\pdfoutput=1
\documentclass[letterpaper]{article}
\usepackage{iclr2027_conference,times}
\usepackage{xcolor}

\newif\ifshowcomments
\showcommentsfalse

\newif\ifincludeappendix
\includeappendixtrue

\newif\ifincludebibliography
\includebibliographytrue

\usepackage[T1]{fontenc}
\usepackage[utf8]{inputenc}
\usepackage{amsmath,amssymb,bm}
\usepackage{graphicx}
\usepackage{booktabs,multirow,tabularx}
\usepackage{float}
\usepackage{placeins}
\usepackage{wrapfig}
\usepackage{needspace}
\usepackage{capt-of}
\usepackage[table]{xcolor}
\usepackage{enumitem}
\usepackage{url}
\usepackage{marvosym}
\usepackage{hyperref}
\definecolor{citegreen}{HTML}{006B3D}
\hypersetup{
  colorlinks=true,
  citecolor=citegreen,
  linkcolor=red,
  urlcolor=black,
  filecolor=black,
  pdftitle={RouteREPA: Routed Representation Alignment for Just Image Transformers},
  pdfauthor={Yi Tan, Beier Zhu, Sirui Chen, Cheng Chen, Xingyu Zhu, Yuan Zhou, Zhenshan Tan, Tianrun Chen, Hanwang Zhang}
}
\setlist[itemize]{leftmargin=1.2em}

\newcommand{\ours}{\mbox{RouteREPA}}

\newcommand{\vect}[1]{\bm{#1}}
\newcommand{\mat}[1]{\mathbf{#1}}
\newcommand{\R}{\mathbb{R}}
\newcommand{\E}{\mathbb{E}}
\definecolor{tabhighlight}{RGB}{237,247,248}

\newcommand{\para}[1]{\noindent\textbf{#1.}}

\title{RouteREPA: Routed
Representation Alignment
for Just Image Transformers}
\author{%
Yi Tan\textsuperscript{1},\ Beier Zhu\textsuperscript{2,\Letter},\ Sirui Chen\textsuperscript{3},\ Cheng Chen\textsuperscript{1},\ Xingyu Zhu\textsuperscript{4},\\
\bf Yuan Zhou\textsuperscript{1},\ Zhenshan Tan\textsuperscript{5},\ Tianrun Chen\textsuperscript{6},\ Hanwang Zhang\textsuperscript{1}\\[3pt]
\textsuperscript{1}NTU\quad \textsuperscript{2}USTC\quad \textsuperscript{3}Tongji University\quad \textsuperscript{4}NUS\quad \textsuperscript{5}NUIST\quad \textsuperscript{6}Moxin Technology\\[2pt]
\texttt{yi.tan@ntu.edu.sg}\quad \texttt{beier.zhu@ustc.edu.cn}
}

\iclrfinalcopy

\begin{document}
\maketitle
\lhead{}
\renewcommand{\headrulewidth}{0pt}
{\renewcommand{\thefootnote}{\Letter}\footnotetext{Corresponding author.}}

\begin{abstract}

Representation alignment (REPA) is widely used to accelerate the training of latent-space diffusion transformers (DiTs).
However, applying it to pixel-space models such as Just Image Transformers (JiT) is challenging.
Our analysis reveals a \emph{pronounced representation conflict} between discriminative alignment and pixel-space generation.
Compared with its latent counterpart, JiT learns representations that are substantially less aligned with DINOv2.
Applying REPA to JiT consistently reduces the reconstructability of its intermediate representations, whereas this effect is not observed in Latent JiT.
We attribute this conflict to JiT directly reconstructing pixels, which requires its representations to hierarchically encode discriminative semantics and reconstruction-relevant visual information across network depths.
To address these challenges, we propose Routed Representation Alignment (RouteREPA), which uses lightweight, representation-dependent routers to adaptively route task-oriented representations for alignment and generation.
We further apply routing across multiple network depths to accommodate their depth-dependent representation requirements.
Extensive experiments on ImageNet-1K at $256\times256$ resolution show that RouteREPA consistently improves JiT across model scales and training budgets, outperforming standard REPA.
On JiT-H/16, RouteREPA further surpasses PixelREPA using only one-third of its training epochs without classifier-free guidance.

\end{abstract}

\begin{figure}[H]
  \centering
  \includegraphics[width=\linewidth]{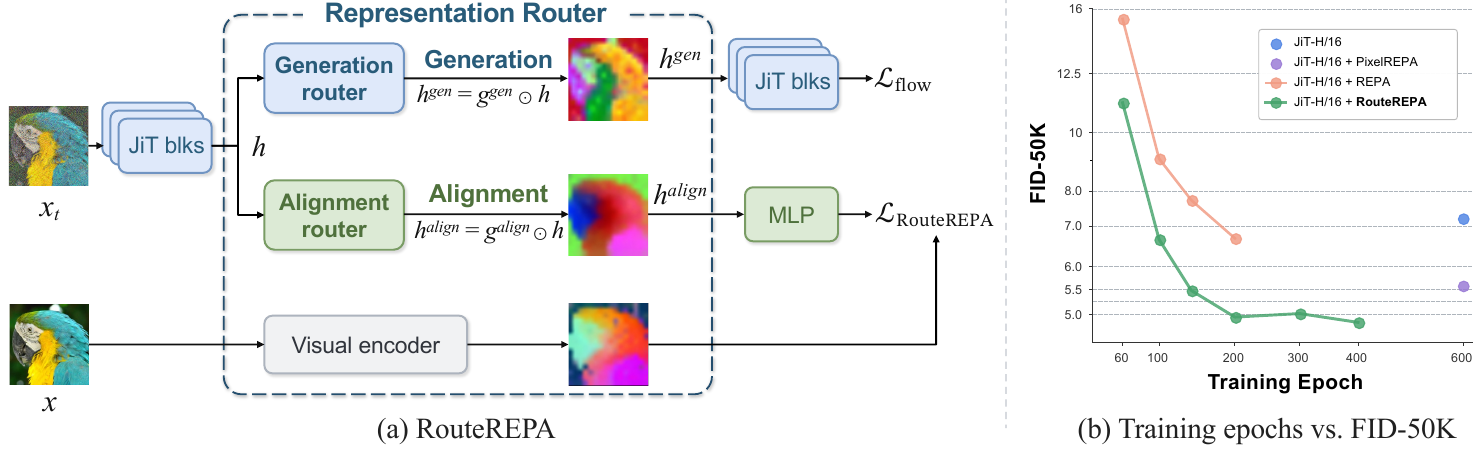}
  \caption{Overview of \ours{}.
  (a) Architecture of \ours{}. Task-specific channel-wise routers direct intermediate JiT representations to the generation and representation alignment branches according to their representation preference.
  (b) FID-50K during ImageNet-1K training at $256\times256$ resolution without CFG. \ours{} requires fewer than 100 and 200 training epochs to match JiT-H/16 and PixelREPA-H/16 trained for 600 epochs, respectively.}
  \label{fig:overview}
  \vspace{-0.2cm}
\end{figure}

\section{Introduction}
\label{sec:intro}

Latent-space diffusion transformers (DiTs), which perform denoising in the compact latent representations produced by pretrained variational autoencoders (VAEs)~\citep{rombach2022ldm,peebles2023dit,ma2024sit} or pretrained visual encoder~\citep{zheng2026rae,singh2026raev2}, have become a dominant paradigm for efficient image generation.
To further improve training efficiency, Representation Alignment (REPA)~\citep{yu2025repa} introduces an auxiliary objective that aligns intermediate representations of the latent-space DiTs with features from DINOv2, providing semantic supervision that accelerates convergence.

More recently, pixel-space DiTs such as Just Image Transformers (JiT)~\citep{li2025jit} have attracted increasing attention as they remove the need for a pretrained image encoder.
Removing the encoder simplifies the generative pipeline and eliminates the reconstruction bottleneck of latent representations, offering potential advantages for scaling.
Inspired by the success of REPA in latent-space DiTs, a natural question is \textit{whether representation alignment can further improve pixel-space generation with JiT?}

Although representation alignment encourages DiTs to learn discriminative visual features, it may conflict with reconstruction-relevant information for image generation. This challenge is particularly pronounced in pixel-space DiTs:
Latent-space DiTs operate on pretrained  representations, where \textit{the encoder already captures perceptually meaningful structures}, including semantic information useful for alignment.
Therefore, REPA provides complementary supervision rather than \textit{requiring the model to learn these features from scratch}.
In contrast, JiT directly models RGB images and must learn both discriminative and generation-related features from raw pixels.
Applying representation alignment may force JiT to allocate additional capacity toward recognition-oriented features that are not directly required for pixel reconstruction, reducing the effectiveness of standard REPA.

\begin{figure}[t]
    \centering
    \includegraphics[width=\linewidth]{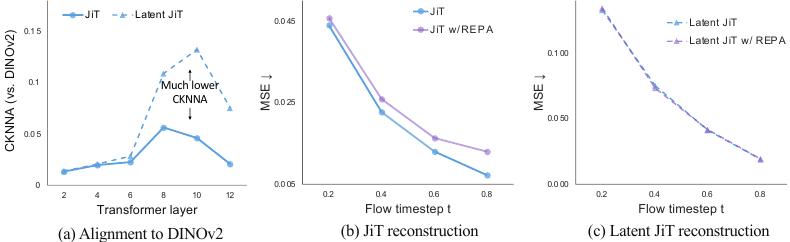}
    \caption{ Pixel-space generation of JiT introduces a \textit{pronounced representation conflict}:
        (a) JiT aligns less closely with DINOv2 than Latent JiT.
        (b) REPA lowers the reconstruction capability of JiT intermediate representations.
        (c) In contrast, no such degradation is observed in Latent JiT.
    }
    \label{fig:pixel-latent-representations}
\end{figure}

To better understand the \textit{pronounced representation conflict} in JiTs, we conduct two analyses:
\textit{(1) JiT representations are less suited for direct alignment than latent diffusion model.}
We first compare the learned representations of pixel-space JiT and a VAE-latent counterpart built with the same Transformer architecture and training recipe.
The two models are trained to comparable class-conditional generation quality on ImageNet-256 (FID $\approx$31).
Following REPA, we use CKNNA to measure alignment with DINOv2.
As shown in Fig.~\ref{fig:pixel-latent-representations}(a), JiT exhibits substantially lower CKNNA than its latent counterpart.
This suggests that JiT representations differ substantially from those of the discriminative teacher.
\textit{(2) Representation alignment suppresses reconstruction-relevant information in JiT representations.}
We further compare the effect of directly applying REPA to JiT and Latent JiT.
Specifically, we examine the changes in their reconstruction performance before and after applying REPA using decoder heads trained on frozen representations.
As shown in Fig.~\ref{fig:pixel-latent-representations}(b, c), JiT representations trained with REPA become consistently harder to decode back into pixel images, indicating reduced reconstruction capability of the aligned intermediate representations.
\footnote{This does not imply that representation alignment degrades JiT's generation performance; rather, it indicates that pixel images become harder to reconstruct directly from the aligned intermediate representations.}
In contrast, no comparable degradation in reconstruction capability is observed for Latent JiT.

Motivated by the analyses above, we propose Routed Representation Alignment (\ours{}) for JiTs.
Its core idea is to route intermediate JiT features into two task-oriented representations for pixel-space generation and alignment with discriminative teacher features, thereby alleviating the severe representation conflict between the two objectives.
We perform this modulation along the channel dimension, which allows feature responses to be adjusted without altering their spatial organization.
Since the relationship between discriminative and reconstruction-relevant information in JiT representations varies across noise levels, training stages, and network depths, \ours{} is designed to adapt to these changing requirements.
Specifically, it uses lightweight, input-dependent routers to separately modulate the representations supplied to the alignment and generation branches, and applies them hierarchically at multiple intermediate layers.
The routers are jointly optimized with the generator, enabling adaptive representation allocation for the two tasks.

Extensive experiments in Sec.~\ref{sec:experiments} demonstrate the effectiveness of \ours.
With negligible additional parameters and training overhead, \ours{} improves training efficiency and generation quality across JiT scales and training budgets.
On JiT-H, \ours{} achieves a lower unguided FID than PixelREPA~\citep{shin2026pixelrepa} using only one-third of its training epochs.

Our contributions are summarized as follows:
\begin{itemize}
    \item We analyze the pronounced representation conflict in representation alignment on JiT: (1) JiT representations are less suited for direct alignment than latent diffusion model; and (2) Representation alignment suppresses reconstruction-relevant infor-
mation in JiT representations.

    \item We propose \ours{}, which adaptively routes features for generation and alignment across multiple network depths to alleviate this pronounced representation conflict.

    \item Across different model scales and training budgets, \ours{} consistently improves JiT generation performance and outperforms both standard REPA and PixelREPA.
\end{itemize}
\par\allowbreak
\section{Related work}
\label{sec:related}

\noindent\textbf{Image generation models.}
Pioneering diffusion models, such as DDPM~\citep{ho2020ddpm} and DDIM~\citep{song2021ddim}, generate images through iterative denoising directly in pixel space.
Latent diffusion models (LDMs) separate perceptual compression from generative modeling by performing denoising in the latent spaces of pretrained autoencoders~\citep{rombach2022ldm}.
This reduces the burden of modeling perceptually insignificant pixel-level details while improving computational efficiency and become a widely adopted paradigm.
Transformer-based frameworks, including DiT~\citep{peebles2023dit} and SiT~\citep{ma2024sit}, further advanced this paradigm by using self-attention to model spatial relationships and scale image generation.
More recently, pixel-space models~\citep{chen2026dip,ma2026deco,yu2026pixeldit} such as JiT~\citep{li2025jit} have gained attention for generating high-resolution images without pretrained autoencoders while remaining competitive with latent-space models.

\noindent\textbf{Representation alignment.}
Representation alignment (REPA)~\citep{yu2025repa} and its extensions~\citep{wu2025reg,leng2025repae,singh2026irepa} have emerged as effective strategies for accelerating latent DiTs training.
REPA aligns projected intermediate features of a denoising network with clean-image representations from an external pretrained visual encoder~\citep{caron2021dino,oquab2023dinov2,radford2021clip,zhai2023siglip,he2022mae}.
This supervision helps the generator learn useful representations more efficiently, improving both convergence speed and generation quality.
However, the success of representation alignment in latent space does not automatically carry over to pixel-space generators.
\par\allowbreak
\section{Method}
\label{sec:method}

\ours{} equips JiT with lightweight, channel-wise routers that modulate the intermediate representations used by the alignment and generation branches.
We first summarize the generative formulation and alignment objective in Sec.~\ref{sec:preliminaries}, then present our framework in Sec.~\ref{sec:method-overview}.

\subsection{Preliminaries}
\label{sec:preliminaries}

\para{Flow and diffusion models}
Diffusion and flow-based generation can be studied through continuous trajectories that connect data with noise~\citep{ma2024sit,lipman2023flow}.
For a sample $\vect{x}\sim p_{\mathrm{data}}$ and independently noise $\vect{\epsilon}\sim\mathcal{N}(\vect{0},\mathbf{I})$, their mixture at $t\in[0,1]$ takes the form $\vect{x}_t = \alpha_t\vect{x} + \sigma_t\vect{\epsilon}$.
The schedules $\alpha_t$ and $\sigma_t$ specify the signal and noise amplitudes.
A learned field $v_\theta$ drives the sampling ODE, $\mathrm{d}\vect{x}_t/\mathrm{d}t=v_\theta(\vect{x}_t,t)$, and is fitted to derivatives of sampled interpolation paths:
\begin{equation}
  \mathcal{L}_{\mathrm{flow}}(\theta)
  = \E_{\vect{x},\vect{\epsilon},t}
  \left[\left\|v_\theta(\vect{x}_t,t)
  -\bigl(\dot{\alpha}_t\vect{x}+\dot{\sigma}_t\vect{\epsilon}\bigr)\right\|_2^2\right].
  \label{eq:flow-loss}
\end{equation}
Here, $\dot{\alpha}_t$ and $\dot{\sigma}_t$ are schedule derivatives.
Linear interpolation with $\alpha_t=t$ and $\sigma_t=1-t$ yields the target $\vect{x}-\vect{\epsilon}$, with generation proceeding from noise at $t=0$ toward data at $t=1$.
Fixing $\alpha_t=1$ instead leaves a noise-only derivative, $\dot{\sigma}_t\vect{\epsilon}$, as in variance-exploding diffusion.
For brevity, class conditioning is not shown.

\para{Just Image Transformers}
JiT~\citep{li2025jit} performs generation directly in pixel space using a Transformer, without a pretrained autoencoder.
Under the linear interpolation above, JiT predicts the clean RGB image $\hat{\vect{x}}_\theta(\vect{x}_t,t)$ rather than velocity, which is recovered as
\begin{equation}
  v_\theta(\vect{x}_t,t)
  = \frac{\hat{\vect{x}}_\theta(\vect{x}_t,t)-\vect{x}_t}{1-t},
  \qquad 0\leq t<1.
  \label{eq:jit-velocity}
\end{equation}
Thus, the network outputs a clean-image estimate but optimized with the velocity loss in Eq.~\eqref{eq:flow-loss}.

\para{Representation alignment}
REPA~\citep{yu2025repa} uses clean-image features to guide intermediate denoising representations.
With a frozen teacher $F$ and a trainable MLP $p_\phi$, the regularizer is
\begin{equation}
  \mathcal{L}_{\mathrm{REPA}}(\theta,\phi)
  = -\frac{1}{N}\sum_{n=1}^{N}
  \E_{\vect{x},\vect{\epsilon},t}
  \left[\operatorname{cos}\!\left(p_\phi(\vect{h}_t^{[n]}),F(\vect{x})^{[n]}\right)\right],
  \label{eq:repa-loss}
\end{equation}
Here, $n$ indexes image patches and $\operatorname{cos}$ denotes cosine similarity.
The total objective is $\mathcal{L}=\mathcal{L}_{\mathrm{flow}}+\lambda\mathcal{L}_{\mathrm{REPA}}$, with alignment weight $\lambda>0$.

\subsection{Routed representation alignment}
\label{sec:method-overview}

Standard representation alignment applies the alignment objective to the same intermediate representation that is subsequently used for image generation.
For JiT, this shared representation must both align with semantically compressed teacher representations and retain the reconstruction-relevant information required for direct pixel prediction, resulting in the representation conflict discussed in Sec.~\ref{sec:intro}.
To address this conflict, \ours{} routes intermediate JiT representations into two task-oriented representations: one for alignment with the discriminative teacher and the other for subsequent image generation.
The routing is learned along the channel dimension and conditioned on the current representation, without manually assigning particular channels to semantics or visual details.
Figure~\ref{fig:overview}(a) illustrates the resulting architecture.

\para{Adaptive channel routing}
Let $\vect{h}_{\ell}\in\R^{N\times C}$ denote the image patch representations after block $\ell$, with $N$ patches and $C$ channels.
We spatially average the representations as $\bar{\vect{h}}_{\ell}=N^{-1}\sum_{n=1}^{N}\vect{h}_{\ell}^{[n]}$ and define task-specific routers for generation and alignment:
\begin{equation}
  \vect{g}_{\ell}^{\tau}
  = \operatorname{sigmoid}\!\left(\mat{W}_{\ell}^{\tau}\bar{\vect{h}}_{\ell}+\vect{b}_{\ell}^{\tau}\right),
  \qquad
  \vect{h}_{\ell}^{\tau}=\vect{g}_{\ell}^{\tau}\odot\vect{h}_{\ell},
  \quad \tau\in\{\mathrm{gen},\mathrm{align}\},
  \label{eq:channel-router}
\end{equation}
where $\tau$ indexes the generation or alignment task, $\mat{W}_{\ell}^{\tau}\in\R^{C\times C}$ and $\vect{b}_{\ell}^{\tau}\in\R^C$ are trainable, and $\odot$ denotes hadamard product and broadcasts over patch positions.
The generation features $\vect{h}_{\ell}^{\mathrm{gen}}$ are passed to the next block, while the alignment features $\vect{h}_{\ell}^{\mathrm{align}}$ are used to align with features from an external pretrained encoder.
Our default design couples these task-specific routers through a shared parameterization: a single prediction $\vect{g}_{\ell}$ is used as $\vect{g}_{\ell}^{\mathrm{gen}}=\vect{g}_{\ell}$ and $\vect{g}_{\ell}^{\mathrm{align}}=\vect{1}-\vect{g}_{\ell}$, yielding complementary routing.
The routers softly modulate channel responses rather than imposing a hard feature partition, so the two tasks can still share useful information.
Within this general formulation, standard REPA corresponds to identity routing for both branches, such that generation and alignment consume the same unmodulated representation.
Gradients from both objectives update the routing parameters jointly with the generator.
This design enables JiT to adaptively route representations between generation and alignment, with routing decisions that evolve during training and respond to the timestep and characteristics of each sample.

\para{Hierarchical routing}
JiT learns the complete pixel-representation-pixel compression and reconstruction process across different noise levels during training.
As representations evolve through the network, the conflict between alignment and generation likewise changes across depths.
This motivates revisiting hierarchical representation routing at multiple locations instead of relying on a single alignment point.
We therefore apply \ours~at a set of intermediate layers $\mathcal{S}$.
Each pair of generation and alignment routers responds to its current representations rather than imposing the same channel assignment throughout the network.
The generation stream continues through the remaining blocks, while the selected alignment representations are summed across layers at corresponding patch positions and passed through a shared projector.

\para{Training and inference}
Using cosine similarity as in Eq.~\eqref{eq:repa-loss}, the \ours{} alignment loss is
\begin{equation}
  \mathcal{L}_{\mathrm{RouteREPA}}
  = -\frac{1}{N}\sum_{n=1}^{N}\E_{\vect{x},\vect{\epsilon},t}
  \left[\operatorname{cos}\!\left(
    p_\phi\!\left(\sum_{\ell\in\mathcal{S}}\vect{h}_{\ell}^{\mathrm{align},[n]}\right),
    F(\vect{x})^{[n]}\right)\right].
  \label{eq:route-objective}
\end{equation}
The total training loss combines this alignment loss with the flow-matching loss:
\begin{equation}
  \mathcal{L} = \mathcal{L}_{\mathrm{flow}}+\lambda\mathcal{L}_{\mathrm{RouteREPA}}.
  \label{eq:route-total-loss}
\end{equation}
\ours~hierarchically provides adaptive task-oriented representations for alignment and generation, helping to mitigate the representation conflict.
At inference, the teacher, projector, and alignment branches are discarded, while the shared routing parameters continue to modulate the generation streams.
\par\allowbreak
\section{Experiments}
\label{sec:experiments}

We present an extensive empirical evaluation of \ours{}, examining its effectiveness, the contributions of different design choices, and the properties it introduces to JiT.
In particular, we address five research questions:
\begin{enumerate}[label=\textbf{RQ\arabic*:},leftmargin=3.4em,itemsep=2pt,topsep=4pt]
  \item Does \ours{} improve training efficiency and generation quality across JiT scales?
  \item How do different design choices affect the performance of \ours{}?
  \item How does \ours{} route representations?
  \item How computationally efficient is \ours{}?
  \item Can \ours{} generalize to latent-space diffusion models?
\end{enumerate}
\subsection{Experimental setup}
\label{sec:setup}

\para{Training configuration}
Unless otherwise specified, we strictly follow the original JiT settings~\citep{li2025jit}.
Specifically, we train on the ImageNet-1K training split~\citep{deng2009imagenet,russakovsky2015imagenet}.
Images are resized with their aspect ratios preserved, center-cropped to $256\times256$, and randomly flipped horizontally.
The backbones are JiT-B/16, JiT-L/16, and JiT-H/16; each processes RGB inputs as $16\times16$ patches.
We retain the JiT optimization recipe for a fair comparison: AdamW updates use 1,024 images each, with a five-epoch warm-up to a learning rate of $2\times10^{-4}$.
DINOv2-B~\citep{oquab2023dinov2} supplies the frozen alignment targets.
Appendix~\ref{app:experimental-details} provides additional experimental details.

\para{Sampling}
Following JiT's sampling settings, we use a 50-step Heun solver.

\para{Compared methods}
We use JiT and JiT trained with REPA~\citep{yu2025repa} as controlled baselines, using the same backbone architectures and experimental protocols as \ours{}.
We further compare with advanced generative models in two categories.
The latent-space group includes DiT~\citep{peebles2023dit}, SiT~\citep{ma2024sit}, and SiT trained with REPA or REG~\citep{wu2025reg}.
The pixel-space group includes ADM-G~\citep{dhariwal2021adm}, VDM++~\citep{kingma2023vdm}, PixelFlow~\citep{chen2025pixelflow}, PixNerd~\citep{wang2026pixnerd}, DiP~\citep{chen2026dip}, DeCo~\citep{ma2026deco}, PixelDiT~\citep{yu2026pixeldit}, and PixelREPA~\citep{shin2026pixelrepa}.

Additional experimental settings are provided in Appendix~\ref{app:experimental-details}.

\subsection{The effectiveness of RouteREPA (\textbf{RQ1})}
\label{sec:main-results}

To assess the significance of \ours{} across model scales, we first conduct comparisons on three backbone sizes: JiT-B/16, JiT-L/16, and JiT-H/16.
We use vanilla JiT, JiT with REPA, and JiT with PixelREPA as baselines and compare their FID without classifier-free guidance (CFG)~\citep{ho2022cfg}.
As shown in Table~\ref{tab:main-results}, \ours{} substantially lowers the FID of JiT across model sizes and training budgets.
Compared with standard REPA and PixelREPA, which is specifically designed for JiT, \ours{} achieves the lowest FID across model sizes and training stages.
Notably, on JiT-B, the FID of \ours{} at 200 epochs nearly matches that of PixelREPA at 600 epochs (22.2 vs.\ 22.1).
On JiT-H, \ours{} at 200 epochs already achieves a lower FID than PixelREPA at 600 epochs (4.96 vs.\ 5.52).

\begin{table}[t]
  \centering
  \begingroup
  \footnotesize
  \begin{minipage}[t]{0.49\textwidth}
    \centering
    \caption{FID comparison without CFG.
RouteREPA consistently improves JiT across different model scales and training budgets, matching or outperforming REPA and PixelREPA with only one-third of the training epochs.}
    \label{tab:main-results}
    \vspace{3pt}
    \renewcommand{\arraystretch}{1.0}
    \setlength{\tabcolsep}{3.5pt}
    \begin{tabularx}{\linewidth}{@{}
      >{\hsize=1.6\hsize\raggedright\arraybackslash}X
      >{\hsize=.9\hsize\centering\arraybackslash}X
      >{\hsize=.7\hsize\centering\arraybackslash}X|
      >{\hsize=.8\hsize\centering\arraybackslash}X@{}}
      \toprule
      Method & Params & Ep. & FID $\downarrow$ \\
      \midrule
      \multicolumn{4}{l}{\textit{JiT-B/16}} \\
      JiT & 131M & 100 & 42.0 \\
      \quad + REPA & 131M & 100 & 35.1 \\
      \quad + PixelREPA & 131M & 100 & 34.4 \\
      \rowcolor{tabhighlight}\quad + \ours{} & 132M & 100 & 30.4 \\
      \addlinespace[1pt]
      JiT & 131M & 200 & 32.2 \\
      \quad + REPA & 131M & 200 & 27.5 \\
      \quad + PixelREPA & 131M & 200 & 27.6 \\
      \rowcolor{tabhighlight}\quad + \ours{} & 132M & 200 & 22.2 \\
      \addlinespace[1pt]
      JiT & 131M & 600 & 25.5 \\
      \quad + PixelREPA & 131M & 600 & 22.1 \\
      \rowcolor{tabhighlight}\quad + \ours{} & 132M & 200 & 22.2 \\
      \midrule
      \multicolumn{4}{l}{\textit{JiT-L/16}} \\
      JiT & 459M & 100 & 21.8 \\
      \quad + REPA & 459M & 100 & 16.0 \\
      \quad + PixelREPA & 459M & 100 & 17.1 \\
      \rowcolor{tabhighlight}\quad + \ours{} & 461M & 100 & 14.0 \\
      \addlinespace[1pt]
      JiT & 459M & 200 & 16.1 \\
      \quad + REPA & 459M & 200 & 12.8 \\
      \quad + PixelREPA & 459M & 200 & 12.9 \\
      \rowcolor{tabhighlight}\quad + \ours{} & 461M & 200 & 10.6 \\
      \addlinespace[1pt]
      JiT & 459M & 600 & 13.8 \\
      \quad + PixelREPA & 459M & 600 & 11.0 \\
      \rowcolor{tabhighlight}\quad + \ours{} & 461M & 200 & 10.6 \\
      \midrule
      \multicolumn{4}{l}{\textit{JiT-H/16}} \\
      JiT & 953M & 600 & 7.20 \\
      \quad + REPA & 953M & 200 & 6.68 \\
      \quad + PixelREPA & 953M & 600 & 5.52 \\
      \rowcolor{tabhighlight}\quad + \ours{} & 956M & 100 & 6.58 \\
      \rowcolor{tabhighlight}\quad + \ours{} & 956M & 200 & 4.96 \\
      \bottomrule
    \end{tabularx}
  \end{minipage}\hfill
  \begin{minipage}[t]{0.49\textwidth}
    \centering
    \caption{FID and IS comparison with SoTA methods under CFG. Bold marks the best JiT-based result.
RouteREPA improves JiT performance, achieving the best FID among methods using JiT as the backbone while remaining competitive with methods that employ specially designed decoupled semantic backbones.}
    \label{tab:guided-results}
    \vspace{3pt}
    \renewcommand{\arraystretch}{1.25}
    \setlength{\tabcolsep}{3.5pt}
    \begin{tabularx}{\linewidth}{@{}
      >{\hsize=1.7\hsize\raggedright\arraybackslash}X
      >{\hsize=.7\hsize\centering\arraybackslash}X|
      >{\hsize=.8\hsize\centering\arraybackslash}X
      >{\hsize=.8\hsize\centering\arraybackslash}X@{}}
      \toprule
      Method & Ep. & FID $\downarrow$ & IS $\uparrow$ \\
      \midrule
      \multicolumn{4}{l}{\textit{Latent diffusion}} \\
      DiT-XL/2 & 1400 & 2.27 & 278.2 \\
      SiT-XL/2 & 1400 & 2.06 & 277.5 \\
      \quad + REPA & 800 & 1.42 & 305.7 \\
      \quad + REG & 480 & 1.40 & 296.9 \\
      \midrule
      \multicolumn{4}{l}{\textit{Pixel diffusion}} \\
      ADM-G & 400 & 4.59 & 186.7 \\
      CDM & 2160 & 4.88 & 158.7 \\
      RIN & 480 & 3.42 & 182.0 \\
      VDM++ & 560 & 2.12 & 267.7 \\
      RDM & 400 & 1.99 & 260.4 \\
      EPG-XXL/16 & 800 & 1.87 & 287.6 \\
      PixelFlow-XL/4 & 320 & 1.98 & 282.1 \\
      PixNerd-XL/16 & 320 & 1.93 & 298.0 \\
      \arrayrulecolor{black!35}
      \cmidrule(l{0.6em}){1-4}
      \arrayrulecolor{black}
      \multicolumn{4}{l}{\shortstack[l]{\textit{Transformer with decoupled semantic backbones}}} \\
      DiP-XL/16 & 600 & 1.79 & 281.9 \\
      DeCo-XL/16 & 600 & 1.69 & 304.0 \\
      PixelDiT & 320 & 1.61 & 292.7 \\
      \arrayrulecolor{black!35}
      \cmidrule(l{0.6em}){1-4}
      \arrayrulecolor{black}
      \multicolumn{4}{l}{\textit{Transformer only}} \\
      JiT-H/16 & 600 & 1.86 & 303.4 \\
      \quad + PixelREPA & 600 & 1.81 & \textbf{317.2} \\
      \rowcolor{tabhighlight}\quad + \ours{} & 400 & \textbf{1.72} & 316.2 \\
      \bottomrule
    \end{tabularx}

  \end{minipage}
  \endgroup
  \vspace{-1.35\baselineskip}
\end{table}

\label{sec:guided-results}
Finally, we compare RouteREPA against a broader set of methods, including PixelREPA using the same JiT-H backbone; latent diffusion methods based on DiT and SiT; pixel diffusion models including ADM-G, CDM, RIN, VDM++, RDM, EPG-XXL/16, PixelFlow-XL/4, and PixNerd-XL/16; and pixel diffusion Transformers with decoupled semantic backbones (first proposed in DDT~\citep{wang2025ddt}), including DiP, DeCo, and PixelDiT.
As shown in Table~\ref{tab:guided-results}, RouteREPA significantly improves JiT and achieves the lowest FID among methods using the JiT architecture (1.72), together with a high IS of 316.2.
Compared with methods that substantially modify the diffusion Transformer backbone by introducing a separate low-resolution branch for semantic modeling, RouteREPA achieves the highest IS, outperforms DiP and remains competitive with DeCo and PixelDiT in FID.

\begin{figure}
  \centering
  \includegraphics[width=\linewidth]{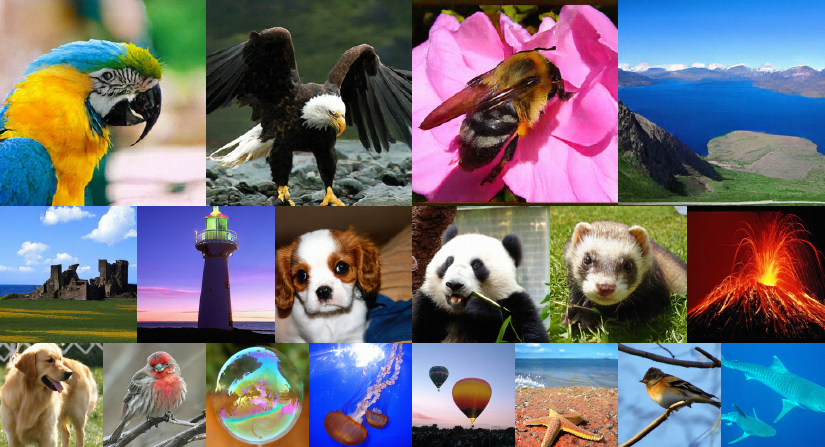}
  \caption{Images generated by JiT-H/16 + \ours{}. Additional samples appear in Appendix~\ref{app:qualitative}.}
  \label{fig:qualitative}
\end{figure}

\vspace{-0.65\baselineskip}
\subsection{Ablation studies (\textbf{RQ2})}

We examine three design choices in RouteREPA: the routing strategy, routing configuration, and routing locations.
Table~\ref{tab:ablations} reports the JiT-B/16 ablations at 200 epochs with class condition.

We first compare standard representation alignment, a learnable orthogonal decomposition, and RouteREPA's shared complementary routing at layer 4.
All three improve upon vanilla JiT.
Relative to standard REPA, orthogonal decomposition yields only a marginal FID improvement (27.5 to 27.4) while reducing IS (58.4 to 57.9).
In contrast, RouteREPA lowers FID to 24.2 and raises IS to 63.4.
This suggests that generation and representation alignment favor different features while still benefiting from shared representations.

We next compare generation-only routing, alignment-only routing, two routers with independent parameters, and our shared-parameter design.
As shown in Table~\ref{tab:ablations}, routing only one branch provides no improvement or only a marginal improvement over standard REPA.
Independent routers provide a clear gain, while sharing the router parameters with complementary generation and alignment weights performs best.

Finally, we compare routing at layer 4, the standard REPA baseline depth, with routing at the deeper layer 6, as well as their combination.
This is motivated by JiT having to model the full hierarchy from pixel-level appearance to high-level semantics within a single backbone, making useful alignment signals span multiple depths.
At 200 epochs, layer 6 gives modestly better FID and IS than layer 4 for both standard REPA and RouteREPA.
Applying RouteREPA with shared routing at both layers gives its best result, with an FID of 22.2 and an IS of 67.0, validating our hypothesis above.
We further investigate whether hierarchical alignment is also beneficial for standard REPA.
Applying REPA at both layers 4 and 6 uses the same alignment configuration as RouteREPA and is equivalent to removing the routers from the latter.
As shown in Table~\ref{tab:ablations}, standard REPA at layers 4 and 6 does not outperform its best single-layer variant, whereas RouteREPA at the same two layers outperforms both single-layer variants. This result shows that adaptive routing of task-oriented representations unlocks the potential of hierarchical representation alignment, which standard REPA fails.

\begin{table}
  \centering
  \setlength{\belowcaptionskip}{4pt}
  \caption{Ablations of \ours{} on JiT-B/16.
  We compare different routing strategies, routing configurations, and routing locations.
  The compared design choices are highlighted in blue, and the best result within each comparison is shown in bold.}
  \label{tab:ablations}
  \begingroup
  \small
  \setlength{\tabcolsep}{10pt}
  \renewcommand{\arraystretch}{1.05}
  \begin{tabular}{lccc|c|cc}
    \toprule
    Method & \shortstack{Routing Strategy} & \shortstack{Routing config.} & Layers & Params & FID $\downarrow$ & IS $\uparrow$ \\
    \midrule
    JiT-B/16 & -- & -- & -- & 131M & 32.2 & 50.1 \\
    \midrule
    \quad + REPA & \cellcolor{tabhighlight}-- & -- & 4 & 131M & 27.5 & 58.4 \\
    \quad + REPA & \cellcolor{tabhighlight}Orthogonal & -- & 4 & 131M & 27.4 & 57.9 \\
    \quad + \ours{} & \cellcolor{tabhighlight}Gated & Shared & 4 & 132M & \textbf{24.2} & \textbf{63.4} \\
    \midrule
    \quad + \ours{} & Gated & \cellcolor{tabhighlight}Gen. only & 4 & 132M & 27.3 & 58.6 \\
    \quad + \ours{} & Gated & \cellcolor{tabhighlight}Align. only & 4 & 131M & 28.0 & 56.5 \\
    \quad + \ours{} & Gated & \cellcolor{tabhighlight}Independent & 4 & 132M & 26.7 & 59.0 \\
    \quad + \ours{} & Gated & \cellcolor{tabhighlight}Shared & 4 & 132M & \textbf{24.2} & \textbf{63.4} \\
    \midrule
    \quad + \ours{} & Gated & Shared & \cellcolor{tabhighlight}4 & 132M & 24.2 & 63.4 \\
    \quad + \ours{} & Gated & Shared & \cellcolor{tabhighlight}6 & 132M & 23.4 & 65.1 \\
    \quad + \ours{} & Gated & Shared & \cellcolor{tabhighlight}$4+6$ & 132M & \textbf{22.2} & \textbf{67.0} \\
    \midrule
    \cellcolor{tabhighlight}\quad + REPA & -- & -- & 4 & 131M & 27.5 & 58.4 \\
    \cellcolor{tabhighlight}\quad + REPA & -- & -- & 6 & 131M & 26.7 & 59.4 \\
    \cellcolor{tabhighlight}\quad + REPA & -- & -- & $4+6$ & 131M & 27.0 & 58.7 \\
    \cellcolor{tabhighlight}\quad + \ours{} & Gated & Shared & $4+6$ & 132M & \textbf{22.2} & \textbf{67.0} \\
    \bottomrule
  \end{tabular}
  \endgroup
  \vspace{-0.6\baselineskip}
\end{table}

\subsection{Analysis of learned routing (\textbf{RQ3})}
\label{sec:routing-analysis}

\textbf{Does \ours~mitigate representation conflict?}
Following REPA~\citep{yu2025repa}, we use CKNNA to quantify the alignment between intermediate JiT features and clean-image DINOv2 representations.
CKNNA measures the agreement between the local neighborhood structures of two representation spaces, where a higher value indicates stronger representation alignment.
As shown in Fig.~\ref{fig:routing-feature-analysis}(a),
the alignment-routed representation achieves comparable alignment to standard REPA at the high-noise timestep ($t=0.2$), and increasingly outperforms REPA as the noise level decreases.
The generation-routed representation, in contrast, remains substantially less aligned with DINOv2.

Under identical decoding probes, Fig.~\ref{fig:routing-feature-analysis}(b,c) shows that the generation route also yields lower MSE and LPIPS than the alignment representations and achieves slightly better reconstruction than the raw representations at every evaluated timestep.
This indicates that RouteREPA uses the alignment and generation routers to select representations suited to their respective objectives, rather than enforcing strong alignment directly on the entire representation intended for generation.
Meanwhile, the system-level results in Tables~\ref{tab:main-results} and~\ref{tab:guided-results} show that RouteREPA also achieves substantially better end-to-end generation performance than JiT and REPA.
Together, these results demonstrate that the learned routing effectively allocates representations according to the distinct needs of representation alignment and image generation.
Qualitative representation-to-pixel reconstructions are provided in Appendix~\ref{app:feature-decoding}.

\vspace{-0.6\baselineskip}
\begin{center}
  \centering
  \setlength{\abovecaptionskip}{3pt}
  \setlength{\belowcaptionskip}{4pt}
  \includegraphics[width=\linewidth]{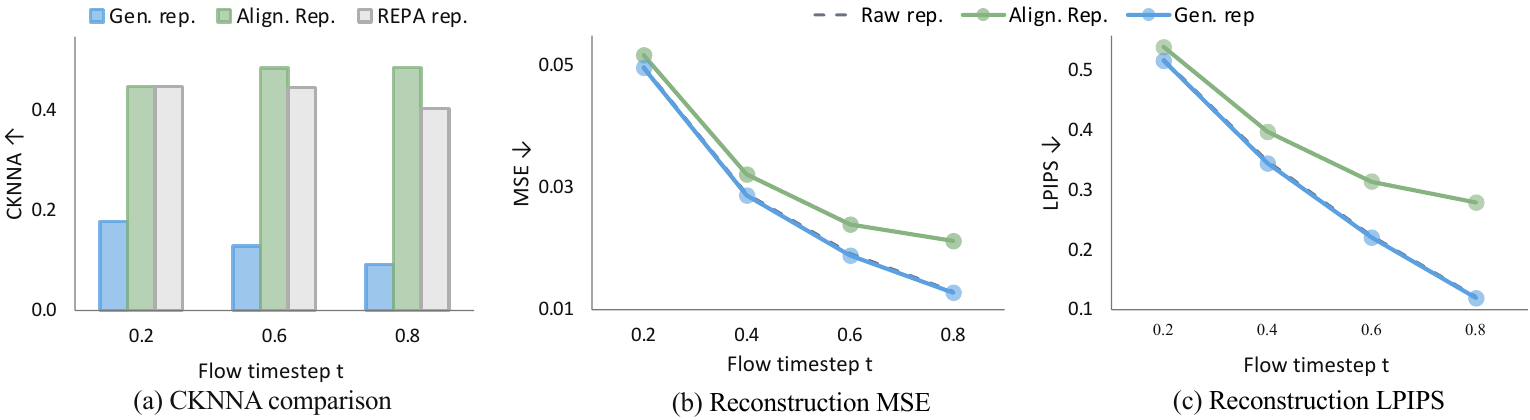}
  \captionof{figure}{Analysis of generation and alignment representations.
  (a) Alignment-routed representations align with DINOv2 even better than REPA.
  (b, c) Generation-routed representations better preserve decodable information for image structure and local appearance, while such reconstruction-related information is suppressed in alignment-routed representations.
  }

\label{fig:routing-feature-analysis}
\end{center}
\vspace{-0.6\baselineskip}

\begin{figure}[t]
  \centering
  \includegraphics[width=\linewidth]{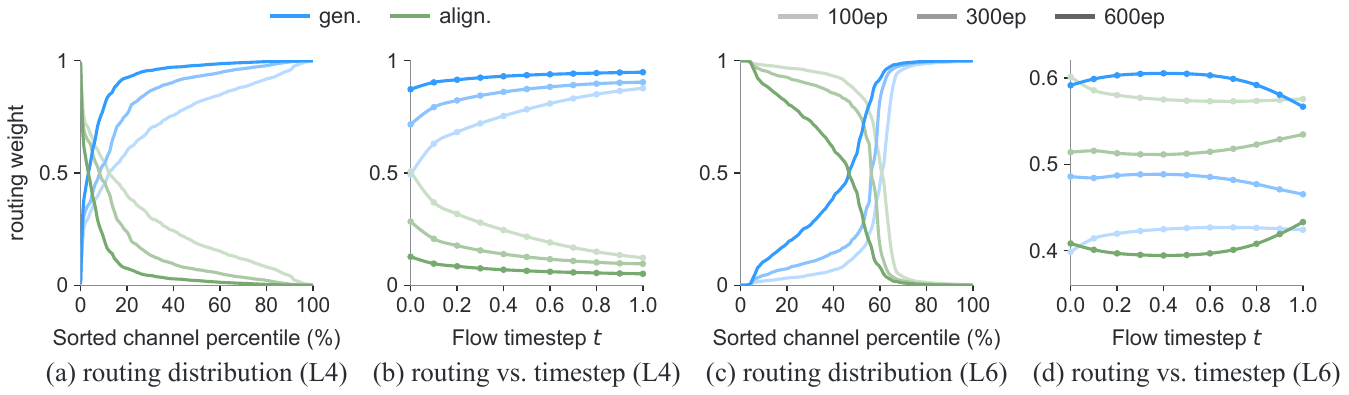}
  \setlength{\abovecaptionskip}{4pt}
  \caption{
Learned routing dynamics at layers 4 and 6.
Blue denotes routing for generation and green for alignment; darker shades indicate later epochs.
Channels are sorted by routing weight.
  }
  \label{fig:routing-analysis}
  \vspace{-1.1\baselineskip}
\end{figure}

\textbf{How Does Routing Evolve?}
We next examine how representation routing evolves across training stages, network depths, and flow timesteps.
Figure~\ref{fig:routing-analysis} shows that the learned routers progressively favor the generation branch at later training stages across timesteps at both layers.
At earlier stages, routers at layers 4 and 6 both assign relatively larger routing weights to the alignment branch in the highly noisy and ambiguous timestep region.
This timestep-dependent variation becomes less pronounced as training proceeds, while the overall routing preference shifts toward generation.
The routers at two layers also exhibit distinct channel-allocation patterns: the router at layer 4 increasingly concentrates alignment on a relatively small subset of channels, whereas the router at layer 6 maintains broader alignment participation.
Together, these observations show that the learned channel-allocation preferences vary with training progress, network depth, and flow timestep.

\vspace{-0.9\baselineskip}
\subsection{Computational efficiency (\textbf{RQ4})}
Table~\ref{tab:jith-efficiency} compares the computational cost of REPA and \ours{} on JiT-H/16.
\ours{} adds only 0.3\% parameters and less than 0.1\% analytical computation, while all measured overheads remain below 2\%.
Full results and the measurement protocol are provided in Appendix~\ref{app:efficiency}.

\vspace{-0.95\baselineskip}
\begin{table}[H]
  \centering
  \setlength{\belowcaptionskip}{4pt}
  \caption{Computational cost on JiT-H/16. \ours{} adds negligible computational overhead compared with REPA.}
  \label{tab:jith-efficiency}
  \begingroup
  \footnotesize
  \renewcommand{\arraystretch}{1.05}
  \setlength{\tabcolsep}{2.6pt}
  \begin{tabular}{l|ccc|ccc}
    \toprule
    Method & Params (M) & \shortstack{Train\\GFLOPs} & \shortstack{Infer.\\GFLOPs} & \shortstack{Peak mem.\\(GiB)} & \shortstack{Train step\\(ms)} & \shortstack{Infer. forward\\(ms)} \\
    \midrule
    REPA & 953 & 1,146 & 364 & 113 & 583 & 48.6 \\
    \rowcolor{tabhighlight}\ours{} & 956 (+0.3\%) & 1,146 ($<0.1\%$) & 364 ($<0.1\%$) & 114 (+0.8\%) & 592 (+1.6\%) & 49.5 (+1.7\%) \\
    \bottomrule
  \end{tabular}
  \endgroup
\end{table}
\vspace{-1.0\baselineskip}

\vspace{-0.95\baselineskip}
\subsection{\ours~in Latent-Space Alignment (\textbf{RQ5})}
\begin{wraptable}{r}{0.39\textwidth}
  \vspace{-1.1\baselineskip}
  \centering
  \setlength{\abovecaptionskip}{0pt}
  \setlength{\belowcaptionskip}{3pt}
  \caption{\ours{} improves SiT-B/2 at 400k iterations (CFG 1).}
  \label{tab:latent-routing}
  \begingroup
  \small
  \setlength{\tabcolsep}{2.8pt}
  \renewcommand{\arraystretch}{0.92}
  \begin{tabular*}{\linewidth}{@{\extracolsep{\fill}}lcc@{}}
    \toprule
    Method & FID $\downarrow$ & IS $\uparrow$ \\
    \midrule
    SiT-B/2 & 33.00 & -- \\
    \quad + REPA & 24.40 & 59.90 \\
    \rowcolor{tabhighlight}\quad + \ours & \textbf{21.85} & \textbf{67.04} \\
    \bottomrule
  \end{tabular*}
  \endgroup
  \vspace{-0.35\baselineskip}
\end{wraptable}

We further examine whether representation routing generalizes beyond pixel-space JiT by applying \ours{} to the latent-space diffusion Transformer SiT-B/2.
All variants are trained for 400K iterations under the same official SiT training setting.
As shown in Table~\ref{tab:latent-routing}, \ours{} substantially improves the vanilla SiT baseline, reducing FID from 33.0 to 21.85.
It also consistently improves upon standard REPA, reducing FID from 24.40 to 21.85 while increasing IS from 59.90 to 67.04.
These results demonstrate that the benefit of task-specific representation routing extends to latent-space diffusion models.

\vspace{-0.5\baselineskip}
\par\allowbreak
\section{Conclusion}
\label{sec:conclusion}
\vspace{-0.5\baselineskip}

In this paper, we introduced \ours{}, a simple and efficient framework for representation alignment in Just Image Transformers.
Our analysis reveals a \emph{pronounced representation conflict} between discriminative alignment and pixel-space generation: JiT representations are naturally less aligned with discriminative teachers, while directly enforcing alignment can suppress reconstruction-relevant information.
To mitigate this conflict, \ours{} learns lightweight, task-specific routing that adaptively allocates intermediate representations to alignment and generation across network depths.
Extensive experiments show that \ours{} consistently improves generation quality and training efficiency across JiT scales, while our analyses further demonstrate meaningful representation specialization and adaptive routing behavior across training stages, noise levels, and network depths.

\textbf{Limitation}. Our current study remains empirical, and the task-dependent routing of intermediate representations still lacks a theoretical understanding.
In addition, due to computational constraints, we have not yet evaluated RouteREPA on substantially larger model scales or more challenging text-to-image generation tasks.
We regard these limitations as important directions for future work and plan to further investigate both the theoretical foundations of representation routing and its scalability to larger and more complex generative settings.\looseness=-1
\par\allowbreak

\ifincludebibliography
  \bibliographystyle{iclr2027_conference}
  \bibliography{refs}
\else
  \section*{References}
\fi

\ifincludeappendix
  \clearpage
  \appendix
  \raggedbottom
  \section{Additional experimental settings}
\label{app:implementation}
\label{app:experimental-details}

This appendix provides the architecture, optimization, alignment, and evaluation settings used in our experiments.
Unless stated otherwise, all configurations follow the JiT recipe~\citep{li2025jit}; the settings specific to representation alignment and routing are summarized separately below.

\subsection{Architecture, training and evaluation configurations}
\label{app:training}

\begin{table}[H]
  \centering
  \caption{Architecture, training, sampling, and alignment configurations. The JiT settings follow the public recipe~\citep{li2025jit}; the final block lists the RouteREPA and standard REPA configurations.}
  \label{tab:appendix-training-config}
  \begingroup
  \footnotesize
  \renewcommand{\arraystretch}{1.05}
  \setlength{\tabcolsep}{6pt}
  \begin{tabular}{@{}p{0.25\linewidth}*{3}{>{\centering\arraybackslash}p{0.13\linewidth}}@{}}
    \toprule
    Setting & JiT-B & JiT-L & JiT-H \\
    \midrule
    \rowcolor{black!10}\multicolumn{4}{l}{\textbf{Architecture}} \\
    Depth & 12 & 24 & 32 \\
    Hidden dimension & 768 & 1,024 & 1,280 \\
    Attention heads & 12 & 16 & 16 \\
    Image size & \multicolumn{3}{c}{$256\times256$} \\
    Patch size & \multicolumn{3}{c}{$16\times16$} \\
    Patch bottleneck & 128 & 128 & 256 \\
    Dropout & 0 & 0 & 0.2 \\
    In-context class tokens & \multicolumn{3}{c}{32} \\
    In-context start block & 4 & 8 & 10 \\
    \midrule
    \rowcolor{black!10}\multicolumn{4}{l}{\textbf{Training}} \\
    Warm-up epochs & \multicolumn{3}{c}{5} \\
    Optimizer & \multicolumn{3}{c}{AdamW, $(\beta_1,\beta_2)=(0.9,0.95)$} \\
    Global batch size & \multicolumn{3}{c}{1,024} \\
    Learning rate & \multicolumn{3}{c}{$2\times10^{-4}$} \\
    Learning-rate schedule & \multicolumn{3}{c}{constant after warm-up} \\
    Weight decay & \multicolumn{3}{c}{0} \\
    EMA decay & \multicolumn{3}{c}{$\{0.9996,0.9999\}$} \\
    Time sampler & \multicolumn{3}{c}{$\operatorname{logit}(t)\sim\mathcal{N}(-0.8,0.8^2)$} \\
    Noise scale & \multicolumn{3}{c}{1.0} \\
    Clip of $(1-t)$ & \multicolumn{3}{c}{0.05} \\
    Class-token drop & \multicolumn{3}{c}{0.1} \\
    \midrule
    \rowcolor{black!10}\multicolumn{4}{l}{\textbf{Sampling}} \\
    ODE solver & \multicolumn{3}{c}{Heun} \\
    ODE steps & \multicolumn{3}{c}{50} \\
    Time steps & \multicolumn{3}{c}{linear in $[0,1]$} \\
    CFG interval & \multicolumn{3}{c}{$[0.1,1.0]$ when used} \\
    \midrule
    \rowcolor{black!10}\multicolumn{4}{l}{\textbf{Representation alignment}} \\
    \multicolumn{4}{l}{\textit{RouteREPA}} \\
    Alignment layers & $4+6$ & $8+12$ & $10+16$ \\
    Frozen teacher & \multicolumn{3}{c}{DINOv2-B} \\
    Alignment coefficient & \multicolumn{3}{c}{$0.05$} \\
    Projector & \multicolumn{3}{c}{shared 3-layer MLP, hidden width $2C$} \\
    \addlinespace[1pt]
    \multicolumn{4}{l}{\textit{Standard REPA}} \\
    Alignment layer & $4$ & $8$ & $10$ \\
    Frozen teacher & \multicolumn{3}{c}{DINOv2-B} \\
    Alignment coefficient & \multicolumn{3}{c}{$0.05$} \\
    Projector & \multicolumn{3}{c}{shared 3-layer MLP, hidden width $2C$} \\
    \bottomrule
  \end{tabular}
  \endgroup
\end{table}

\para{Alignment coefficient}
We use an alignment coefficient of $0.05$ for both standard REPA and RouteREPA.
Empirically, the JiT generation loss is approximately one order of magnitude smaller than the SiT generation loss under the standard REPA setup.
To keep the generation and alignment objectives on comparable scales, we therefore reduce the standard REPA coefficient of $0.5$ used for SiT~\citep{yu2025repa} by the same factor of ten, yielding $0.05$ for JiT.

\para{Image preprocessing}
We exactly follow JiT's preprocessing: images are downsampled, center-cropped to $256\times256$, randomly horizontally flipped, and mapped to $[-1,1]$.

\para{Metrics}
We assess generation quality using 50,000 generated images, reporting Fr\'{e}chet Inception Distance (FID)~\citep{heusel2017fid} and Inception Score (IS)~\citep{salimans2016improved} following the settings of JiT~\citep{li2025jit}.
For our evaluations, FID and IS use torch-fidelity toolkit.
We also report parameter counts (Params, in millions), including the generator and any VAE decoder used at inference, but excluding training-only alignment modules.

\para{Optimization}
We use AdamW with $(\beta_1,\beta_2)=(0.9,0.95)$ and zero weight decay, which is equivalent to Adam under this setting.
The global batch contains 1,024 images.
The learning rate increases linearly over the first five epochs to $2\times10^{-4}$ and remains constant thereafter.
Following JiT, the flow timestep is sampled from a logit-normal distribution with $\mu=-0.8$ and $\sigma=0.8$.

\para{Performance of compared methods}
For comparisons with other state-of-the-art methods in Table~\ref{tab:guided-results}, we directly adopt the results reported in their original papers.
For the unguided comparisons with JiT and JiT+PixelREPA in Table~\ref{tab:main-results}, we evaluate the official 600-epoch checkpoints of JiT-B/L/H and JiT-B/L/H+PixelREPA using exactly the same evaluation pipeline as RouteREPA.
For the corresponding 100- and 200-epoch results, we reproduce these models using their official code and evaluate them under the same protocol.
For JiT+REPA, we use the same training protocol as RouteREPA, train the models for 200 epochs, and compare them with RouteREPA under an identical training epochs.

\subsection{Orthogonal-decomposition baseline}
\label{app:orthogonal-baseline}

The orthogonal baseline in Table~\ref{tab:ablations} replaces channel-wise gating with a learnable rank-$r$ subspace at layer 4 of JiT-B/16, where $r=128$.
Let $\vect{B}\in\R^{C\times r}$ be a trainable matrix.
At every forward pass, we compute a reduced QR decomposition and use its orthonormal factor $\vect{Q}$, such that $\vect{Q}^{\mathsf T}\vect{Q}=\vect{I}$.
For patch features $\vect{h}\in\R^{N\times C}$, the alignment coordinates, their reconstruction in the original channel space, and the complementary generation features are
\begin{equation}
  \vect{h}^{\mathrm{align}}=\vect{h}\vect{Q},\qquad
  \vect{h}^{\parallel}=\vect{h}\vect{Q}\vect{Q}^{\mathsf T},\qquad
  \vect{h}^{\mathrm{gen}}=\vect{h}-\vect{h}^{\parallel}.
  \label{eq:orthogonal-routing}
\end{equation}
The alignment projector receives the $r$-dimensional coordinates $\vect{h}^{\mathrm{align}}$, while the subsequent JiT blocks receive $\vect{h}^{\mathrm{gen}}$.
The basis is initialized orthogonally and optimized jointly by the generation and alignment objectives.
This construction enforces an exact orthogonal split, in contrast to RouteREPA's soft, overlapping channel reweighting.

  \section{Computational overhead}
\label{app:efficiency}

We compare the computational cost of REPA and RouteREPA across the JiT-B, JiT-L, and JiT-H backbones in Table~\ref{tab:efficiency}.

\begin{table}[t]
  \centering
  \caption{Computational cost of REPA and RouteREPA. Params follow the inference-generator convention of the main tables and exclude the training-only teacher and alignment projector. Parentheses report the relative overhead of RouteREPA over REPA. Inference latency measures one network forward rather than a complete sampling step. FLOPs are analytical estimates; memory and latency are measured on four NVIDIA H200 GPUs.}
  \label{tab:efficiency}
  \begingroup
  \footnotesize
  \renewcommand{\arraystretch}{1.08}
  \setlength{\tabcolsep}{5pt}
  \textbf{(a) Inference-time parameters and analytical compute}\\[3pt]
  \begin{tabular}{@{}llccc@{}}
    \toprule
    Backbone & Method & Params (M) & Train GFLOPs & Infer. GFLOPs \\
    \midrule
    JiT-B & REPA & 131 & 205 & 50 \\
    JiT-B & RouteREPA & 132 (+0.9\%) & 205 ($<0.1\%$) & 50 ($<0.1\%$) \\
    JiT-L & REPA & 459 & 582 & 176 \\
    JiT-L & RouteREPA & 461 (+0.5\%) & 582 ($<0.1\%$) & 176 ($<0.1\%$) \\
    JiT-H & REPA & 953 & 1,146 & 364 \\
    JiT-H & RouteREPA & 956 (+0.3\%) & 1,146 ($<0.1\%$) & 364 ($<0.1\%$) \\
    \bottomrule
  \end{tabular}

  \vspace{7pt}
  \textbf{(b) Measured memory and latency}\\[3pt]
  \begin{tabular}{@{}llccc@{}}
    \toprule
    Backbone & Method & Peak mem. (GiB) & Train step (ms) & Infer. forward (ms) \\
    \midrule
    JiT-B & REPA & 26 & 121 & 9.8 \\
    JiT-B & RouteREPA & 27 (+2.0\%) & 127 (+4.8\%) & 10.4 (+7.0\%) \\
    JiT-L & REPA & 70 & 324 & 30.5 \\
    JiT-L & RouteREPA & 71 (+1.0\%) & 330 (+2.0\%) & 31.3 (+2.7\%) \\
    JiT-H & REPA & 113 & 583 & 48.6 \\
    JiT-H & RouteREPA & 114 (+0.8\%) & 592 (+1.6\%) & 49.5 (+1.7\%) \\
    \bottomrule
  \end{tabular}
  \endgroup
\end{table}

All measurements use the same node with four NVIDIA H200 GPUs and PyTorch 2.10.0+cu128.
Training uses four-rank distributed data parallelism, a batch size of 128 per GPU (512 in total), BF16 autocast, and FP32 parameters and AdamW states.
Each timed training step includes the teacher forward pass, model forward and backward passes, distributed gradient synchronization, the optimizer update, and two EMA updates.
Inference uses a batch size of 32 per GPU and measures one network forward pass.
We use three warm-up iterations followed by eight measured training steps or 20 measured inference forwards.
Peak memory is the maximum allocated memory on one GPU, FLOPs are analytical estimates, and latency and memory are measured values.
Across all three model scales, RouteREPA adds only 0.3--0.9\% inference-time parameters, less than 0.1\% analytical computation, and at most 2.0\% peak training memory.
The measured training-step overhead ranges from 1.6\% to 4.8\%, while single-forward inference overhead ranges from 1.7\% to 7.0\% and decreases with model scale.
These results show that RouteREPA is a lightweight addition to REPA.

  \section{Additional analysis}
\label{app:experiments}
\label{app:additional-analysis-experiments}

We provide additional analyses of the representations selected by RouteREPA, their semantic content, the behavior of single-layer routers, and model behavior on subsets defined by DINOv2 feature neighborhoods.

\subsection{Feature-to-pixel reconstructions}
\label{app:feature-decoding}

To assess the relative difficulty of reconstructing images from each representation, we quantify the reconstruction performance of the raw, generation-routed, and alignment-routed representations.
For each representation, we freeze the RouteREPA backbone and routing parameters and train a decoder block as in Fig.~\ref{fig:routing-feature-analysis}(b, c).
Figure~\ref{fig:feature-decoding} shows the same fixed validation examples at three flow timesteps; lower $t$ corresponds to a more heavily corrupted input in the JiT convention.
The generation-routed representation can be decoded into pixel-space images more easily than the alignment-routed representation, while the raw and generation-routed reconstructions remain visually closer, consistent with the MSE and LPIPS results in Fig.~\ref{fig:routing-feature-analysis}(b, c).

\begin{figure}[t]
  \centering
  \includegraphics[width=0.76\linewidth]{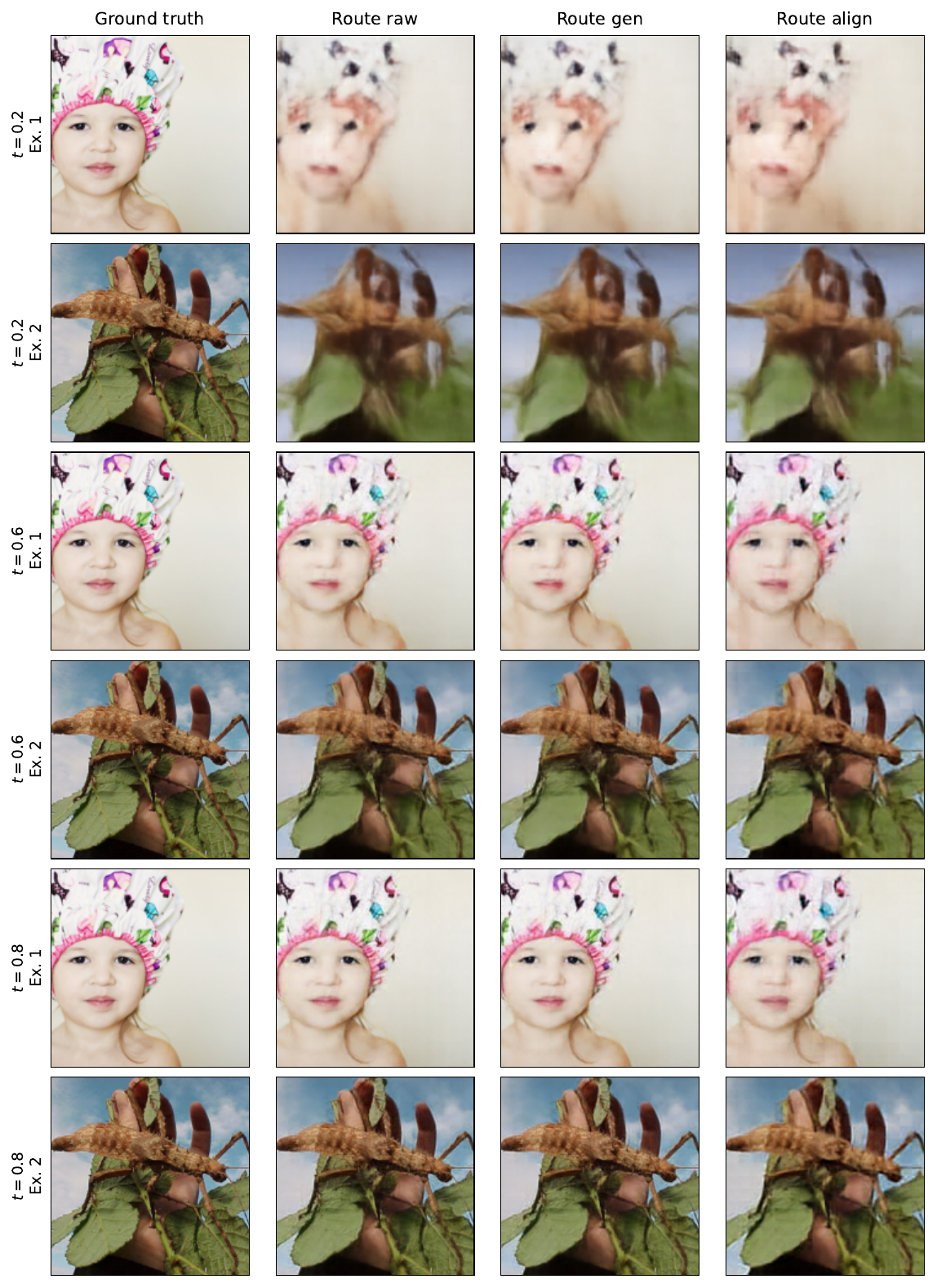}
  \caption{Feature-to-pixel reconstructions at three flow timesteps. Lower $t$ corresponds to a more heavily corrupted input. As more clean-image signal becomes available, the generation route retains more object structure and local appearance than the alignment route.}
  \label{fig:feature-decoding}
\end{figure}

\subsection{Single-layer routing dynamics}
\label{app:single-layer-routing}

We additionally analyze models in which RouteREPA is applied only at layer 4 or only at layer 6.
Figure~\ref{fig:single-layer-routing} compares their routing weights at epochs 100, 300, and 500.
In both single-layer settings, the learned routing weights progressively favor the generation branch as training proceeds.
At earlier checkpoints, the alignment-branch weights are relatively larger for noisier timesteps, while this timestep-dependent variation becomes less pronounced at later checkpoints.

\begin{figure}[H]
  \centering
  \includegraphics[width=\linewidth]{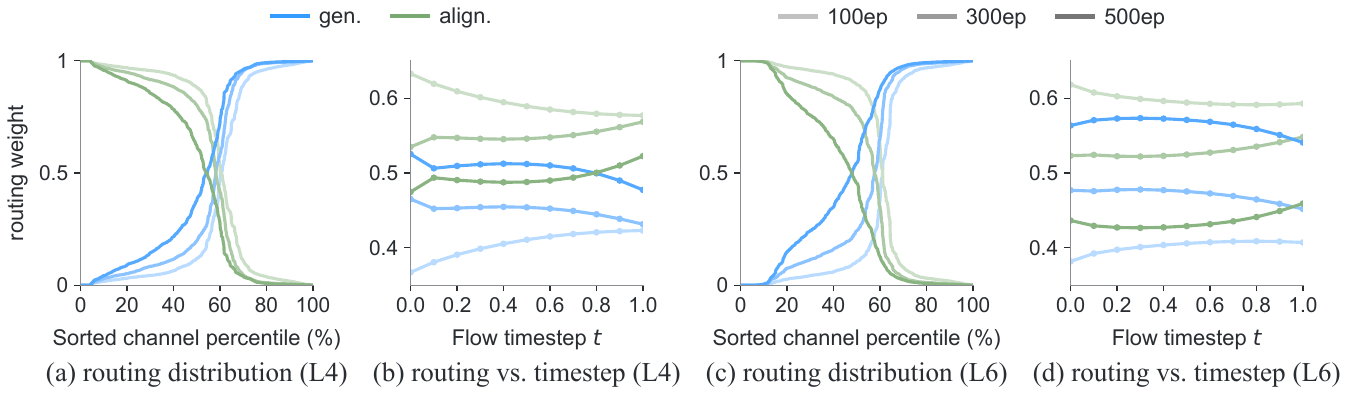}
  \caption{Single-layer routing dynamics in JiT-B/16 models.
  (a,b) L4-only routing; (c,d) L6-only routing.
  Blue denotes generation and green alignment.
}
  \label{fig:single-layer-routing}
\end{figure}

\subsection{An Exploratory Evaluation of the Feature-Hacking Hypothesis}
\label{app:dinov2-neighborhood-analysis}

PixelREPA~\citep{shin2026pixelrepa} proposes the feature-hacking hypothesis to characterize a related challenge in representation alignment for JiT.
Specifically, the hypothesis suggests that alignment with semantically compressed teacher representations may suppress distinctions between images that are similar in representation space but differ in visual details.
PixelREPA evaluates this hypothesis by measuring generation performance drop of REPA on images that are semantically similar in the teacher representation space.
Following its setting, we conduct an exploratory evaluation of this hypothesis using subsets defined by DINOv2 feature similarity.
For each of 100 evaluated ImageNet classes, we compute the class centroid and select the 100 nearest and 100 farthest samples, yielding two 10,000-image subsets.
Each image is perturbed as $x_{0.2}=0.2x_0+0.8\epsilon$ and restored over 50 Heun intervals from $t=0.2$ to $t=1$ using the same image subset and noise across methods.
FID is computed against the corresponding clean subset.
\begin{table}[H]
  \centering
  \caption{FID on DINOv2-neighborhood subsets across guidance scales. We report subset-reference FID after perturbing images at $t=0.2$ and denoising with 50 Heun intervals. Lower is better.}
  \label{tab:dinov2-neighborhood-cfg}
  \begingroup
  \fontsize{9}{10.5}\selectfont
  \setlength{\tabcolsep}{28pt}
  \renewcommand{\arraystretch}{0.92}
  \begin{tabular}{clcc}
    \toprule
    CFG & Method & Nearest 100 & Farthest 100 \\
    \midrule
    1.0 & JiT       & 34.7 & 23.9 \\
        & REPA      & 24.5 & 20.2 \\
        & PixelREPA & 27.5 & 20.6 \\
        & RouteREPA & \textbf{23.1} & \textbf{18.6} \\
    \cmidrule(lr){2-4}
    1.5 & JiT       & 20.0 & 18.0 \\
        & REPA      & 15.2 & 15.9 \\
        & PixelREPA & 16.1 & 16.0 \\
        & RouteREPA & \textbf{14.5} & \textbf{14.7} \\
    \cmidrule(lr){2-4}
    2.5 & JiT       & 10.2 & 15.1 \\
        & REPA      & 9.22 & 13.3 \\
        & PixelREPA & 8.76 & 13.9 \\
        & RouteREPA & \textbf{8.50} & \textbf{12.6} \\
    \cmidrule(lr){2-4}
    3.5 & JiT       & 8.08 & 17.4 \\
        & REPA      & 8.33 & 14.1 \\
        & PixelREPA & 7.20 & 15.4 \\
        & RouteREPA & \textbf{7.11} & \textbf{13.3} \\
    \bottomrule
  \end{tabular}
  \endgroup
\end{table}

Table~\ref{tab:dinov2-neighborhood-cfg} reports the results across guidance scales.
As the CFG scale increases, REPA exhibits different relative behavior on the two subsets.
At CFG 3.5, REPA obtains an FID of 8.33 on the Nearest-100 subset, compared with 8.08 for JiT, while it obtains 14.1 on the Farthest-100 subset, compared with 17.4 for JiT.
Across all evaluated CFG scales, RouteREPA achieves the lowest FID on both subsets.
Figure~\ref{fig:dinov2-neighborhood-examples} presents representative samples from the two subsets.

\begin{figure}[H]
  \centering
  \includegraphics[width=\textwidth]{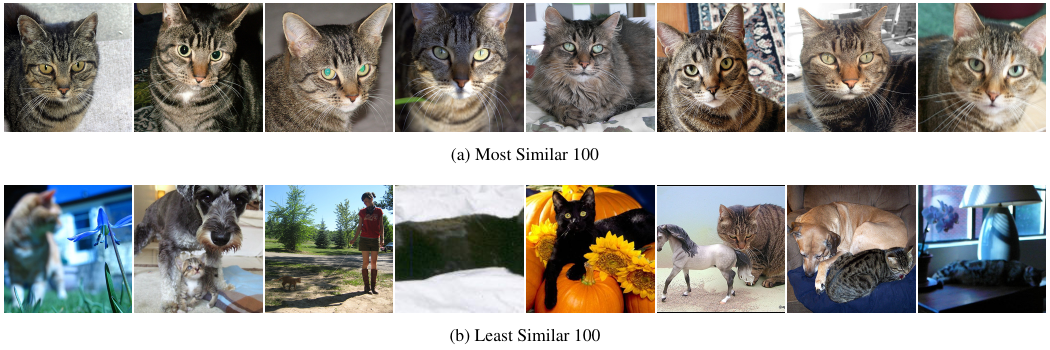}
  \caption{Representative images from the 100 most similar and 100 least similar samples in the \mbox{DINOv2} feature space for the ImageNet tabby class.}
  \label{fig:dinov2-neighborhood-examples}
\end{figure}

  \section{Qualitative results}
\label{app:qualitative}

Figures~\ref{fig:app-qualitative-1}--\ref{fig:app-qualitative-7} present uncurated samples from ImageNet.
All images use JiT-H/16 + \ours{}, a 50-step Heun solver, and CFG 3 throughout the sampling trajectory.

\begingroup
\setlength{\parskip}{0pt}
\setlength{\intextsep}{4pt}
\newcommand{\qualpanel}[2]{%
  \begin{minipage}[t]{0.47\linewidth}
    \centering
    \includegraphics[width=\linewidth]{figs/qualitative/jpeg/#1}\par
    \vspace{1pt}{\fontsize{9}{10}\selectfont class #2}
  \end{minipage}%
}
\newcommand{\qualrow}[4]{%
  \qualpanel{#1}{#2}\hfill\qualpanel{#3}{#4}\par\vspace{4pt}%
}
\newcommand{\qualcenter}[2]{%
  \makebox[\linewidth][c]{\qualpanel{#1}{#2}}\par%
}

\begin{figure}[H]
  \centering
  \qualrow{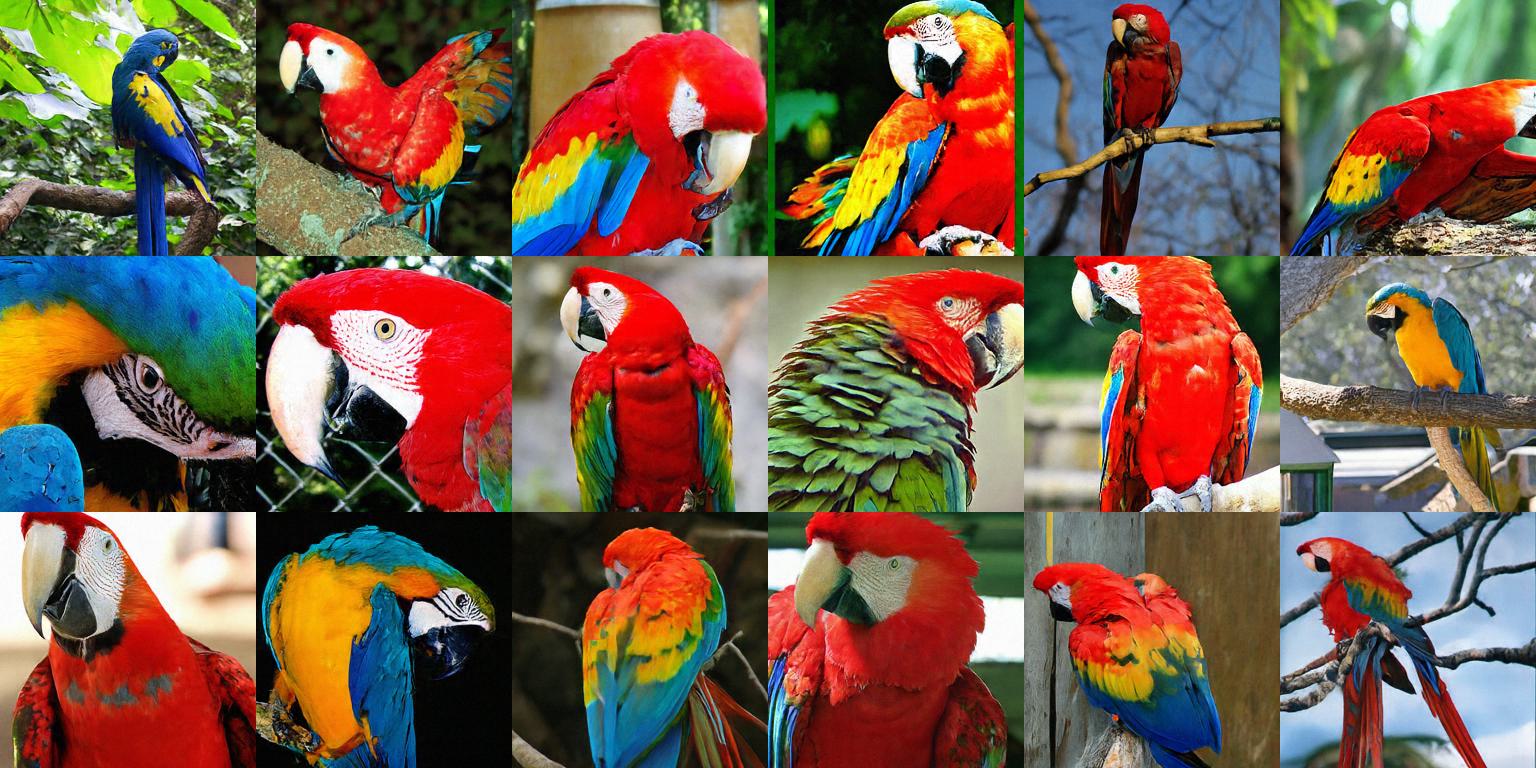}{n01818515}{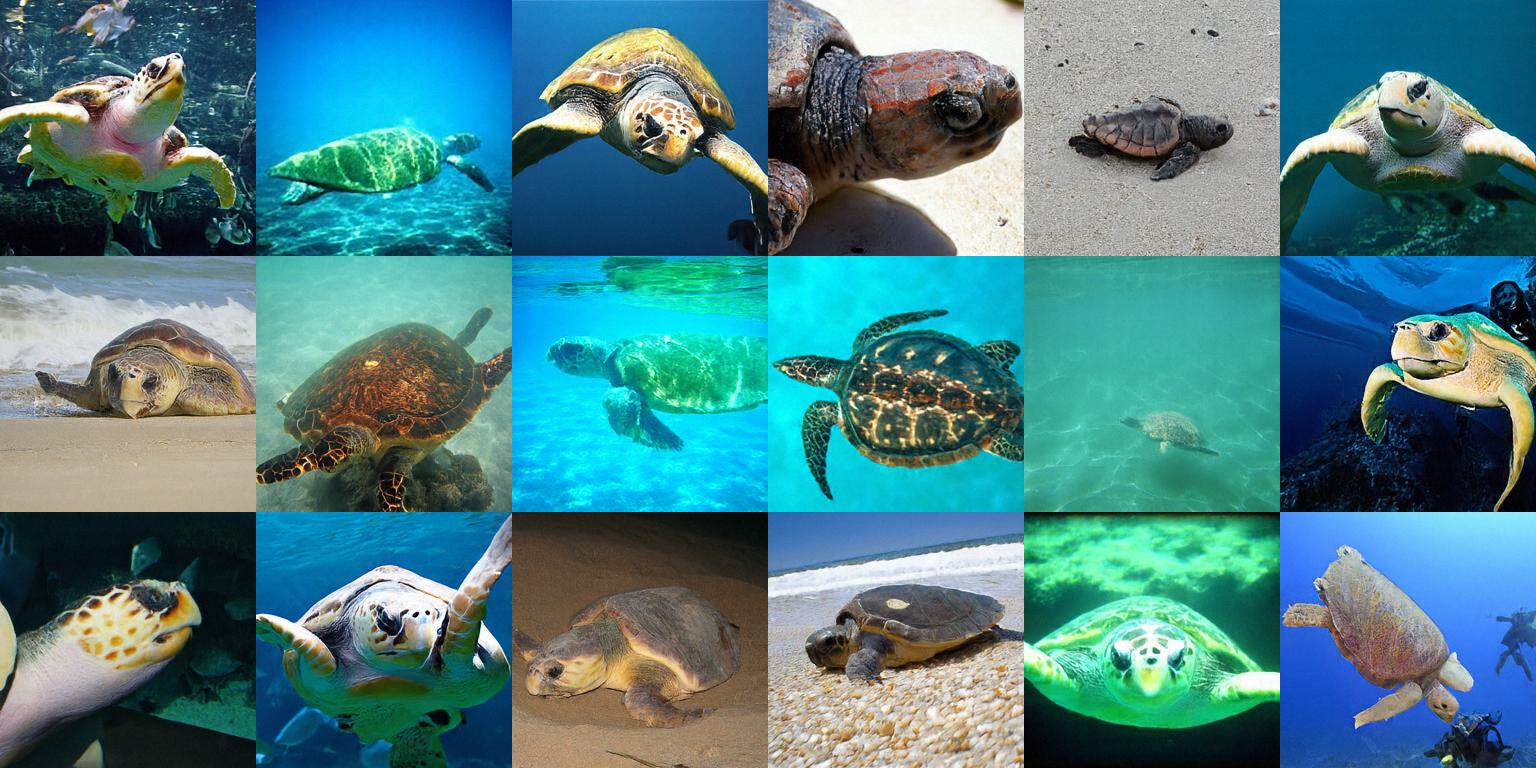}{n01664065}
  \qualrow{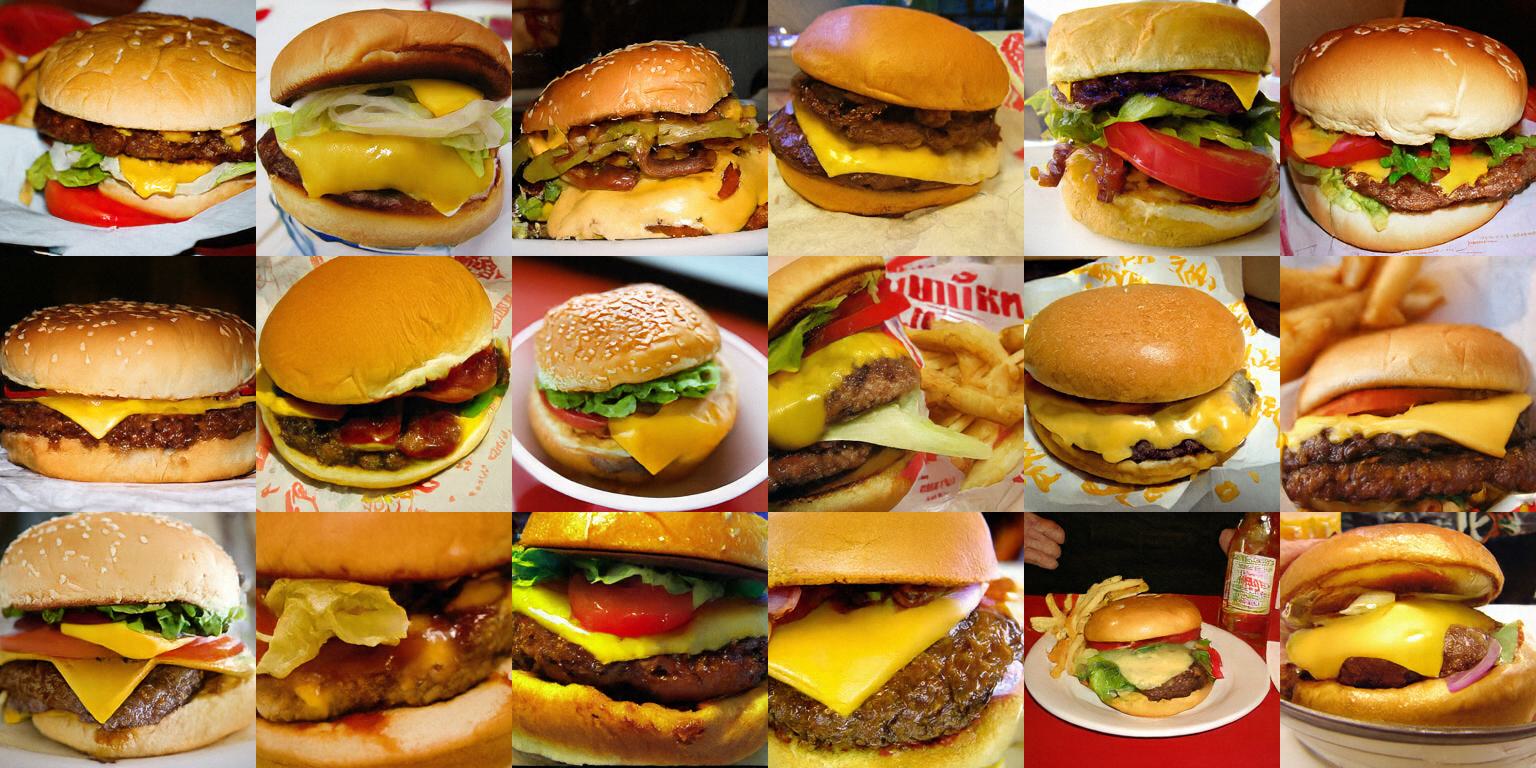}{n07697313}{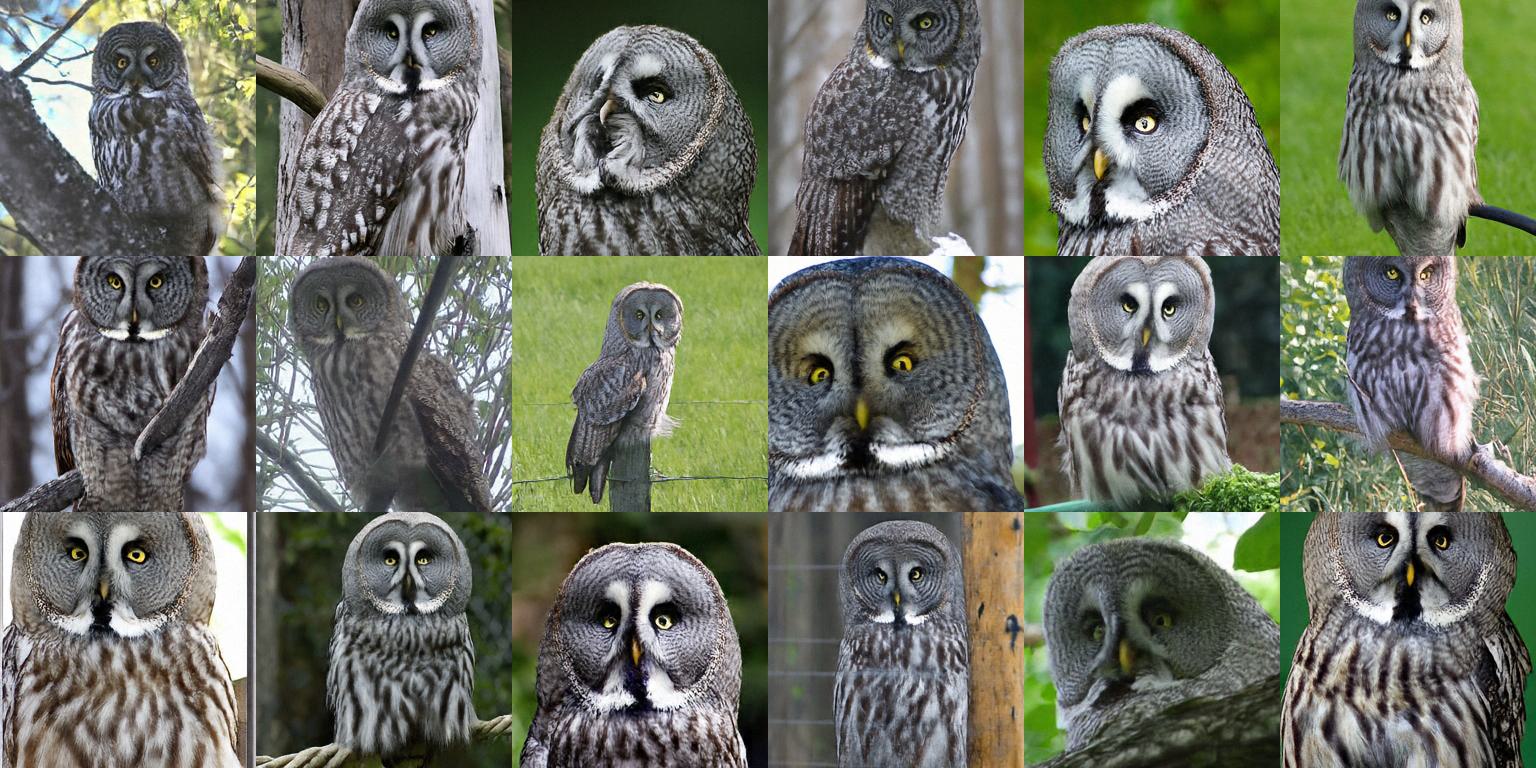}{n01622779}
  \qualrow{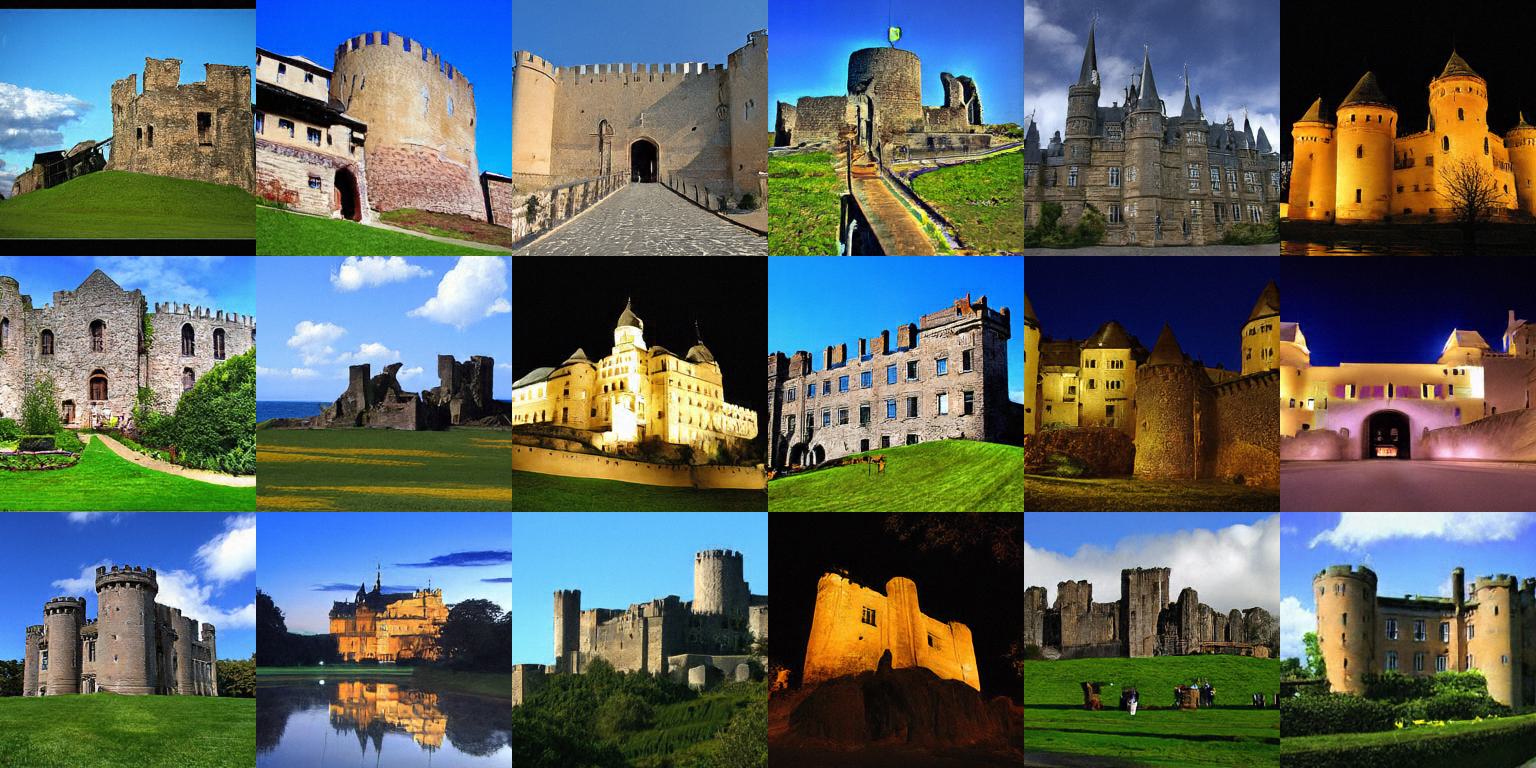}{n02980441}{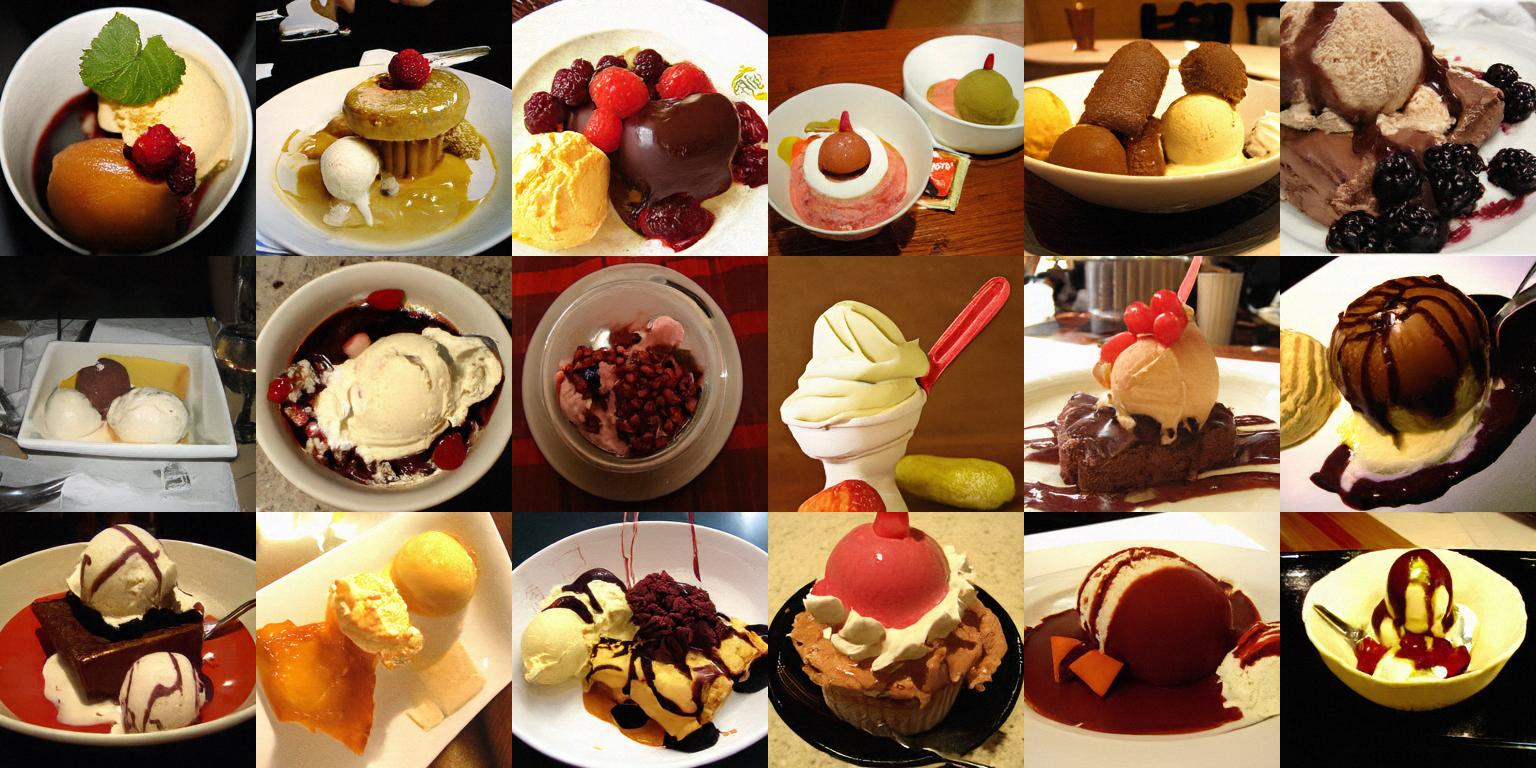}{n07614500}
  \qualrow{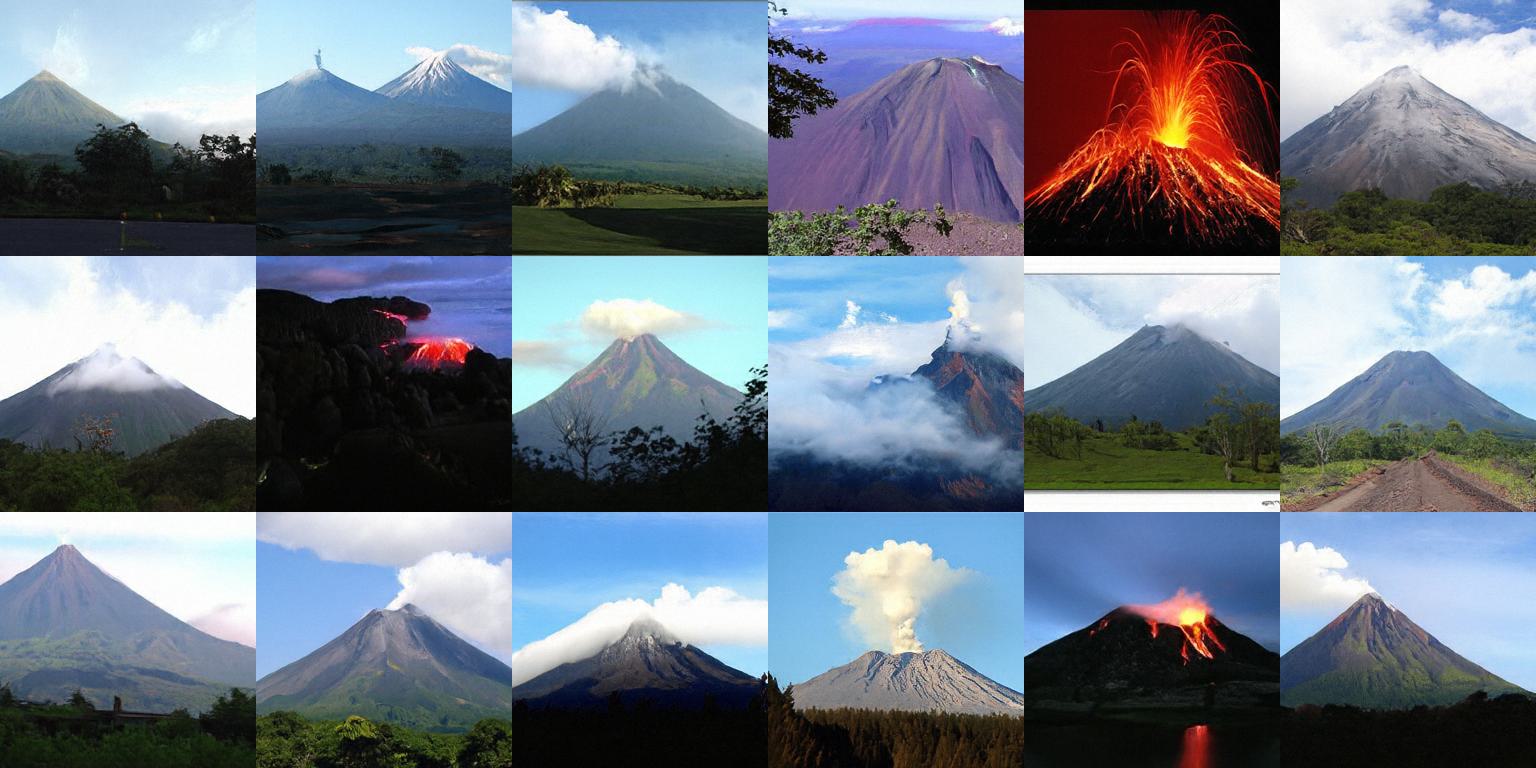}{n09472597}{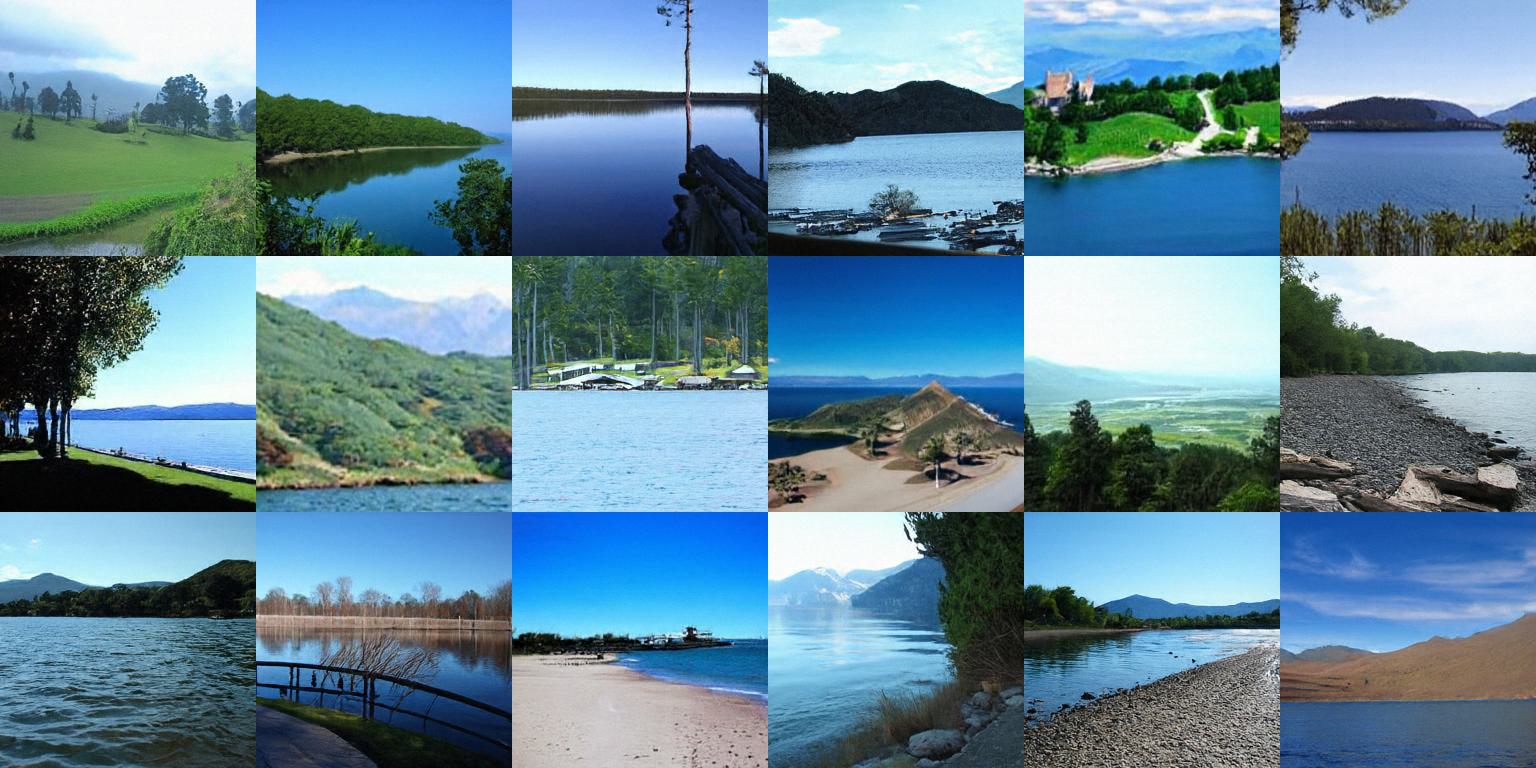}{n09332890}
  \qualcenter{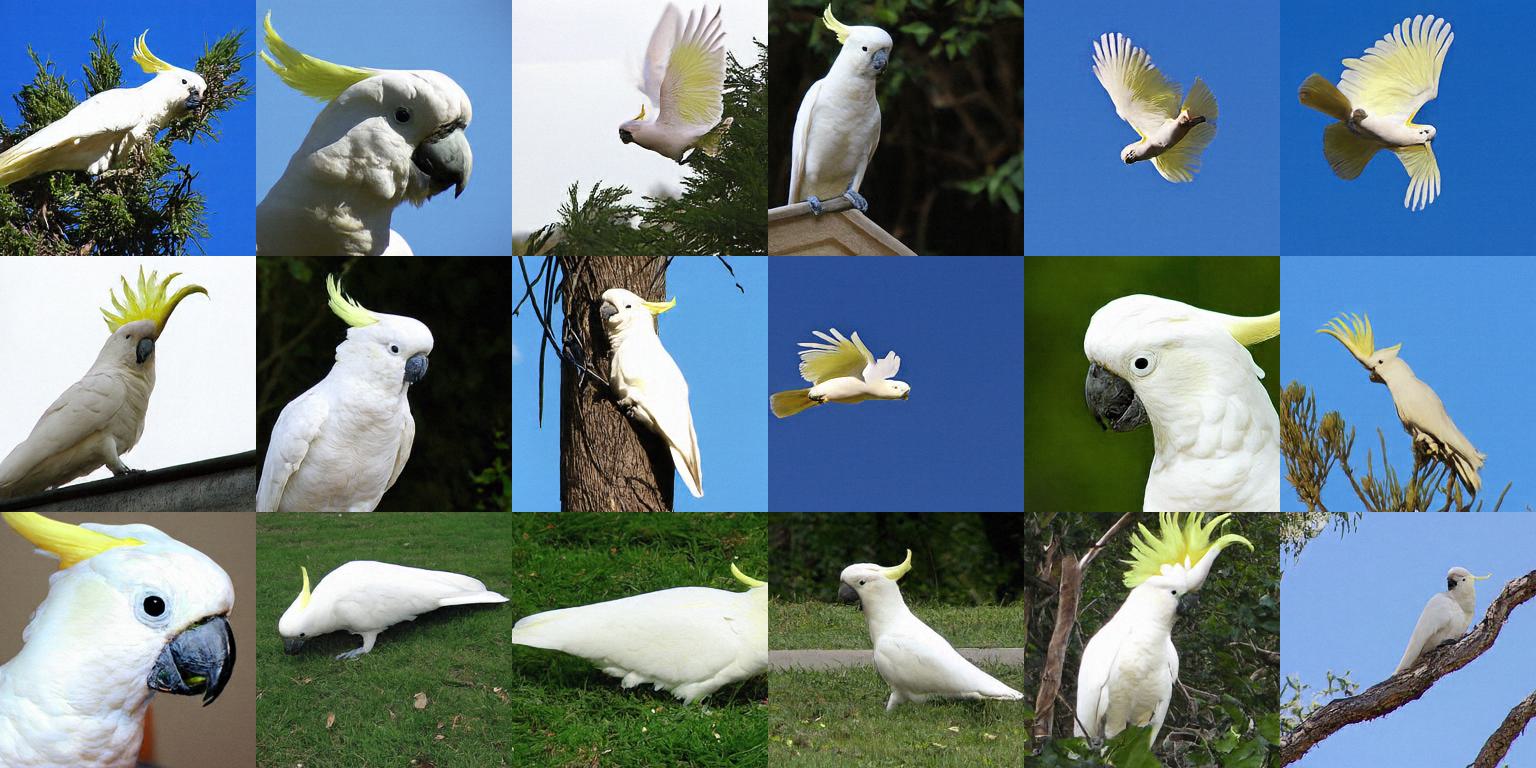}{n01819313}
  \caption{Uncurated samples from JiT-H/16 + \ours{} on ImageNet $256\times256$.}
  \label{fig:app-qualitative-1}
\end{figure}

\begin{figure}[H]
  \centering
  \qualrow{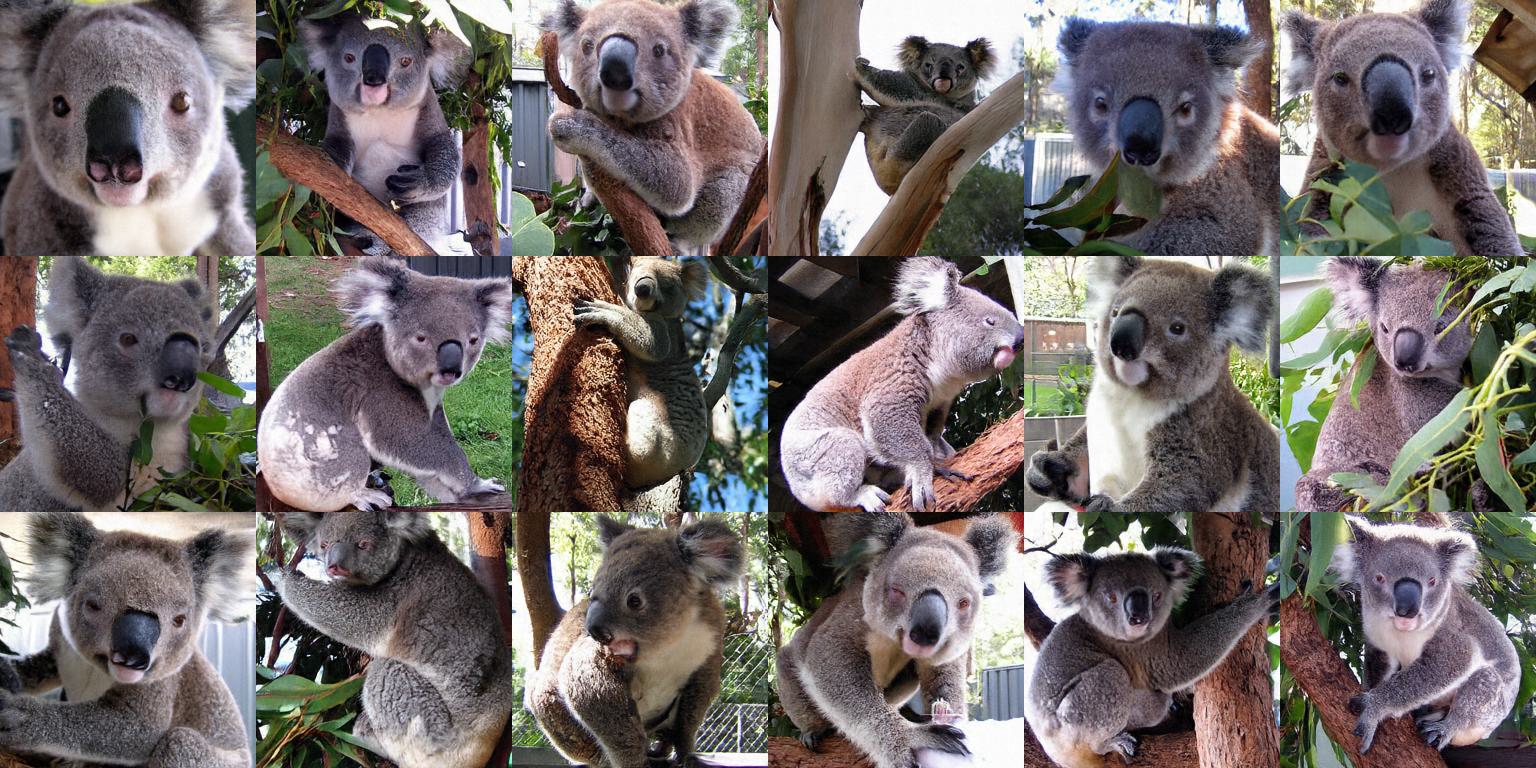}{n01882714}{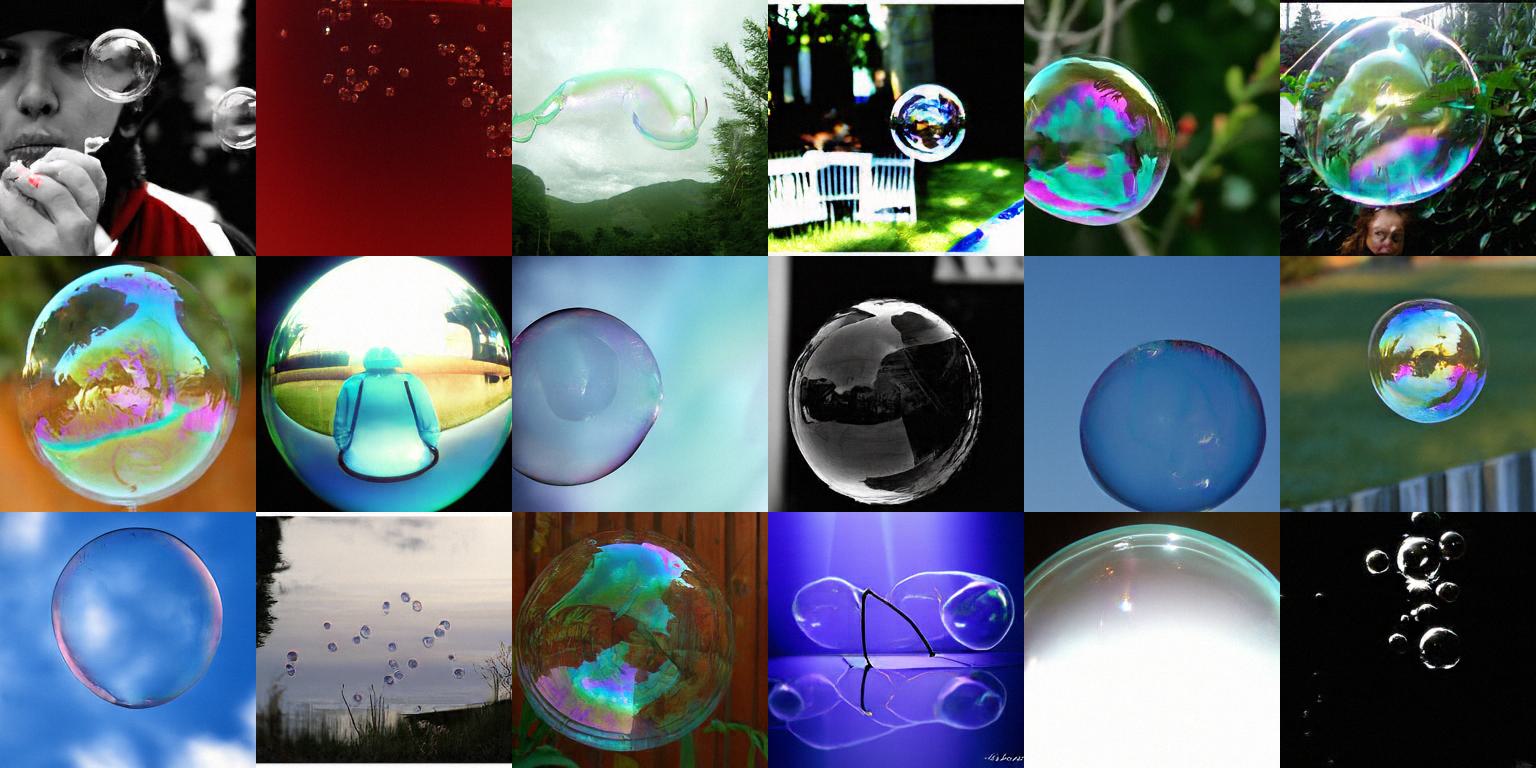}{n09229709}
  \qualrow{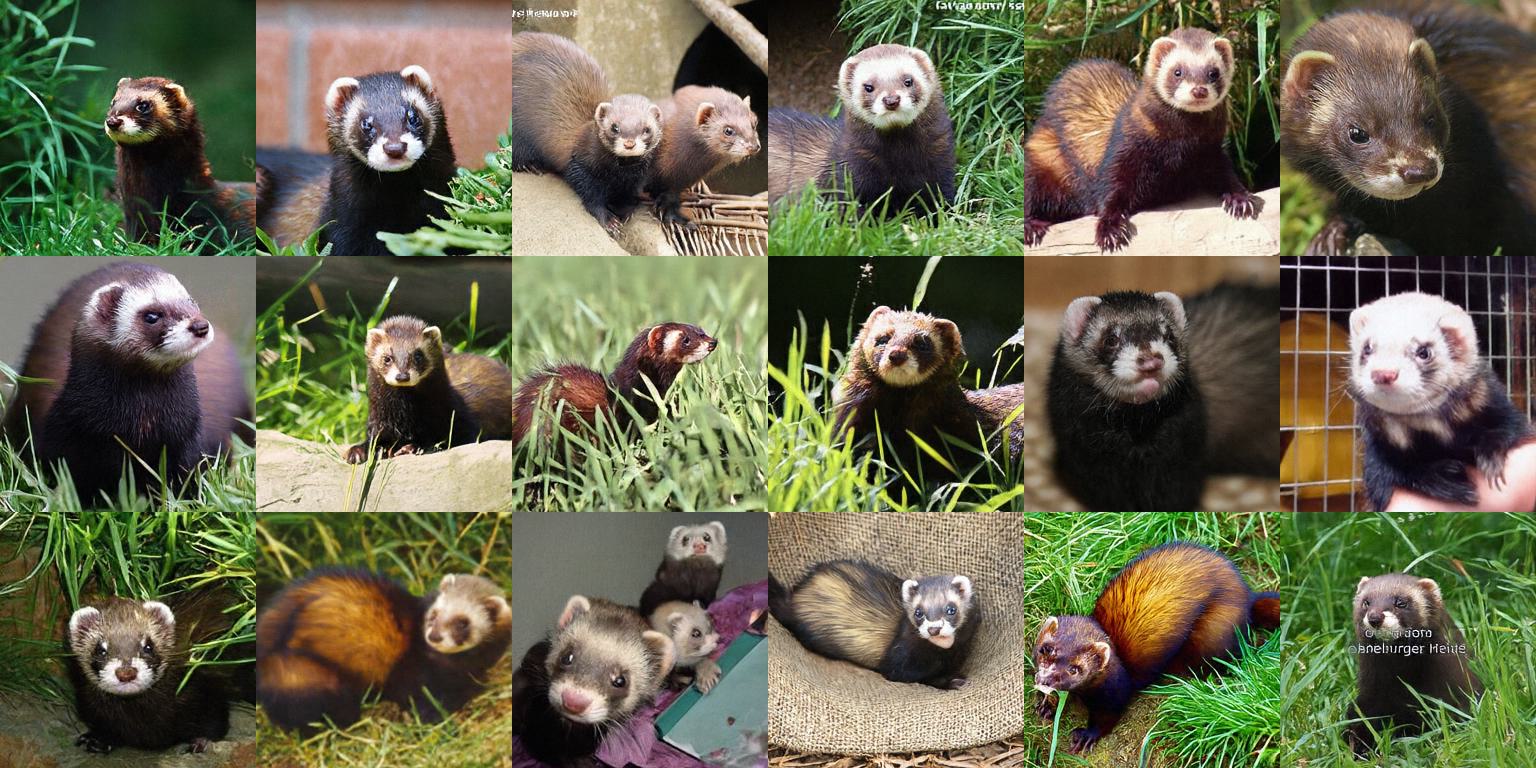}{n02443114}{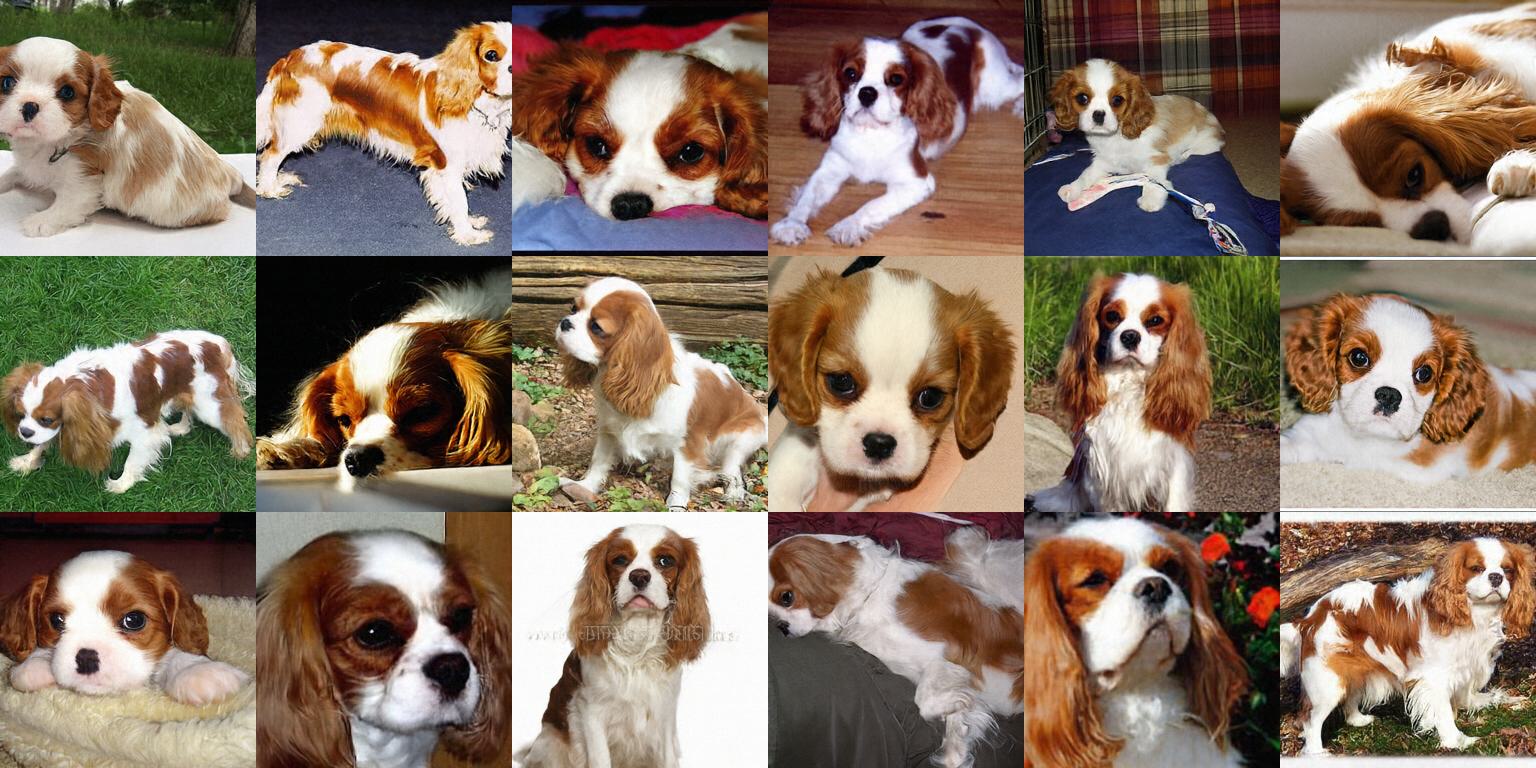}{n02086646}
  \qualrow{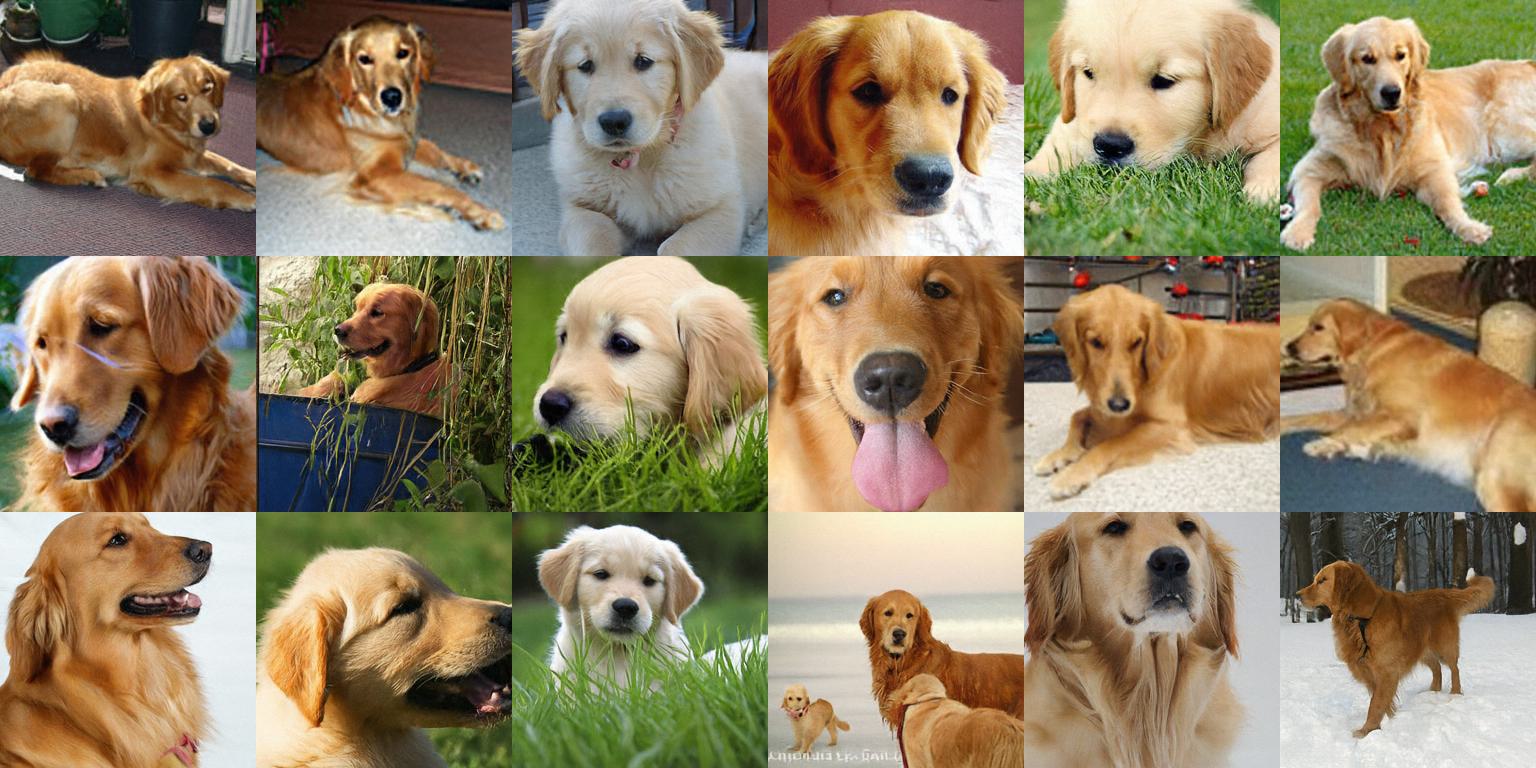}{n02099601}{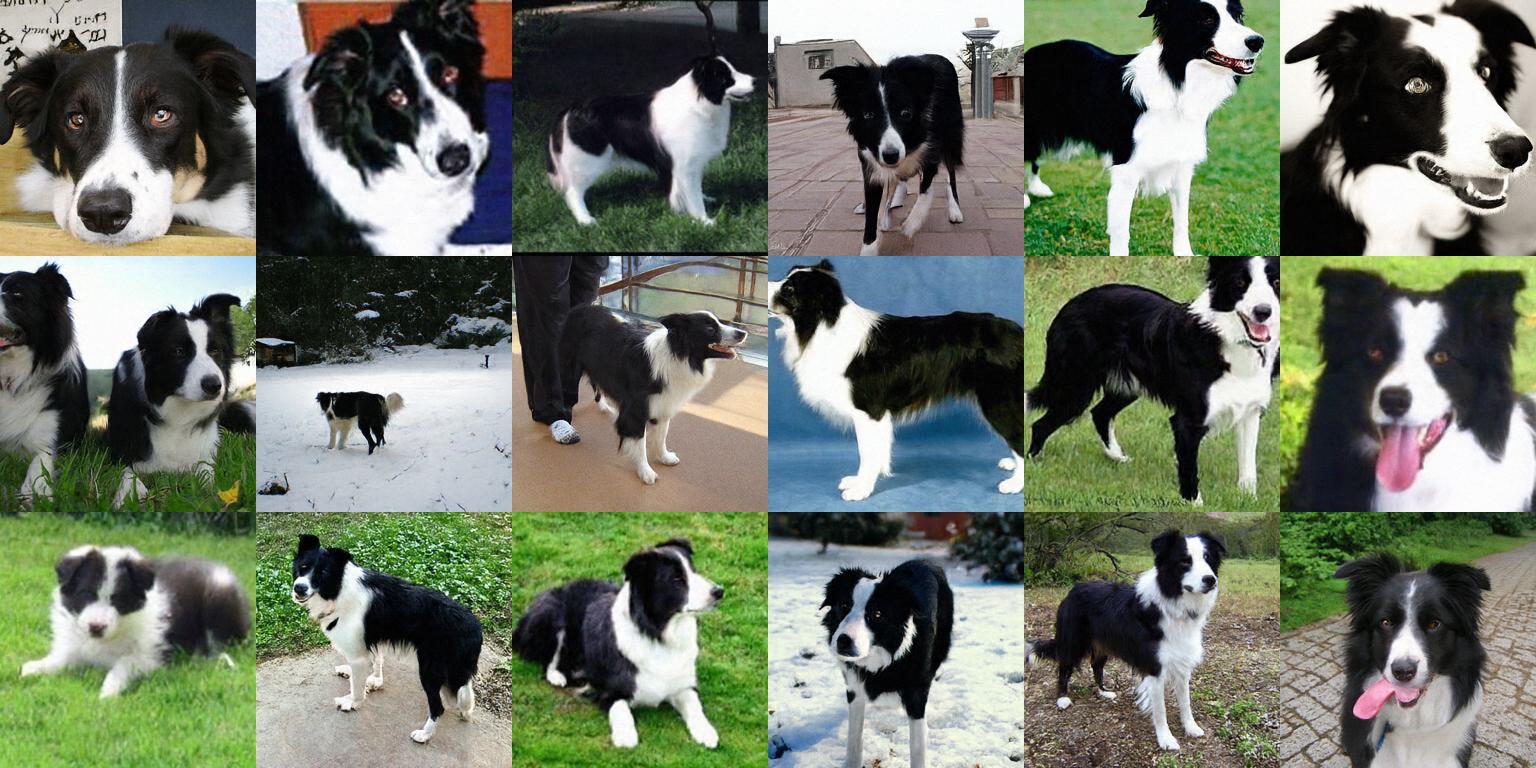}{n02106166}
  \qualrow{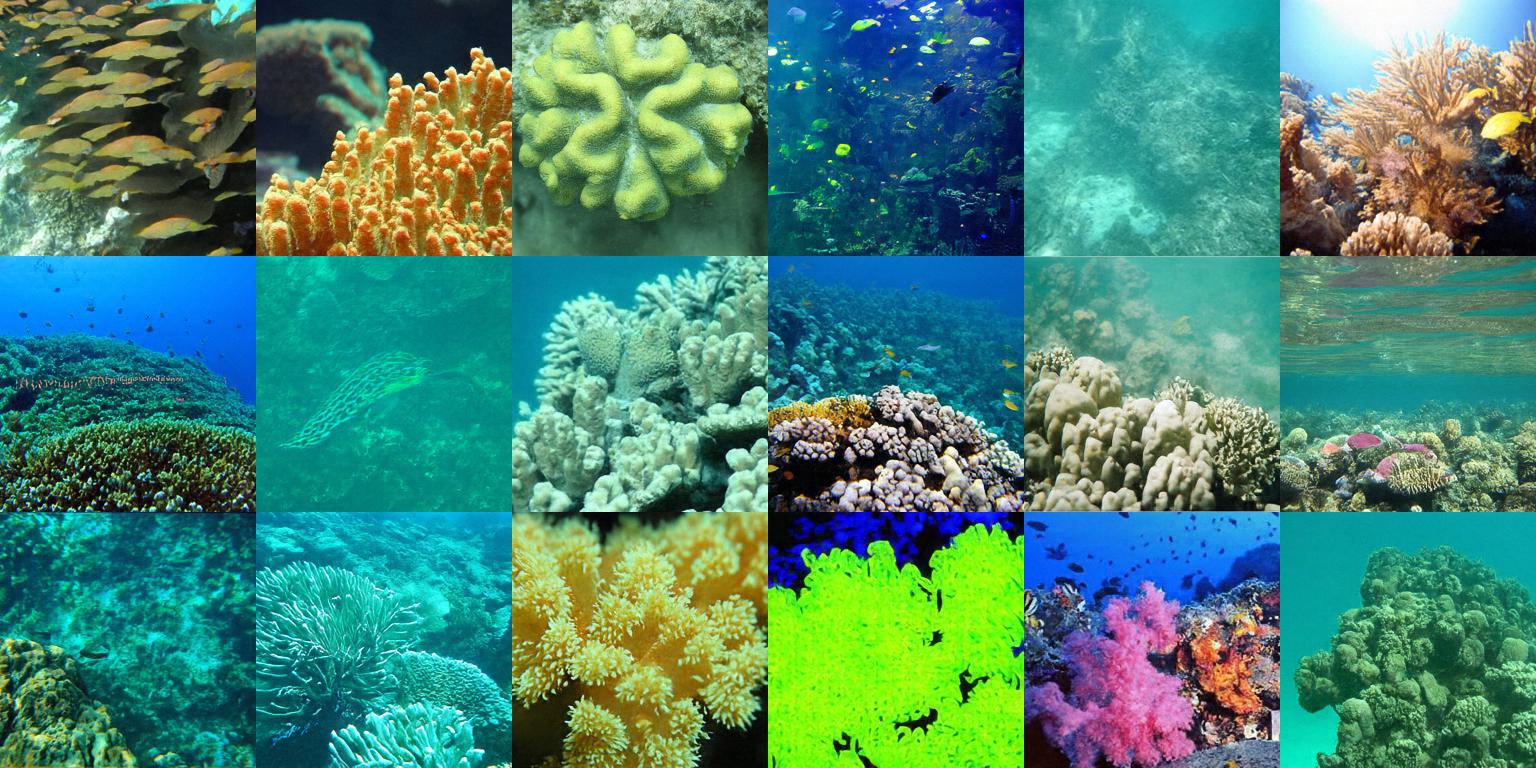}{n09256479}{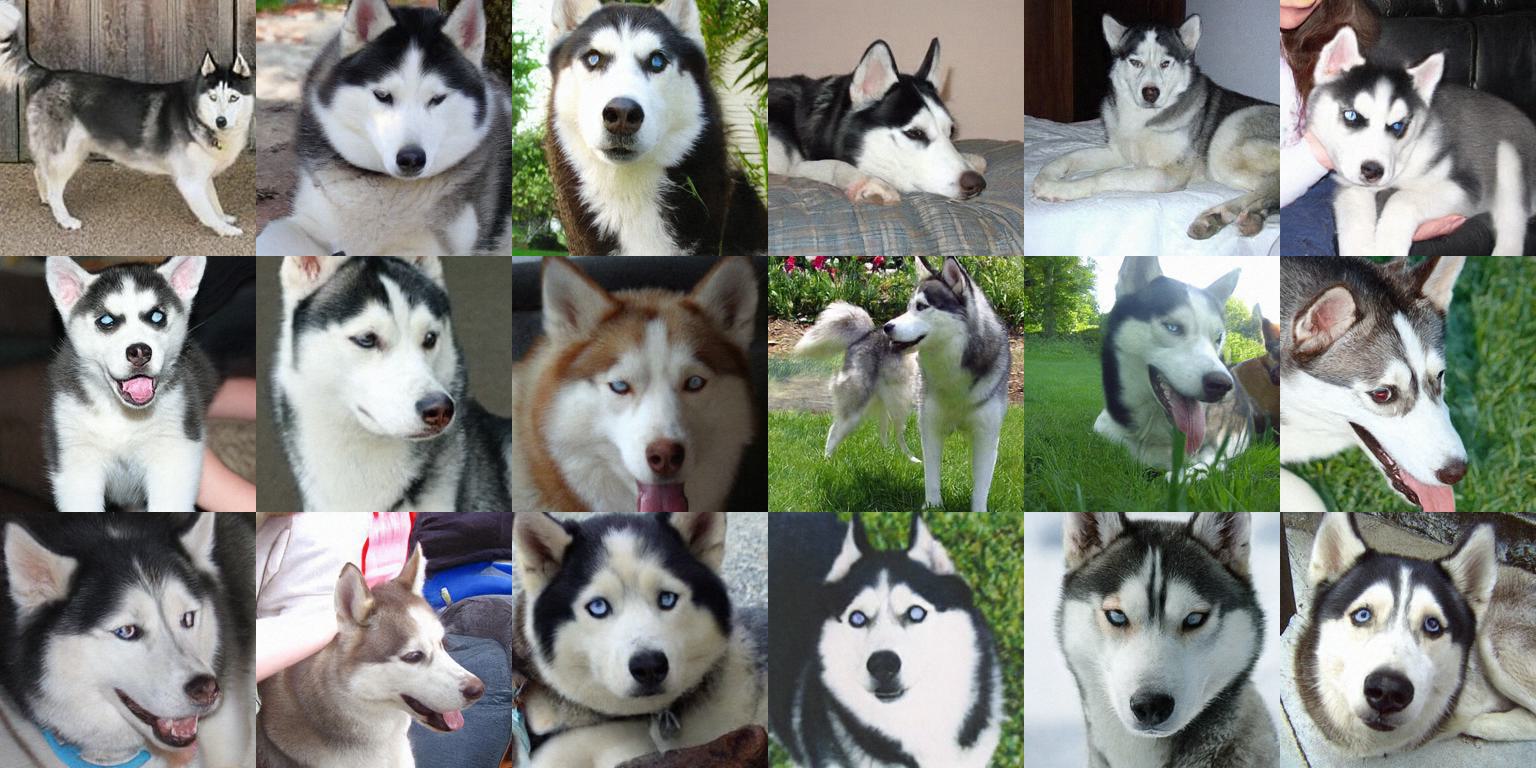}{n02110185}
  \qualcenter{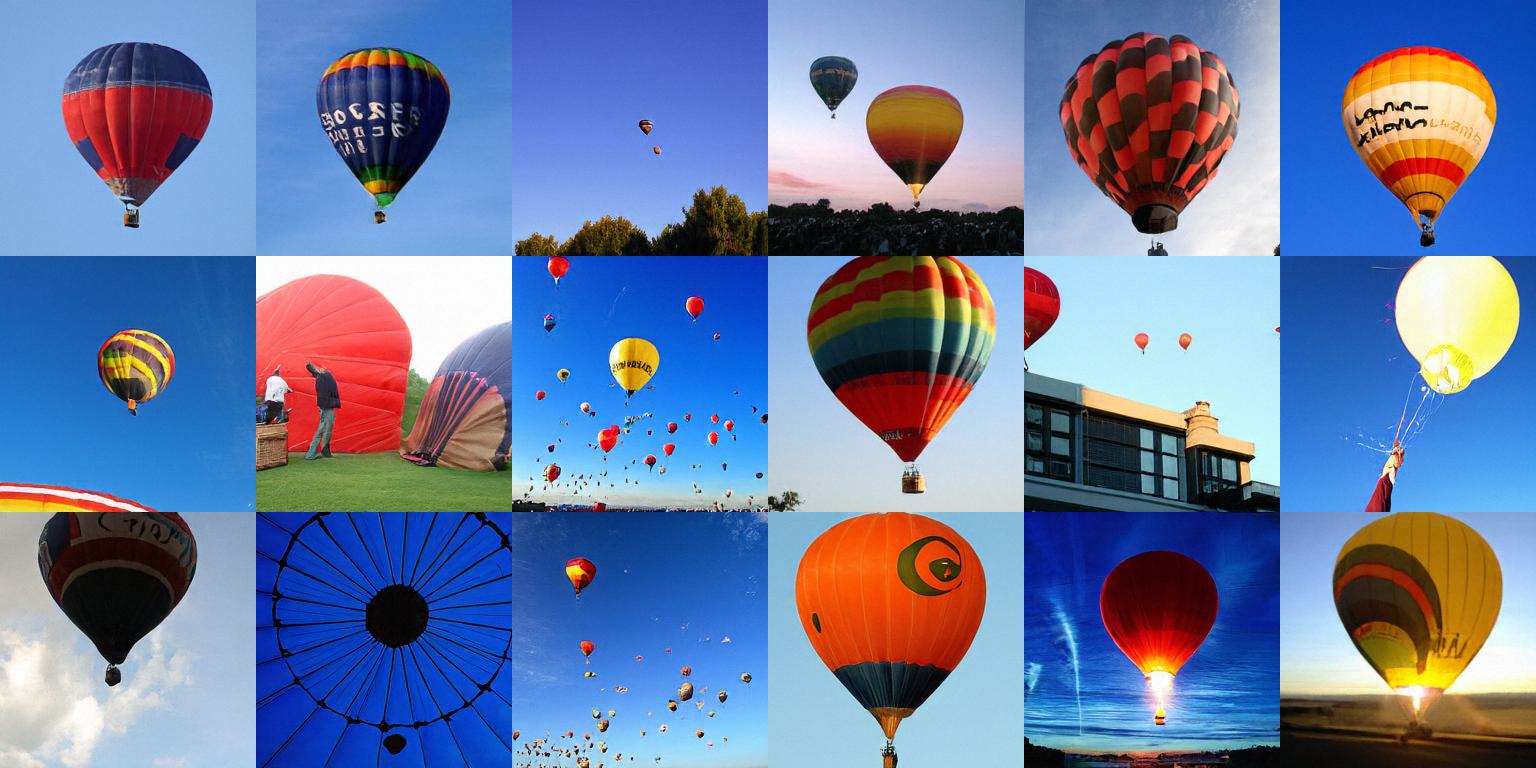}{n02782093}
  \caption{Uncurated samples from JiT-H/16 + \ours{} on ImageNet $256\times256$.}
  \label{fig:app-qualitative-2}
\end{figure}

\begin{figure}[H]
  \centering
  \qualrow{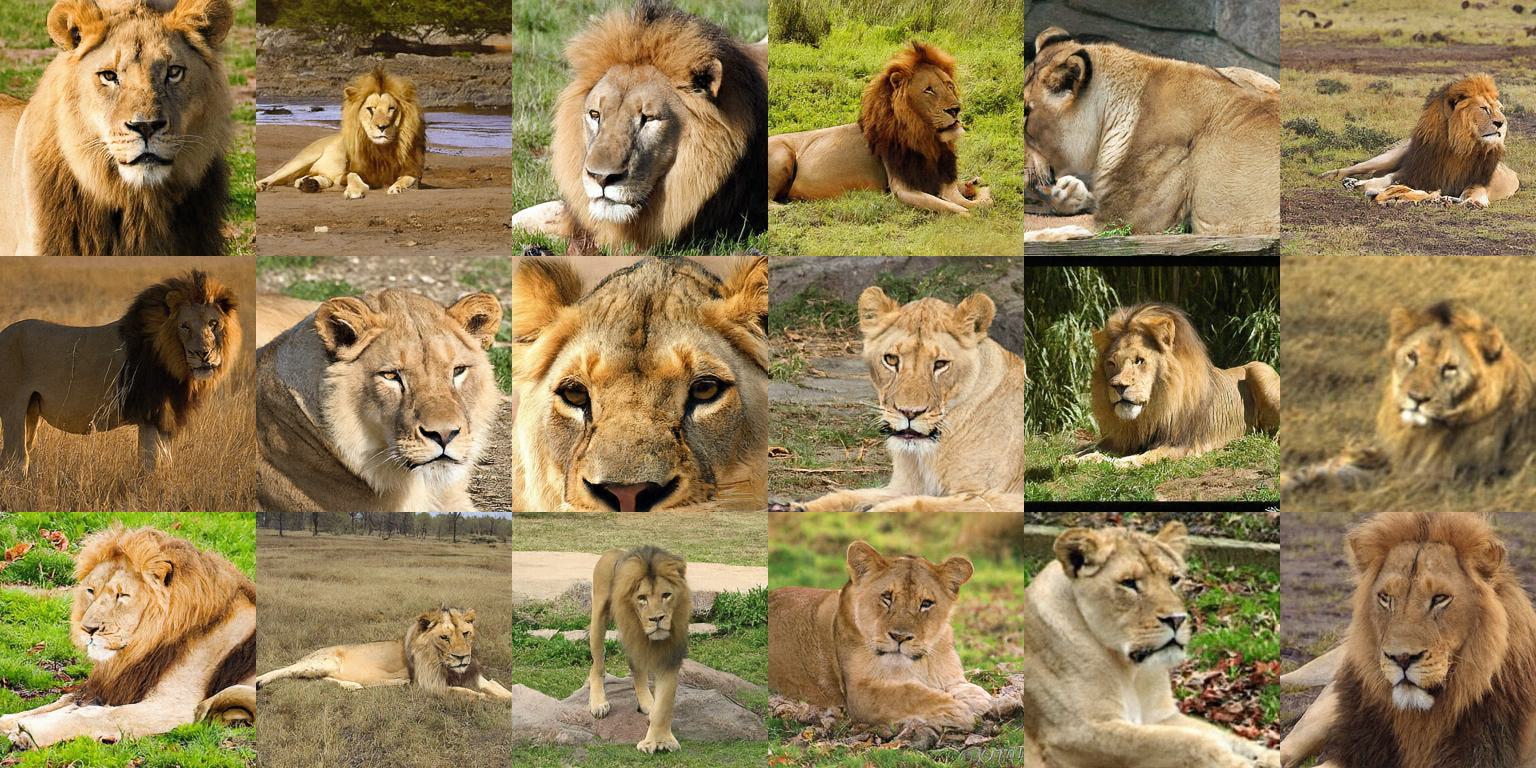}{n02129165}{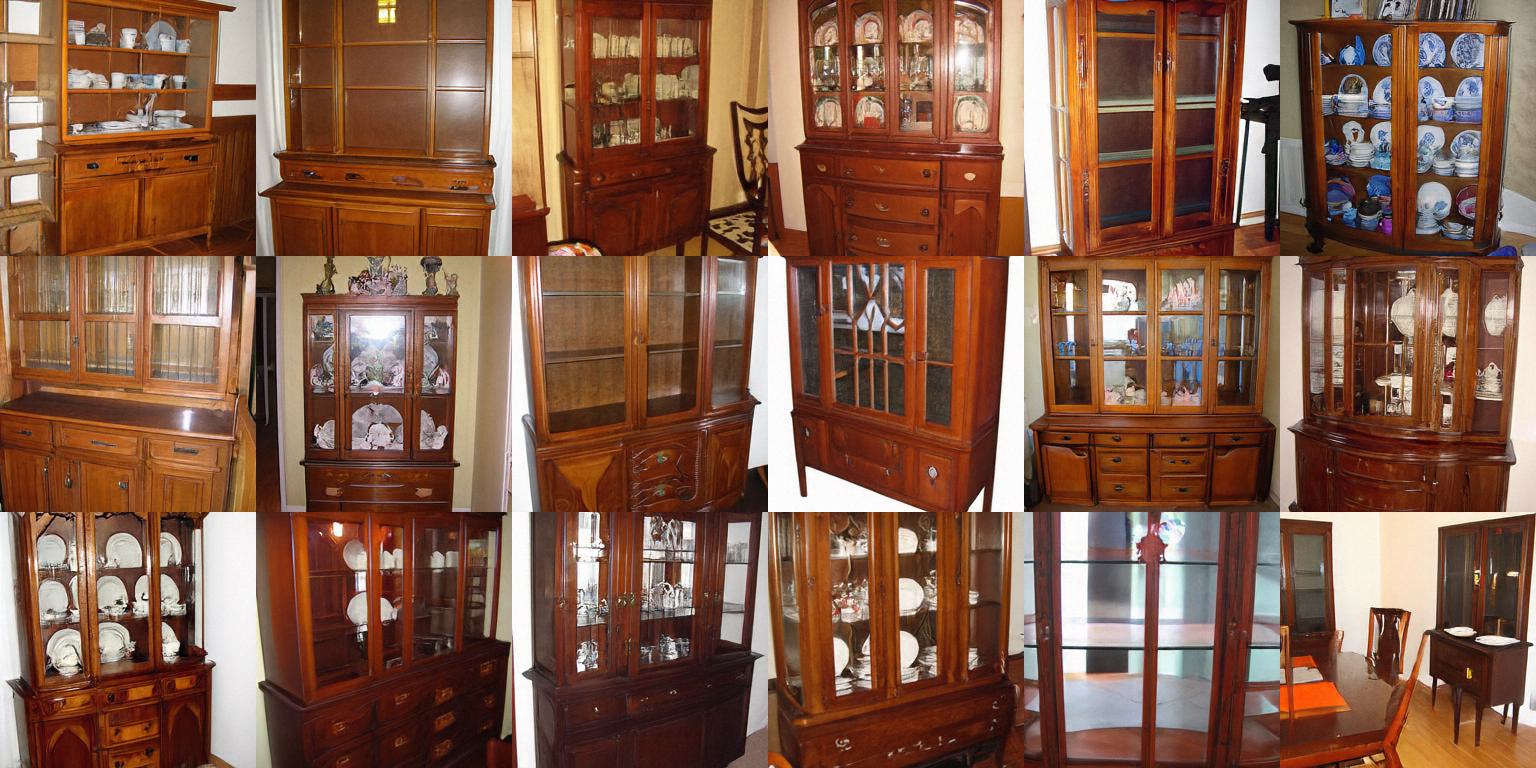}{n03018349}
  \qualrow{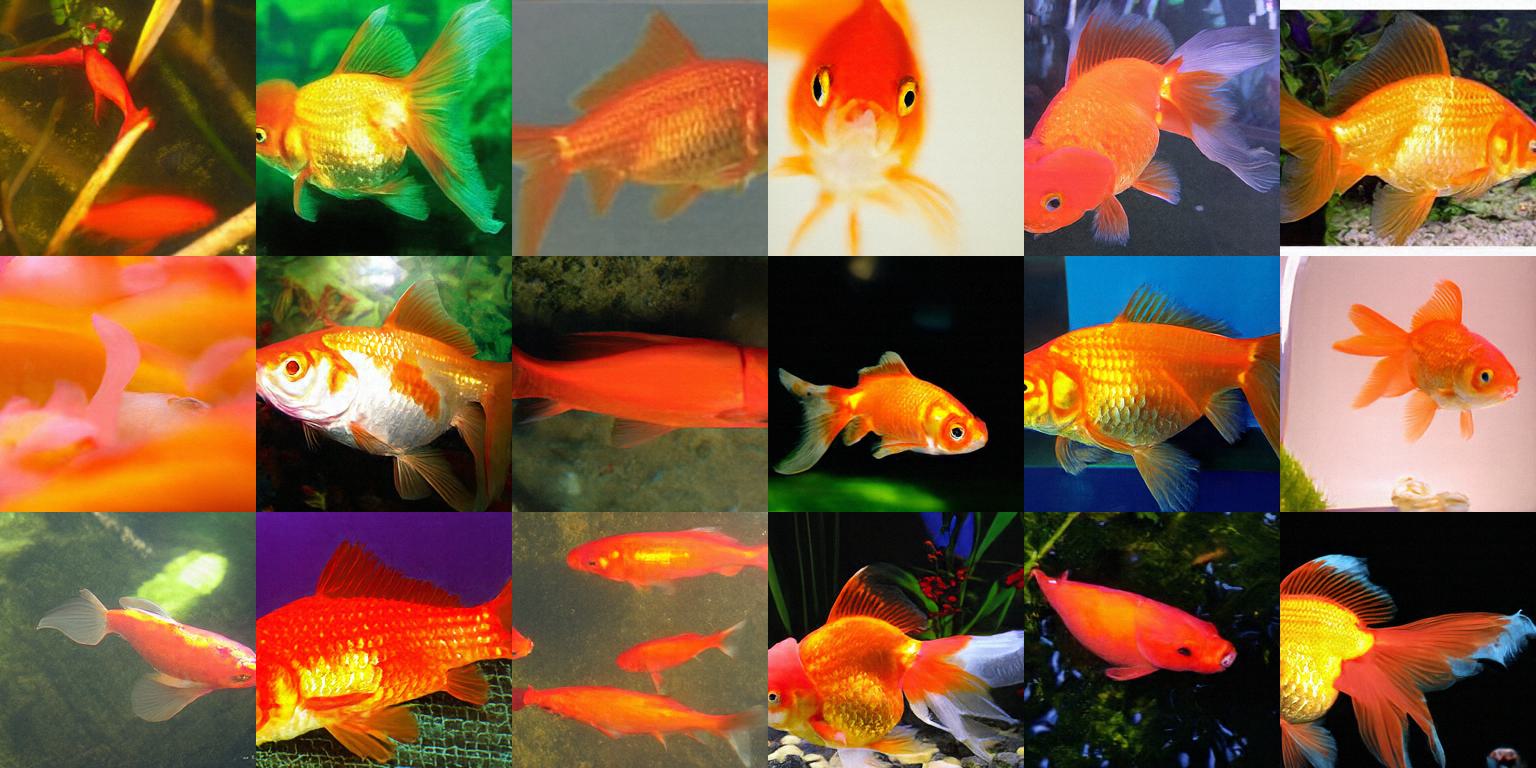}{n01443537}{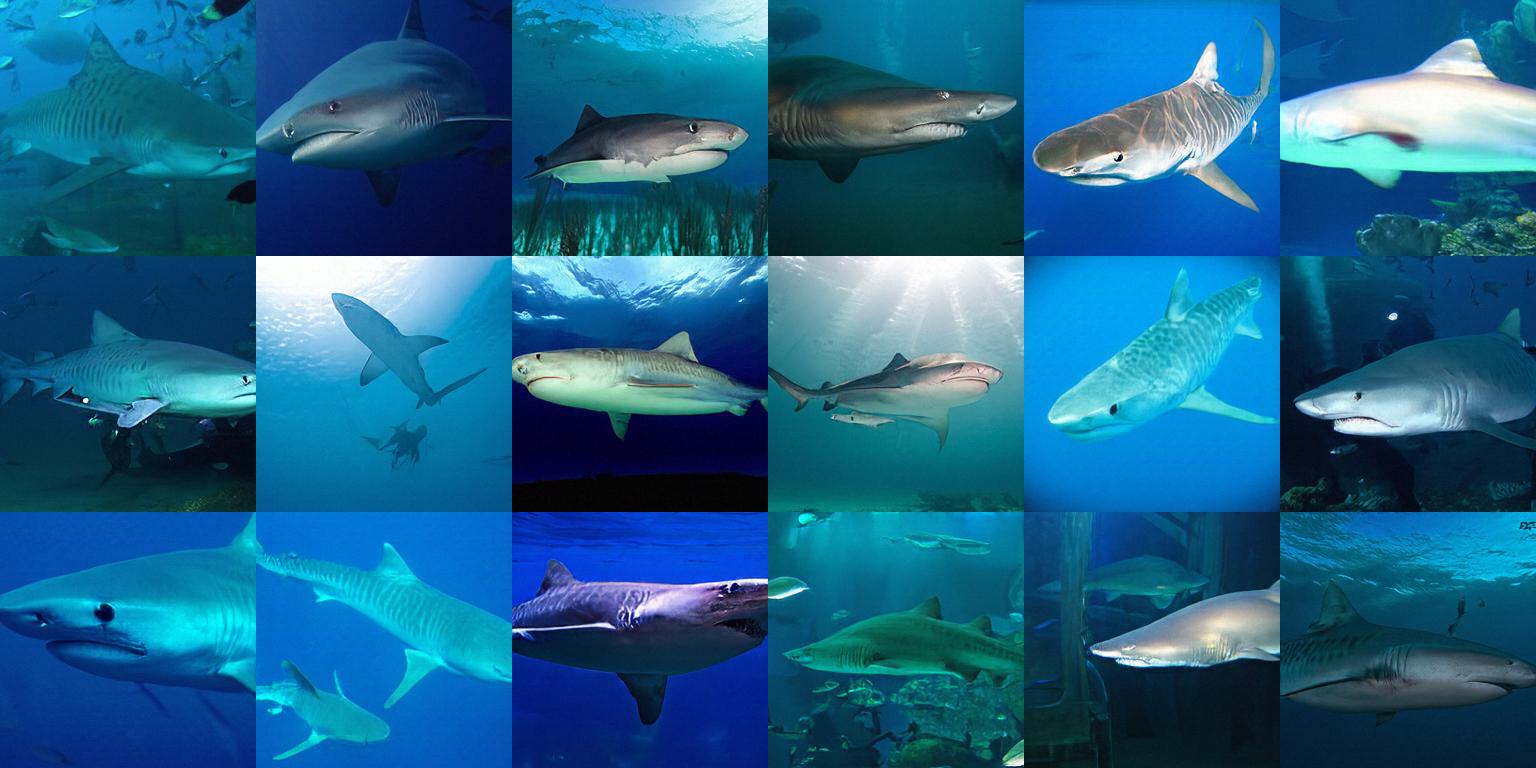}{n01491361}
  \qualrow{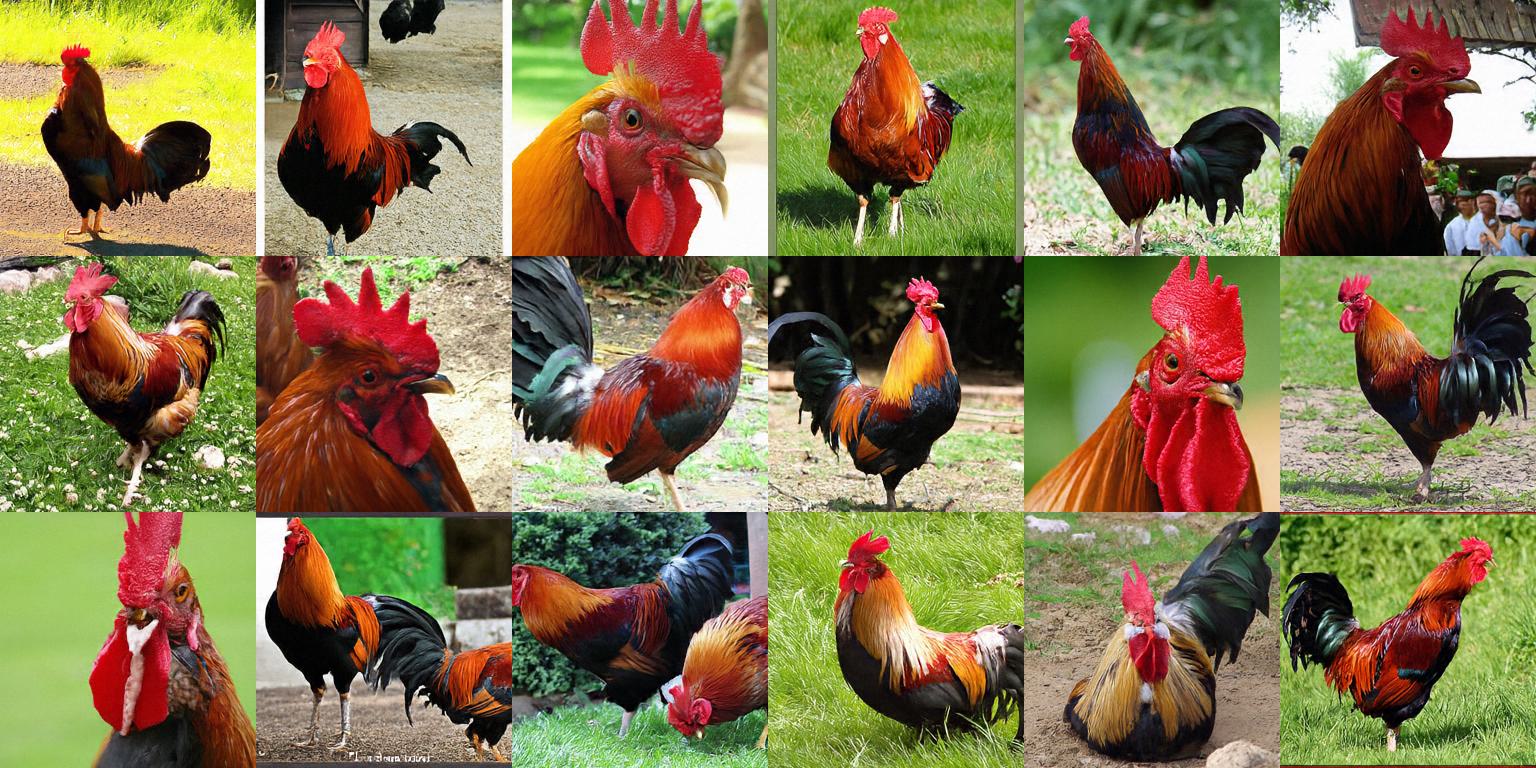}{n01514668}{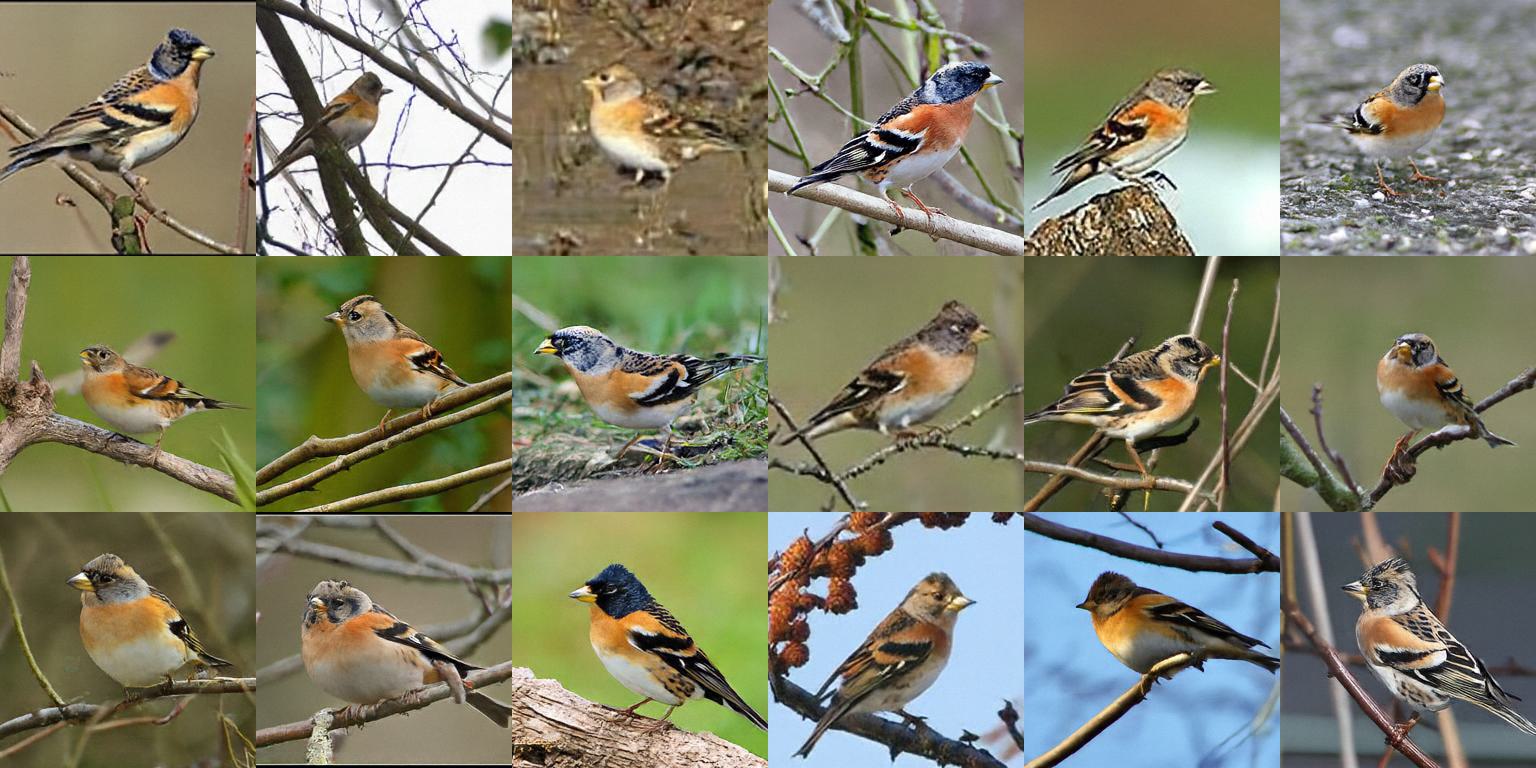}{n01530575}
  \qualrow{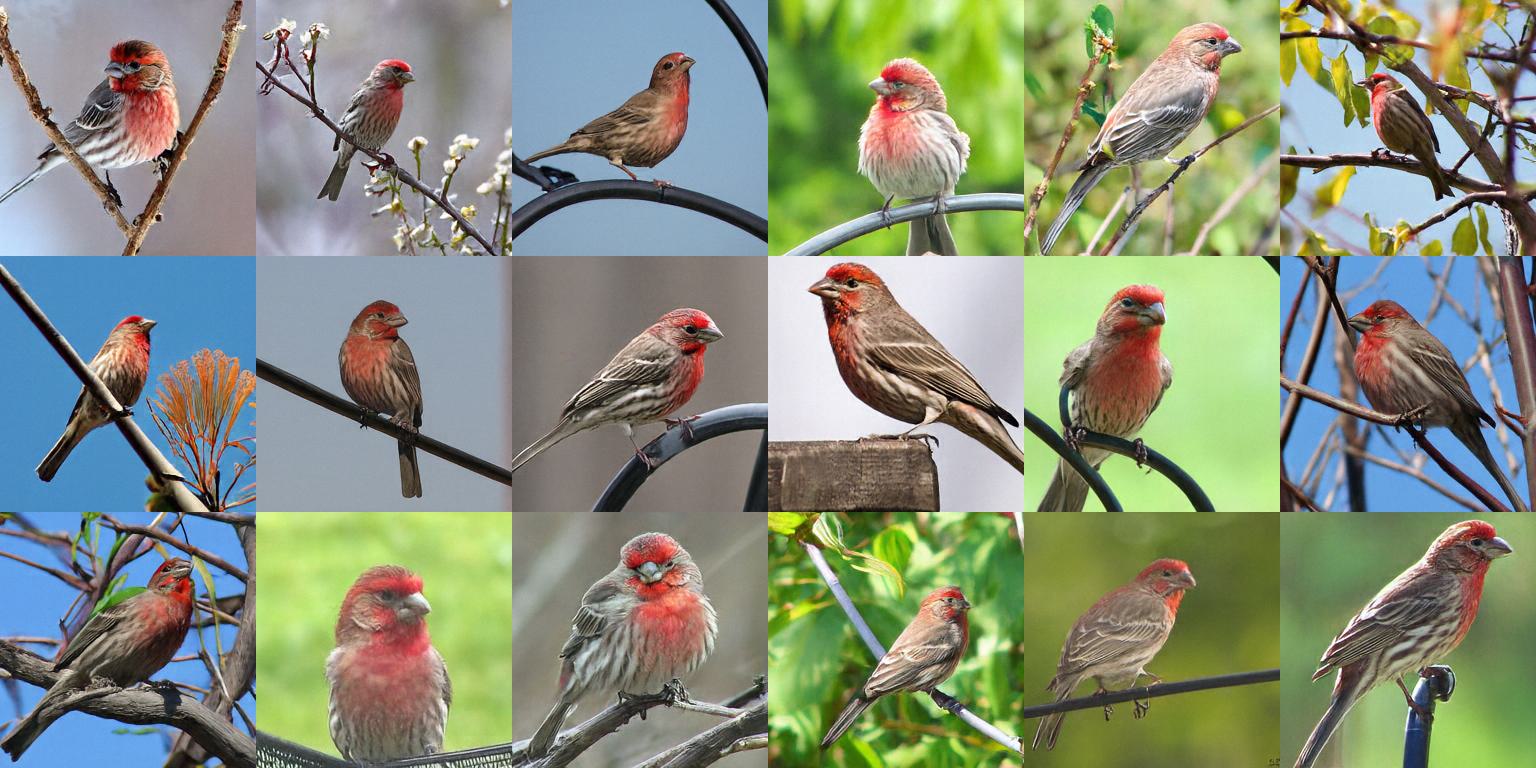}{n01532829}{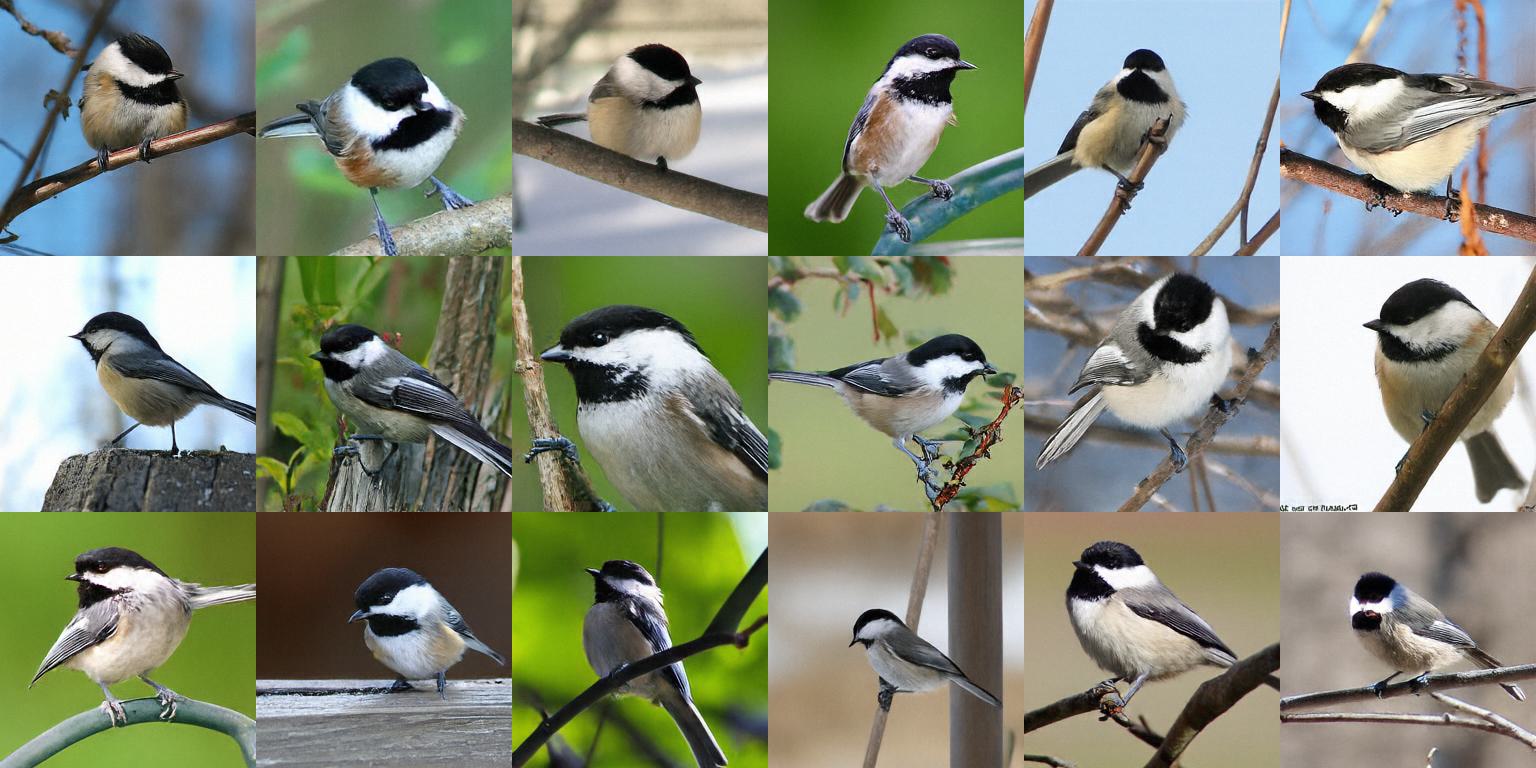}{n01592084}
  \qualcenter{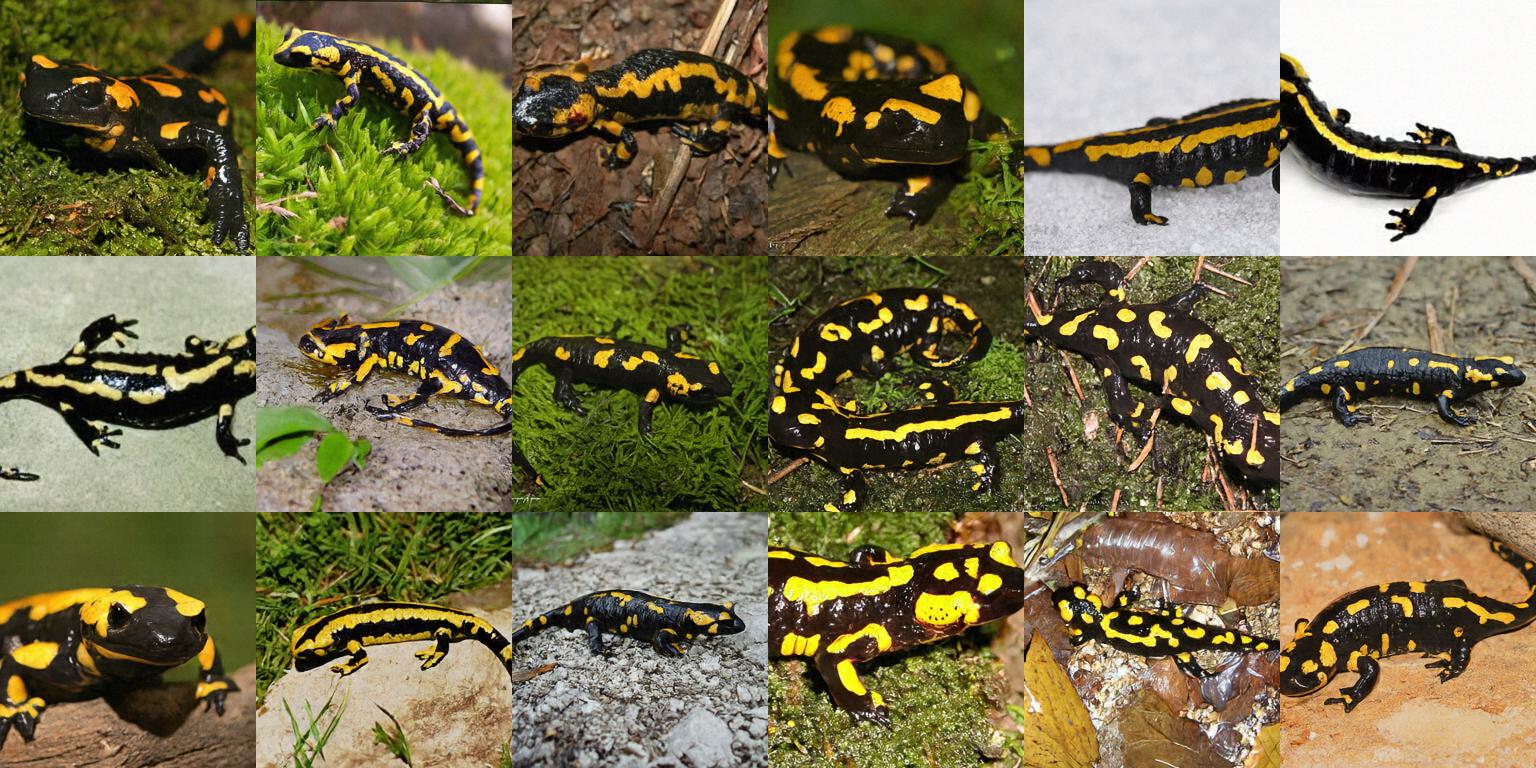}{n01629819}
  \caption{Uncurated samples from JiT-H/16 + \ours{} on ImageNet $256\times256$.}
  \label{fig:app-qualitative-3}
\end{figure}

\begin{figure}[H]
  \centering
  \qualrow{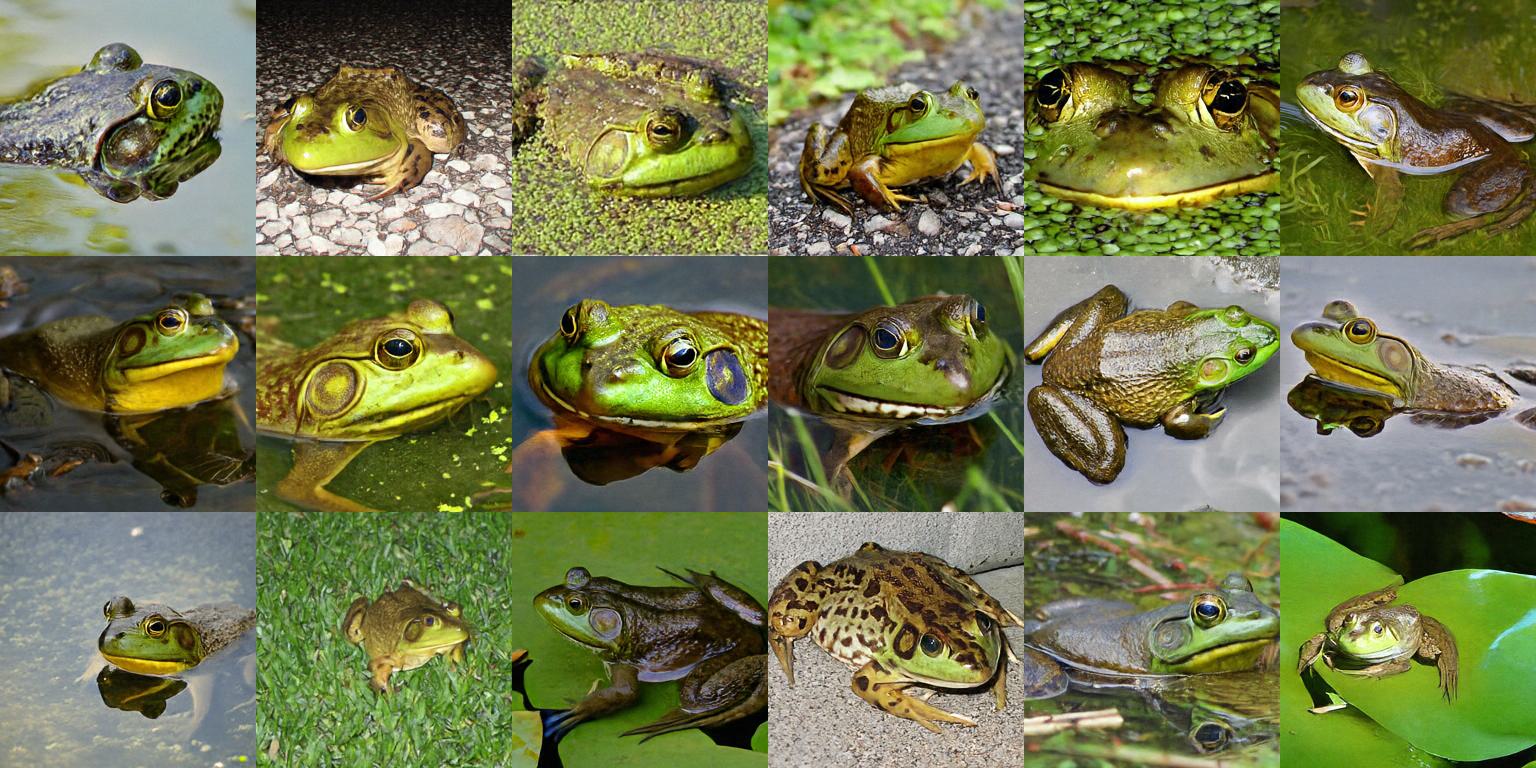}{n01641577}{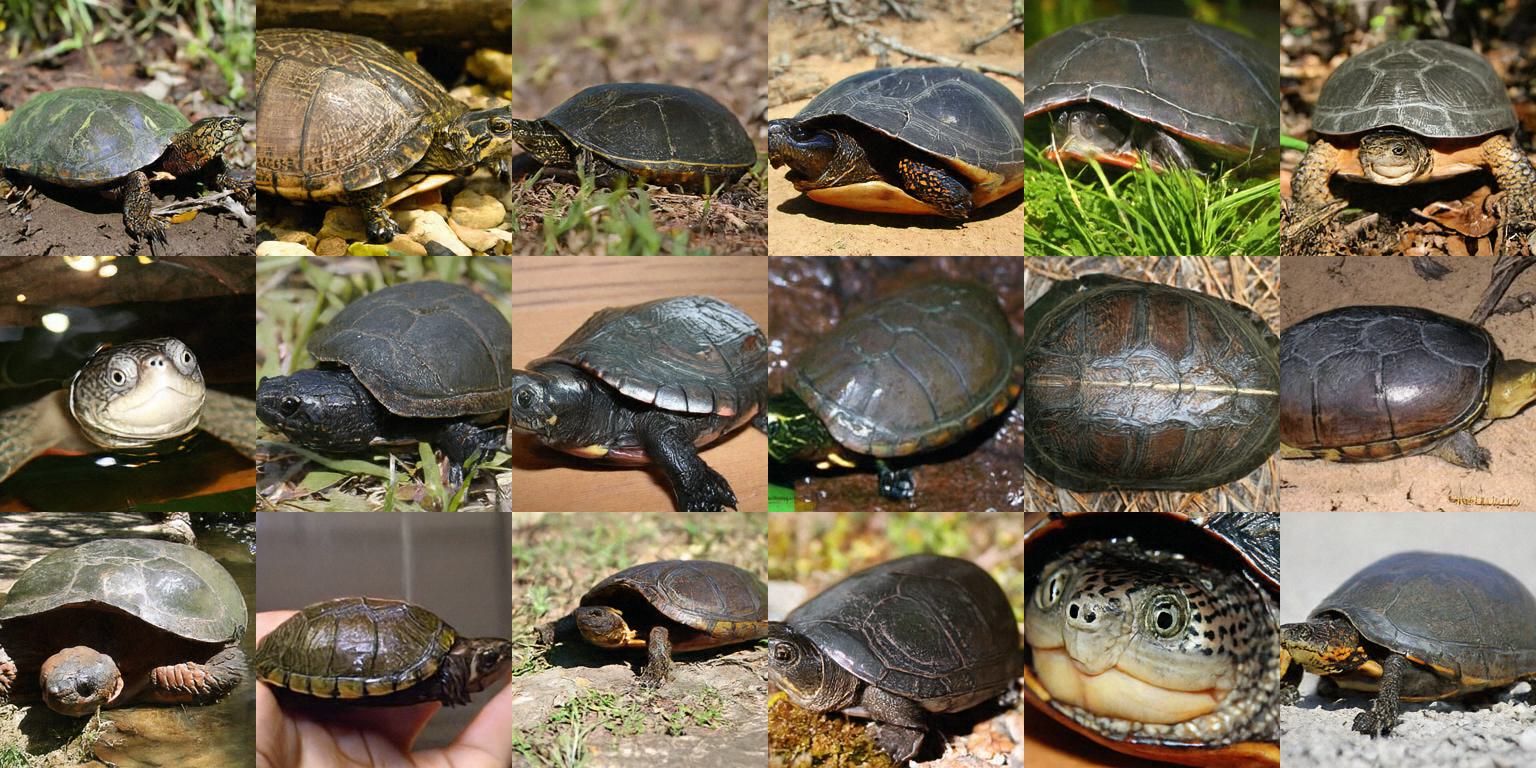}{n01667114}
  \qualrow{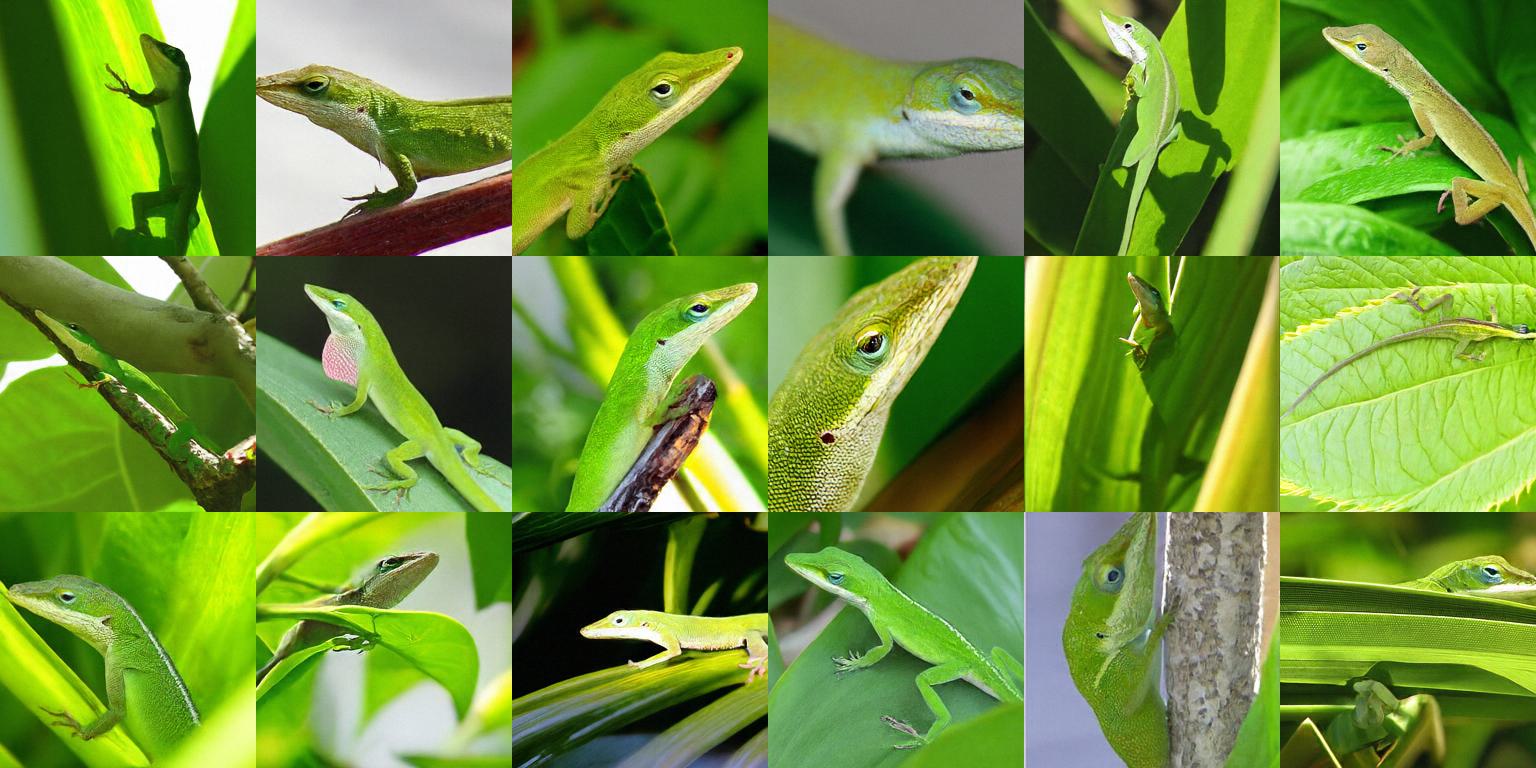}{n01682714}{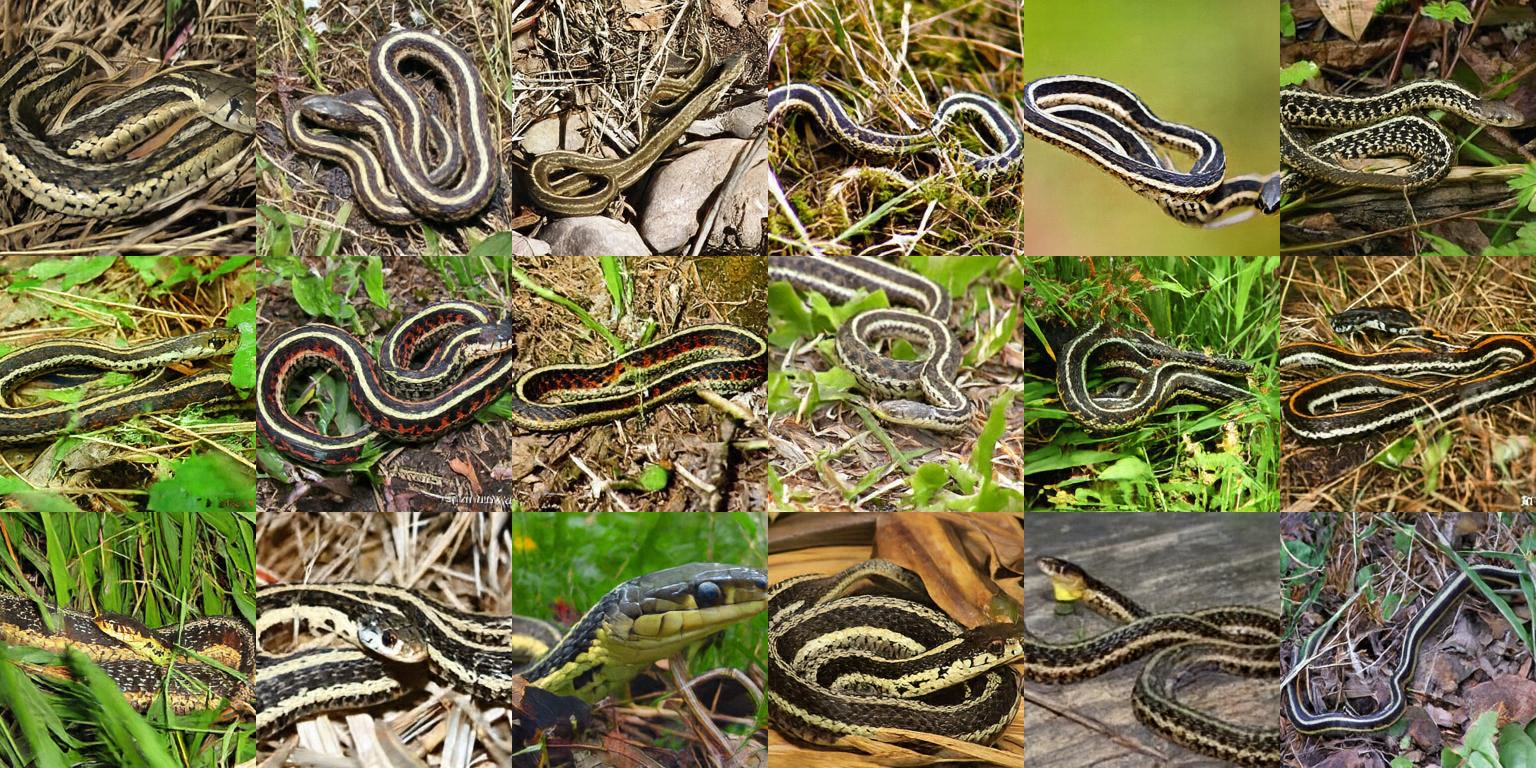}{n01735189}
  \qualrow{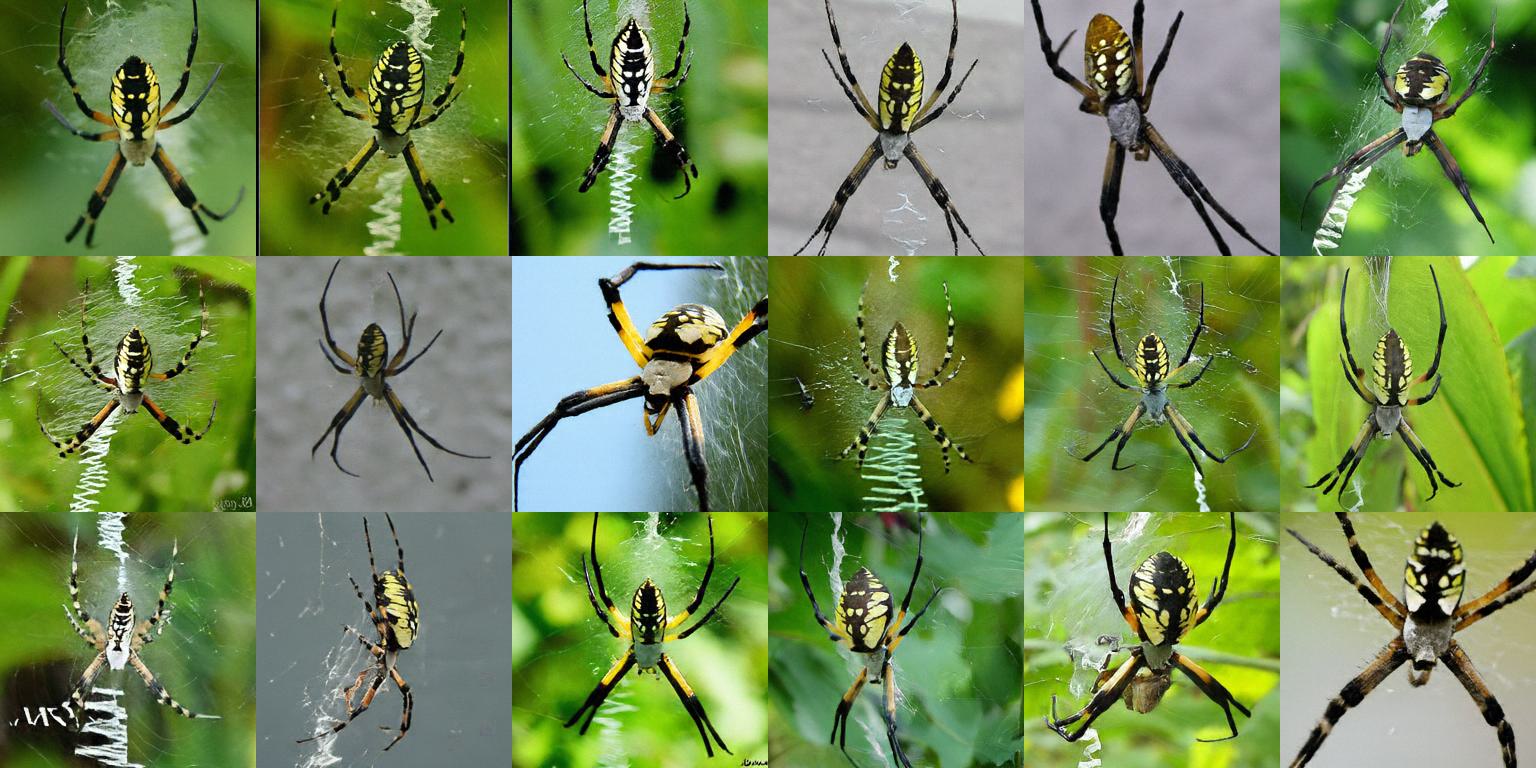}{n01773157}{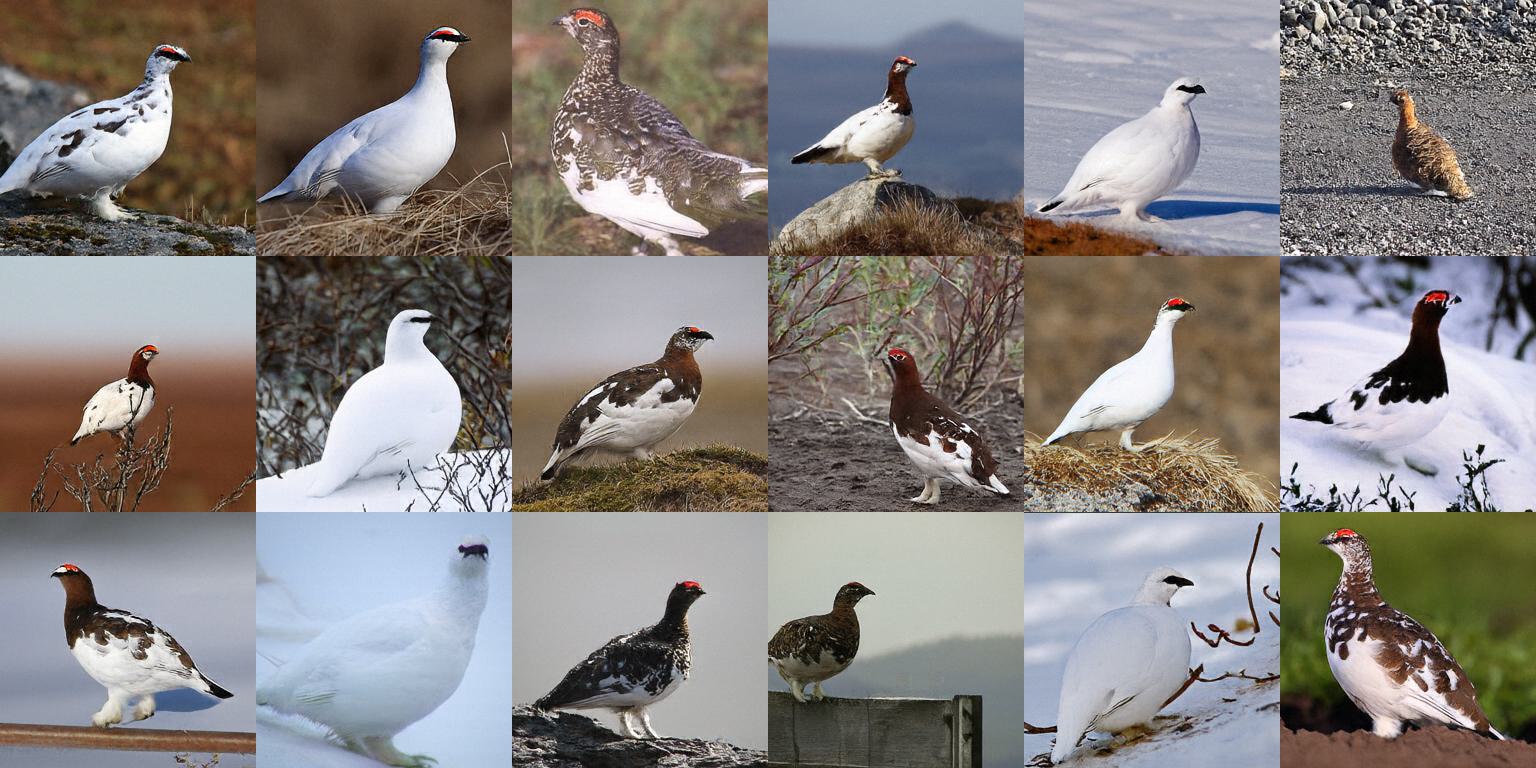}{n01796340}
  \qualrow{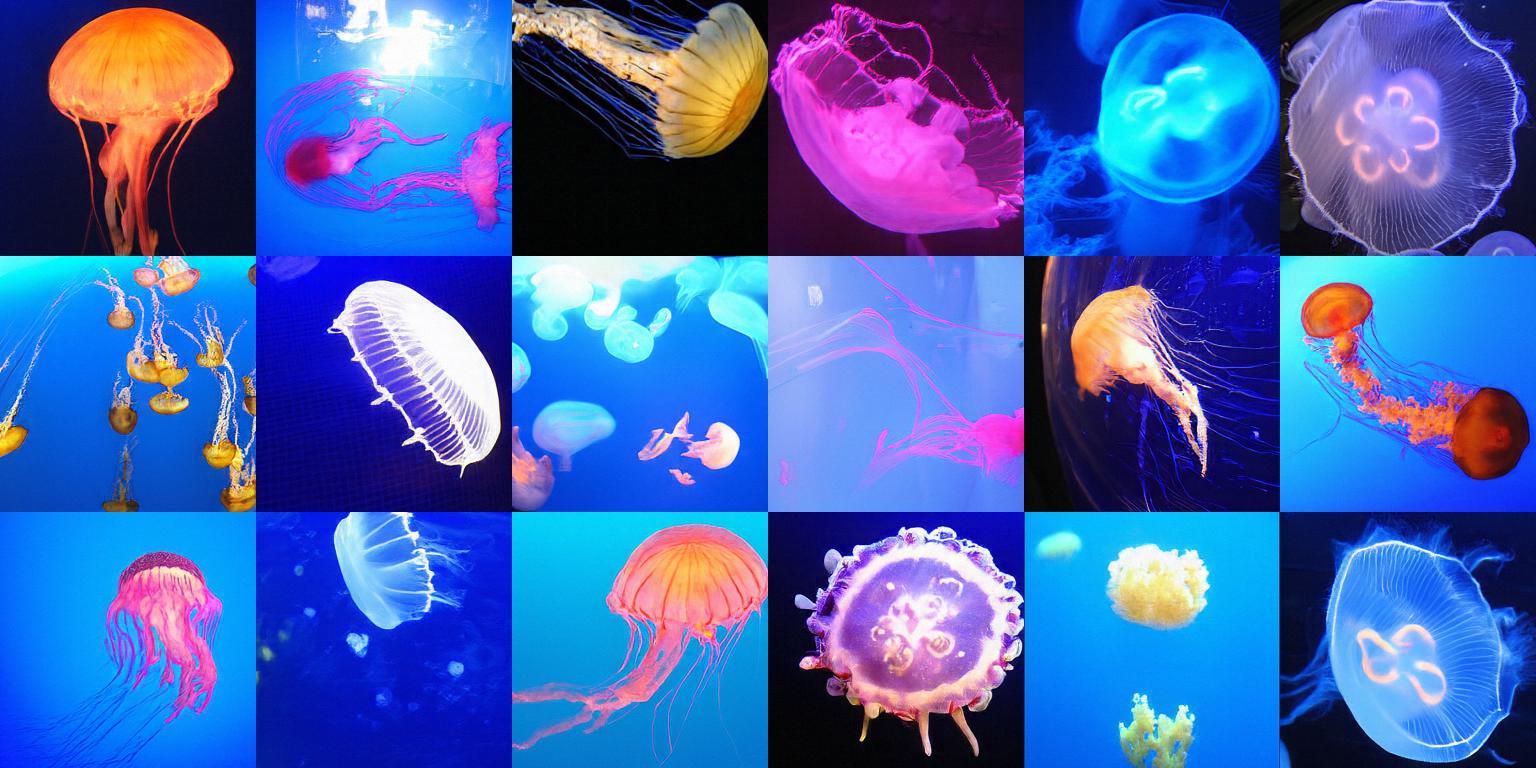}{n01910747}{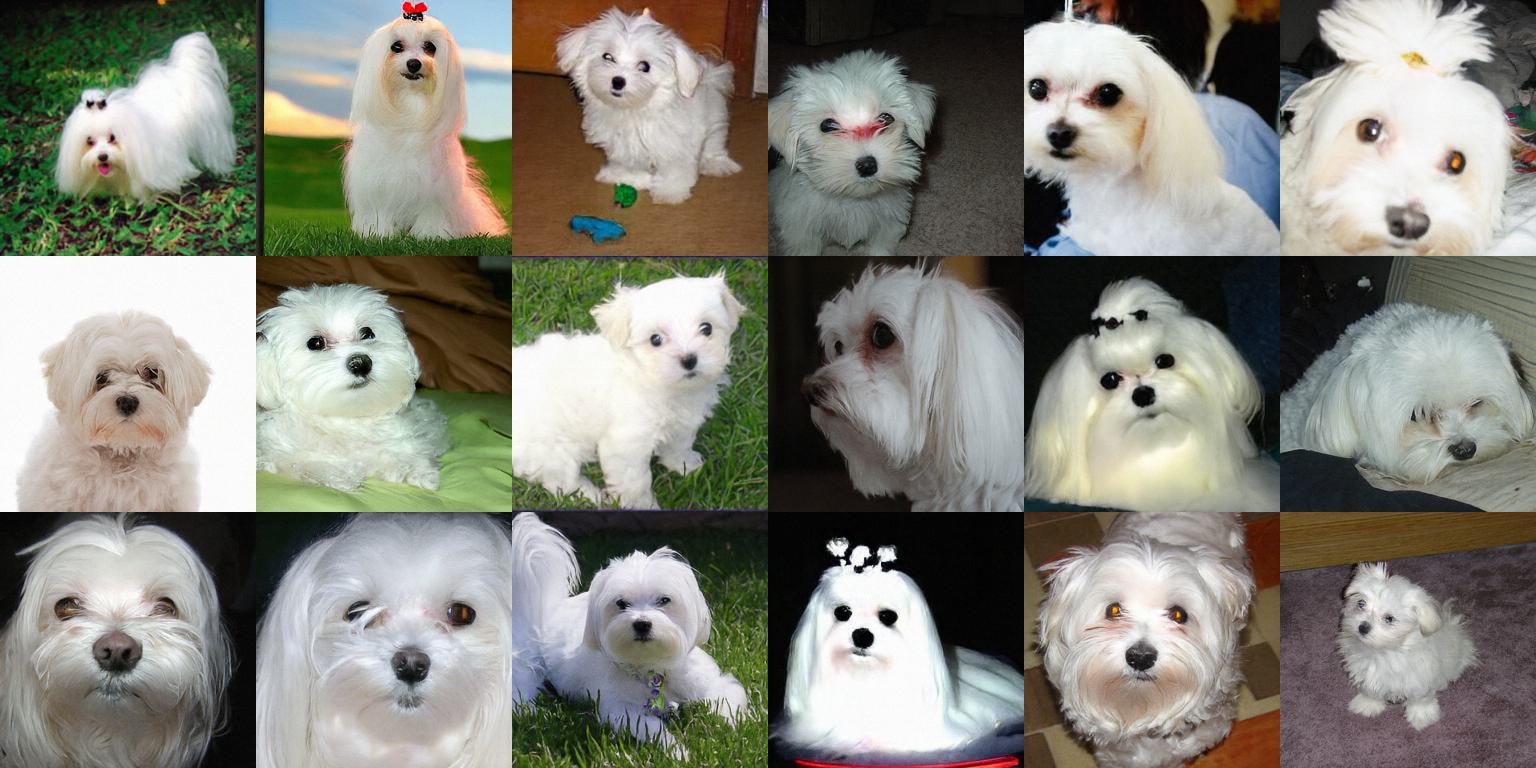}{n02085936}
  \qualcenter{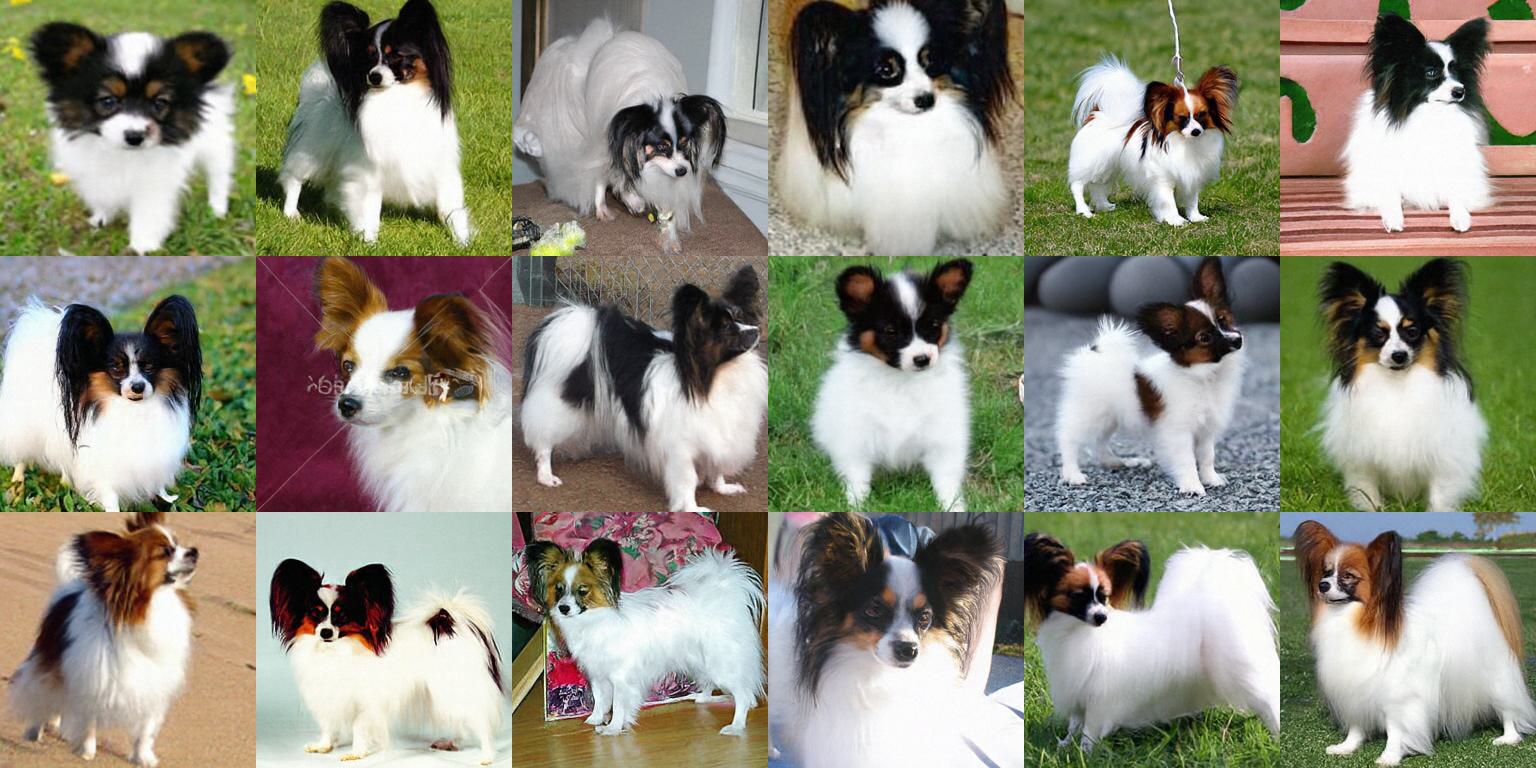}{n02086910}
  \caption{Uncurated samples from JiT-H/16 + \ours{} on ImageNet $256\times256$.}
  \label{fig:app-qualitative-4}
\end{figure}

\begin{figure}[H]
  \centering
  \qualrow{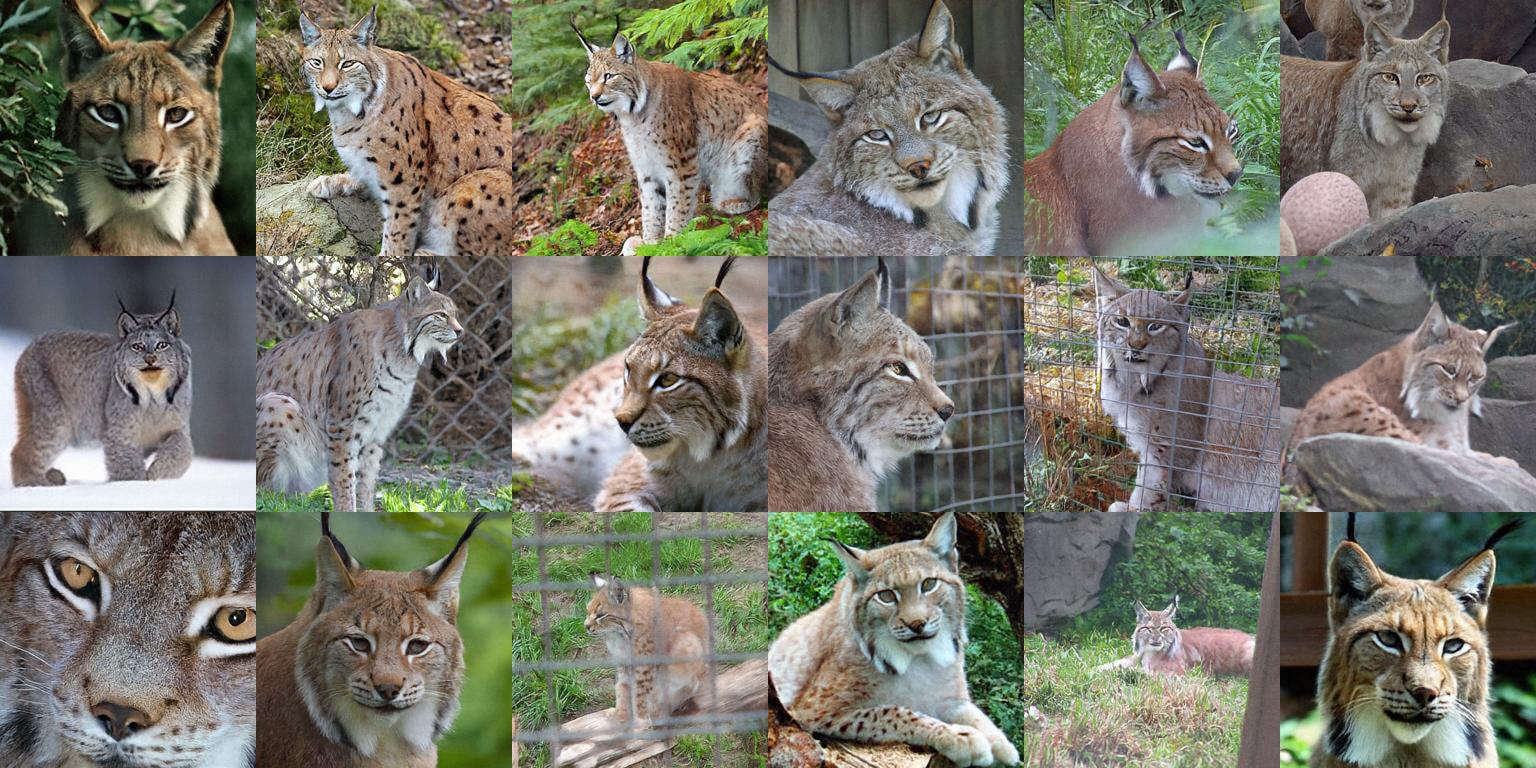}{n02127052}{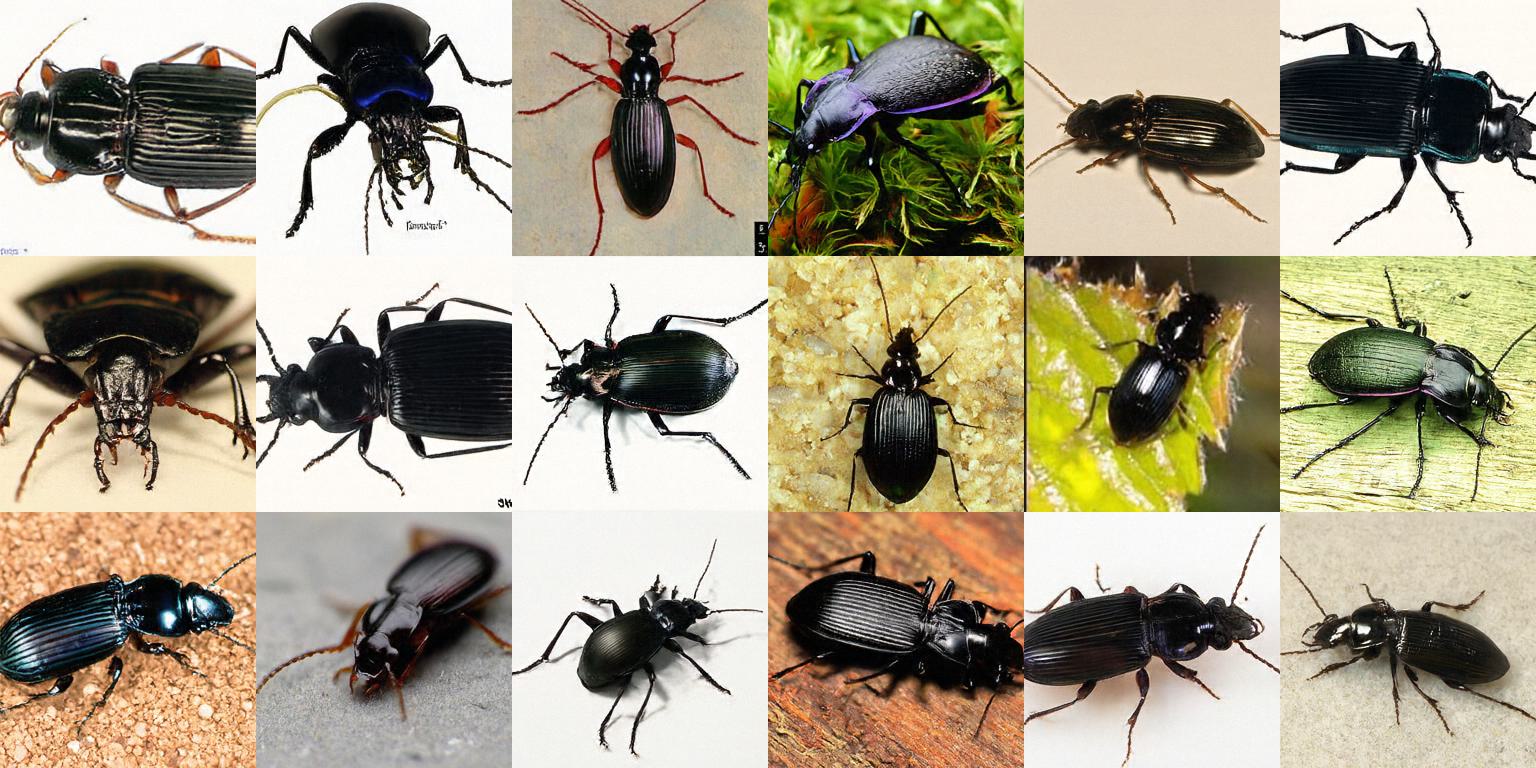}{n02167151}
  \qualrow{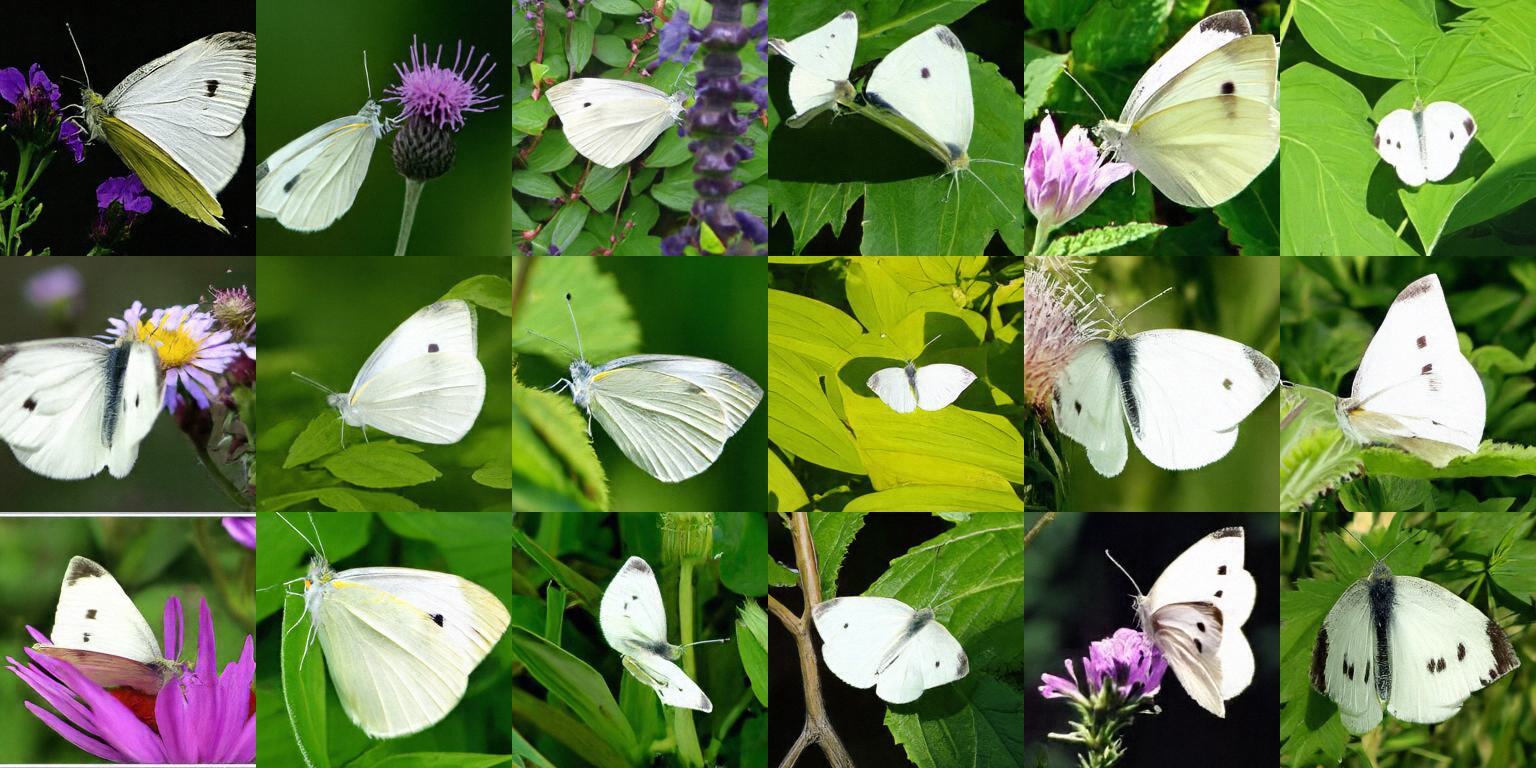}{n02280649}{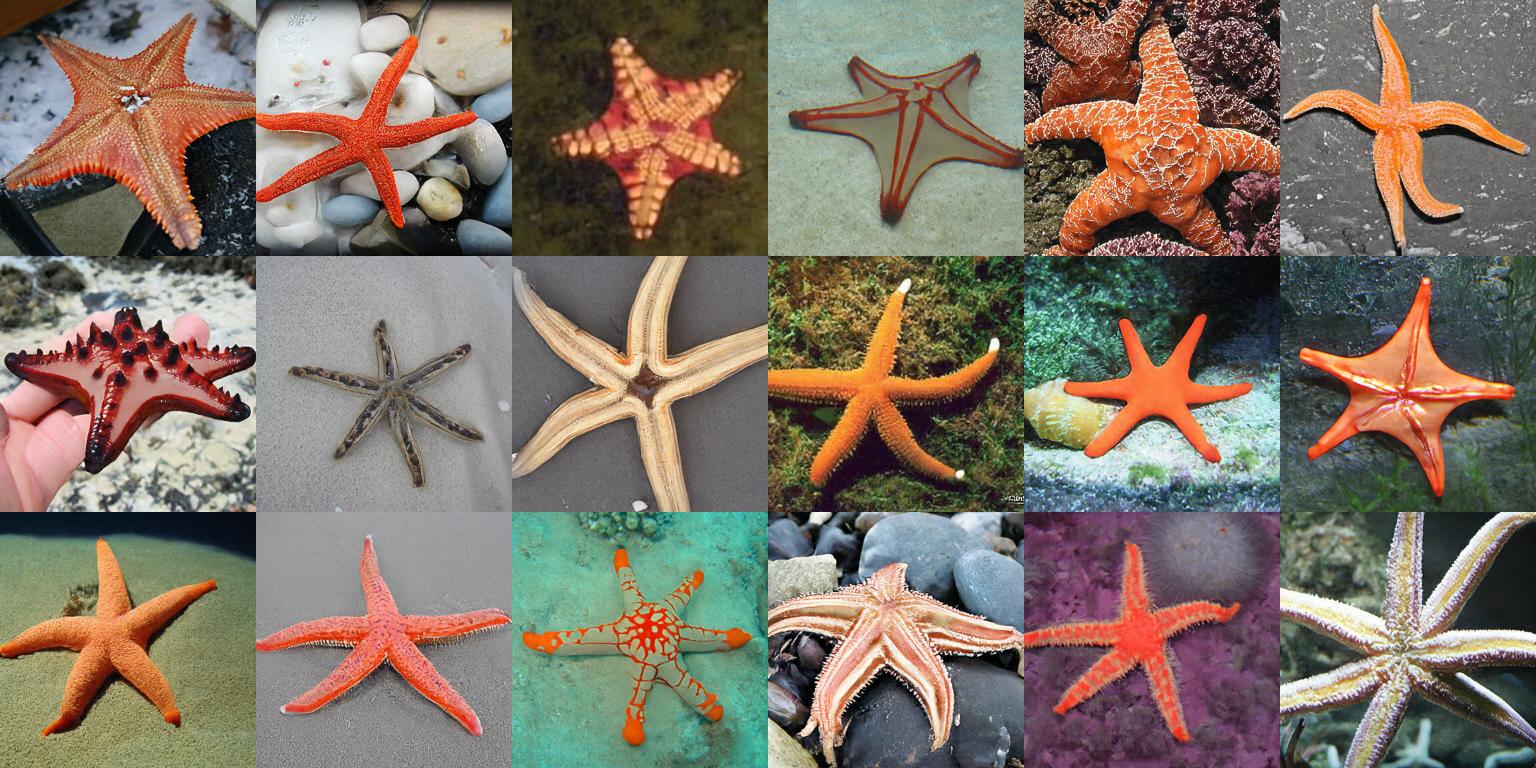}{n02317335}
  \qualrow{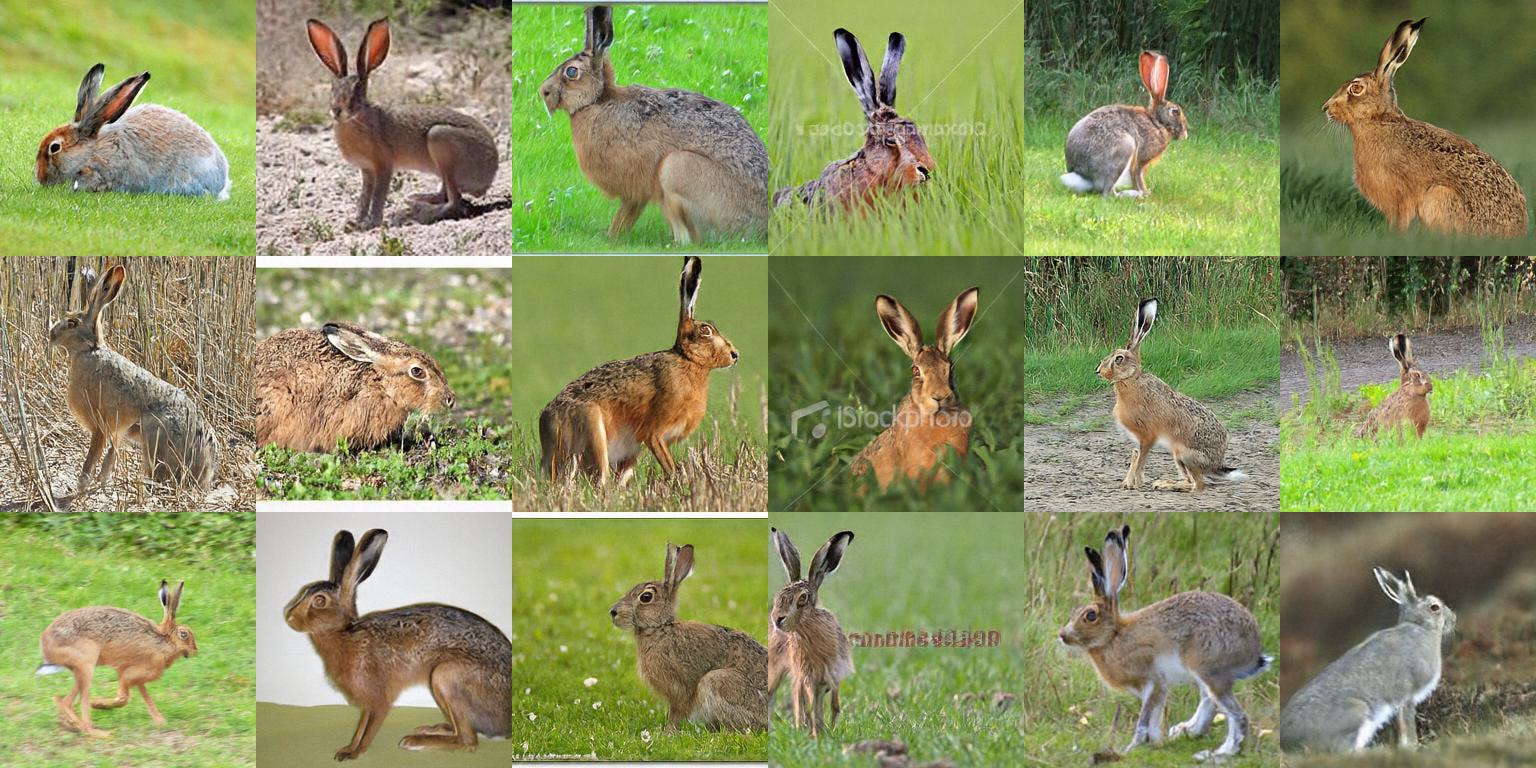}{n02326432}{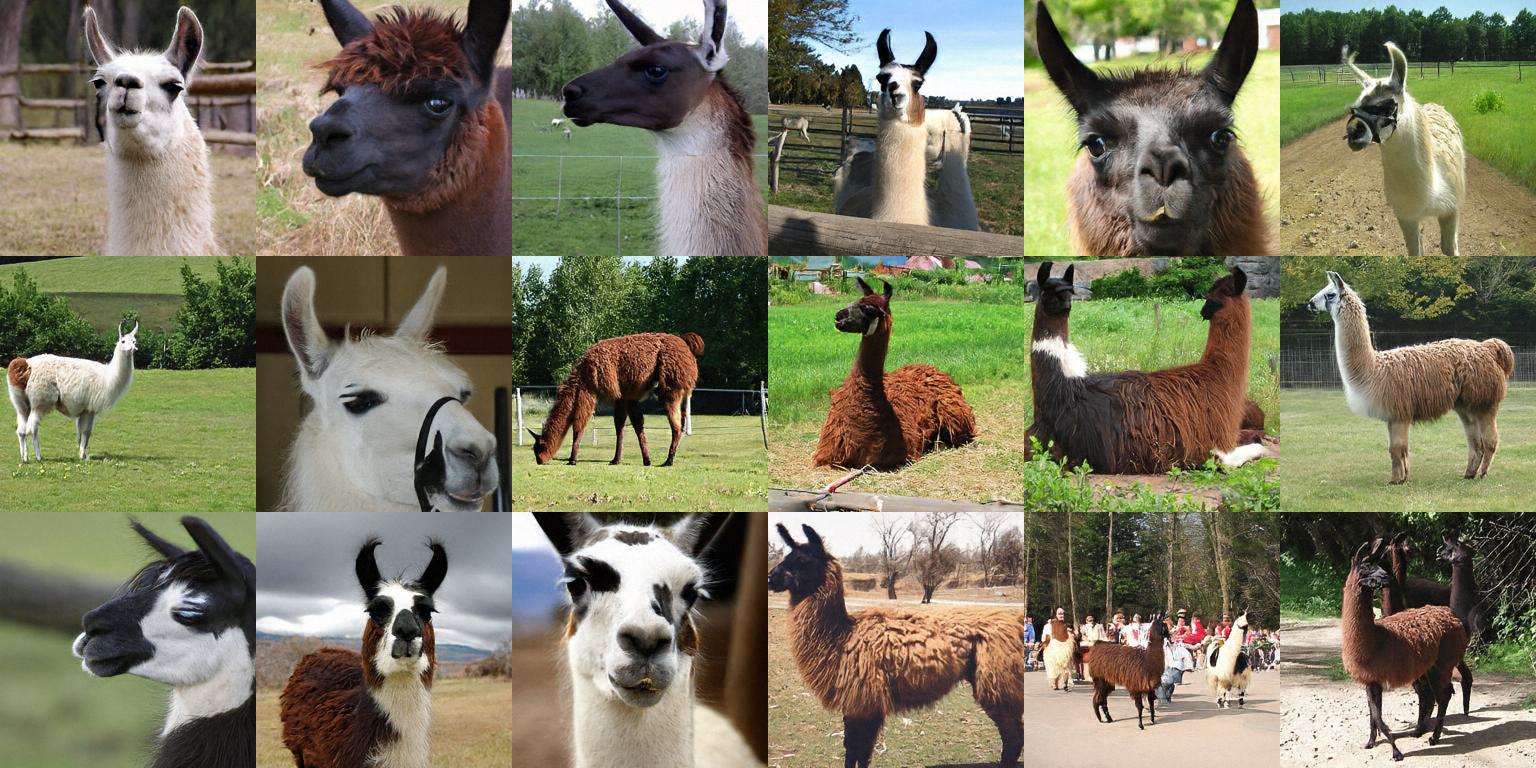}{n02437616}
  \qualrow{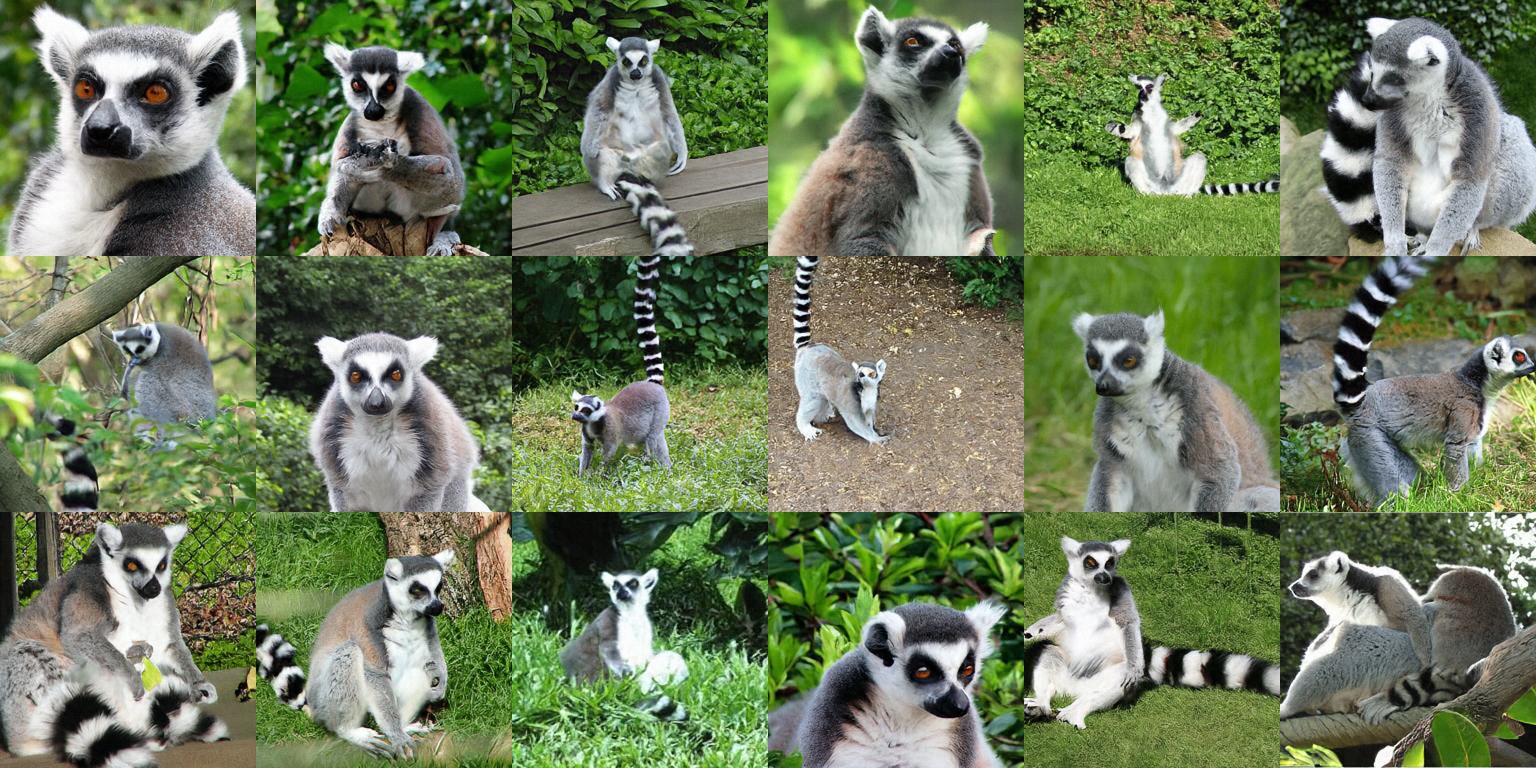}{n02497673}{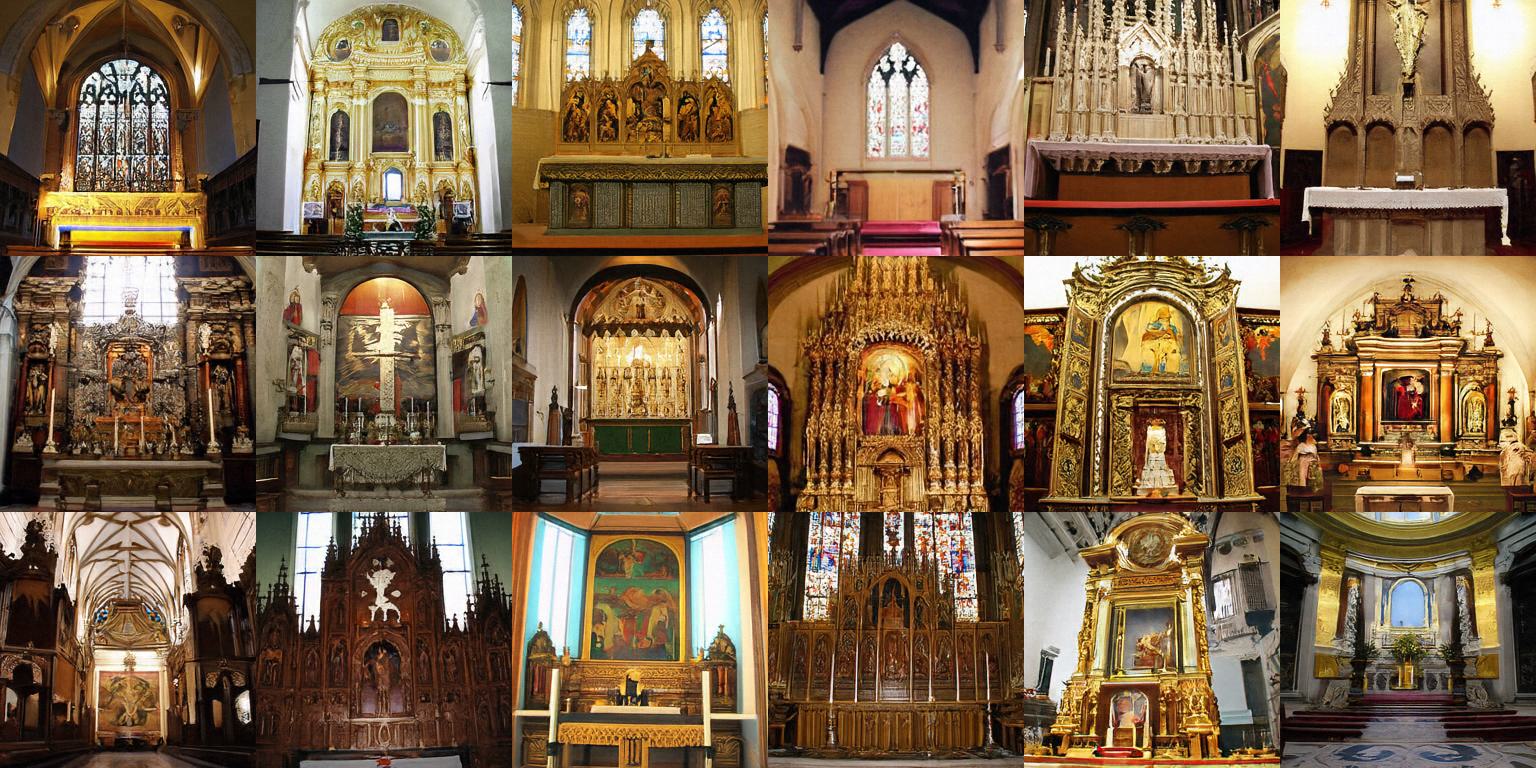}{n02699494}
  \qualcenter{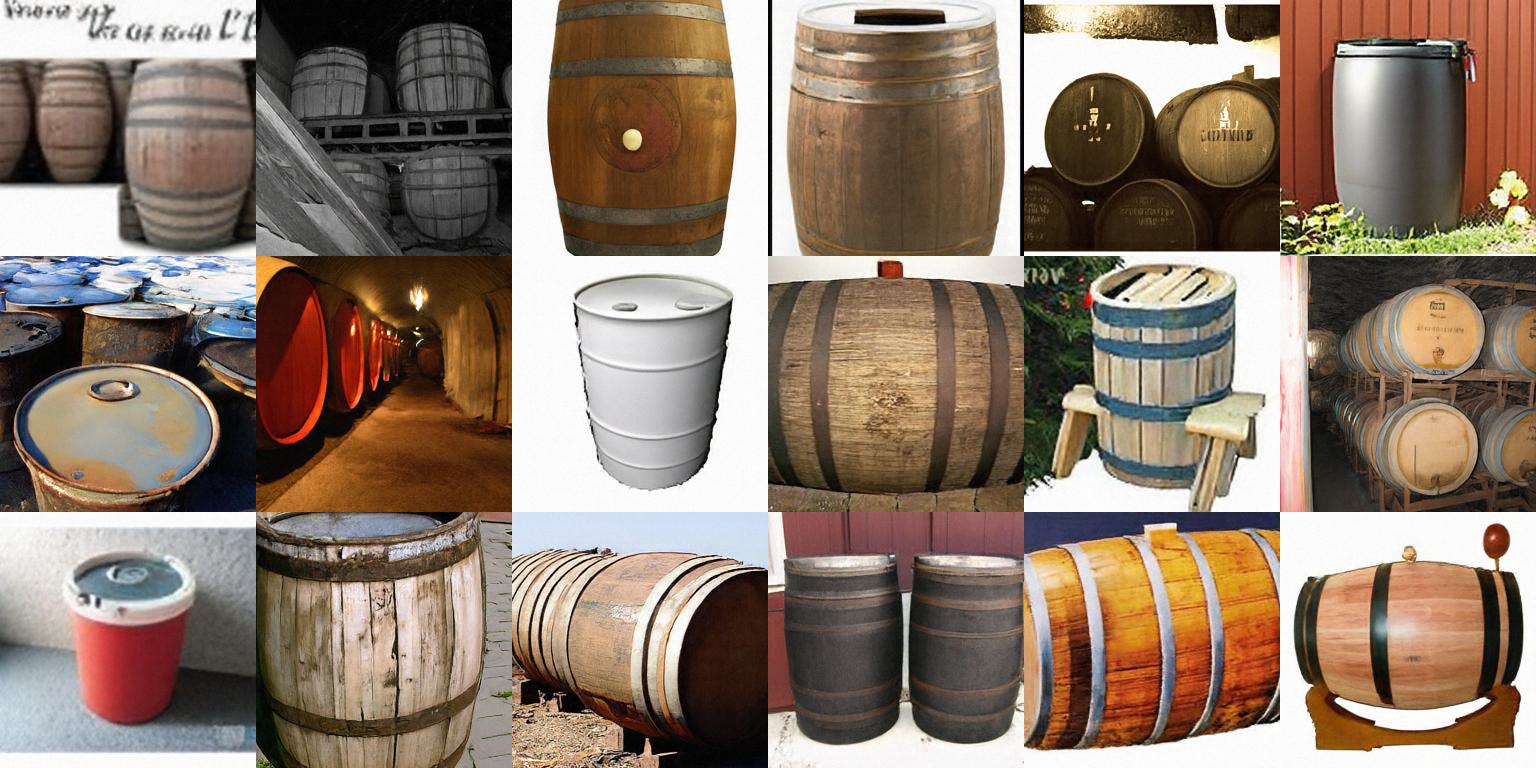}{n02795169}
  \caption{Uncurated samples from JiT-H/16 + \ours{} on ImageNet $256\times256$.}
  \label{fig:app-qualitative-5}
\end{figure}

\begin{figure}[H]
  \centering
  \qualrow{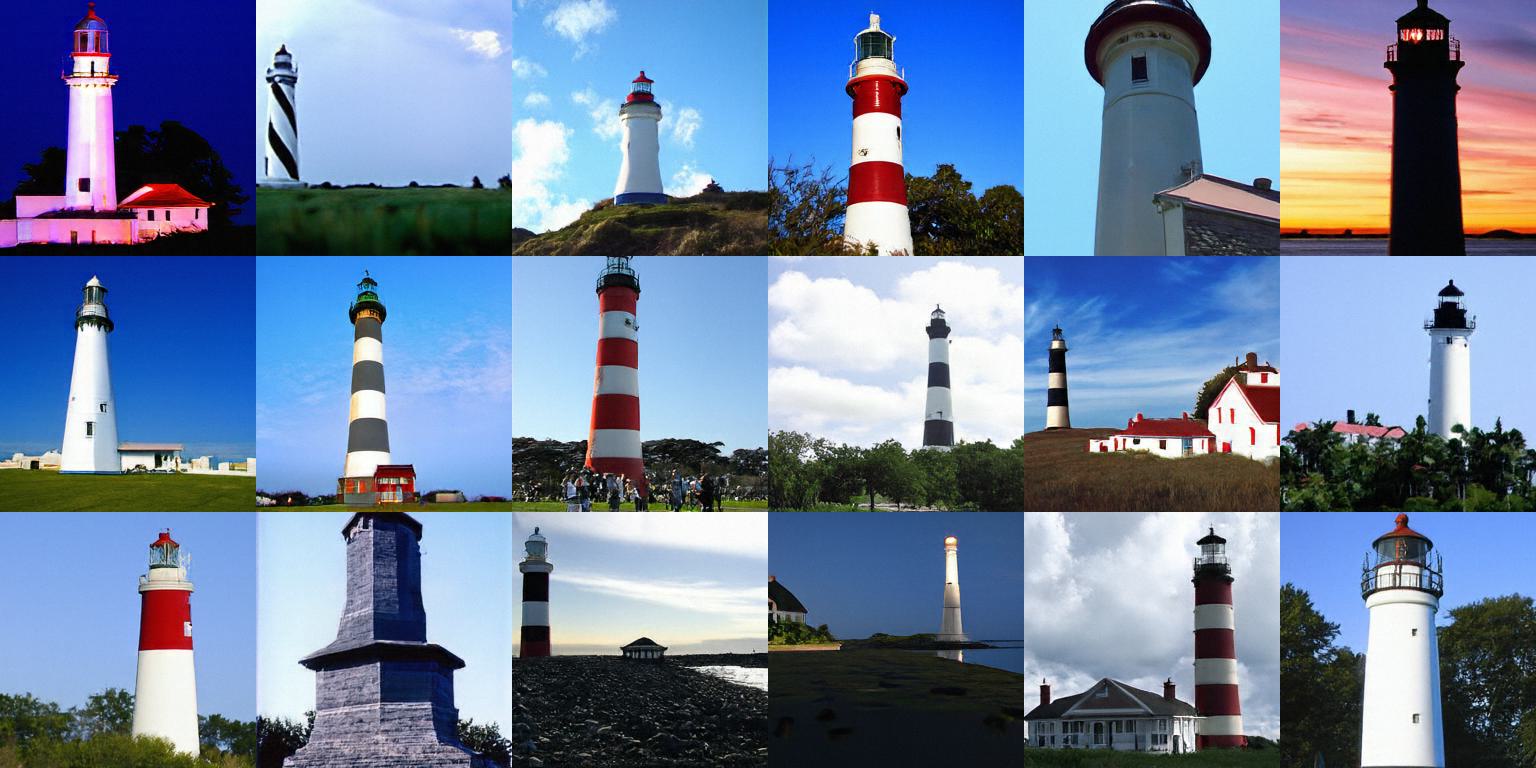}{n02814860}{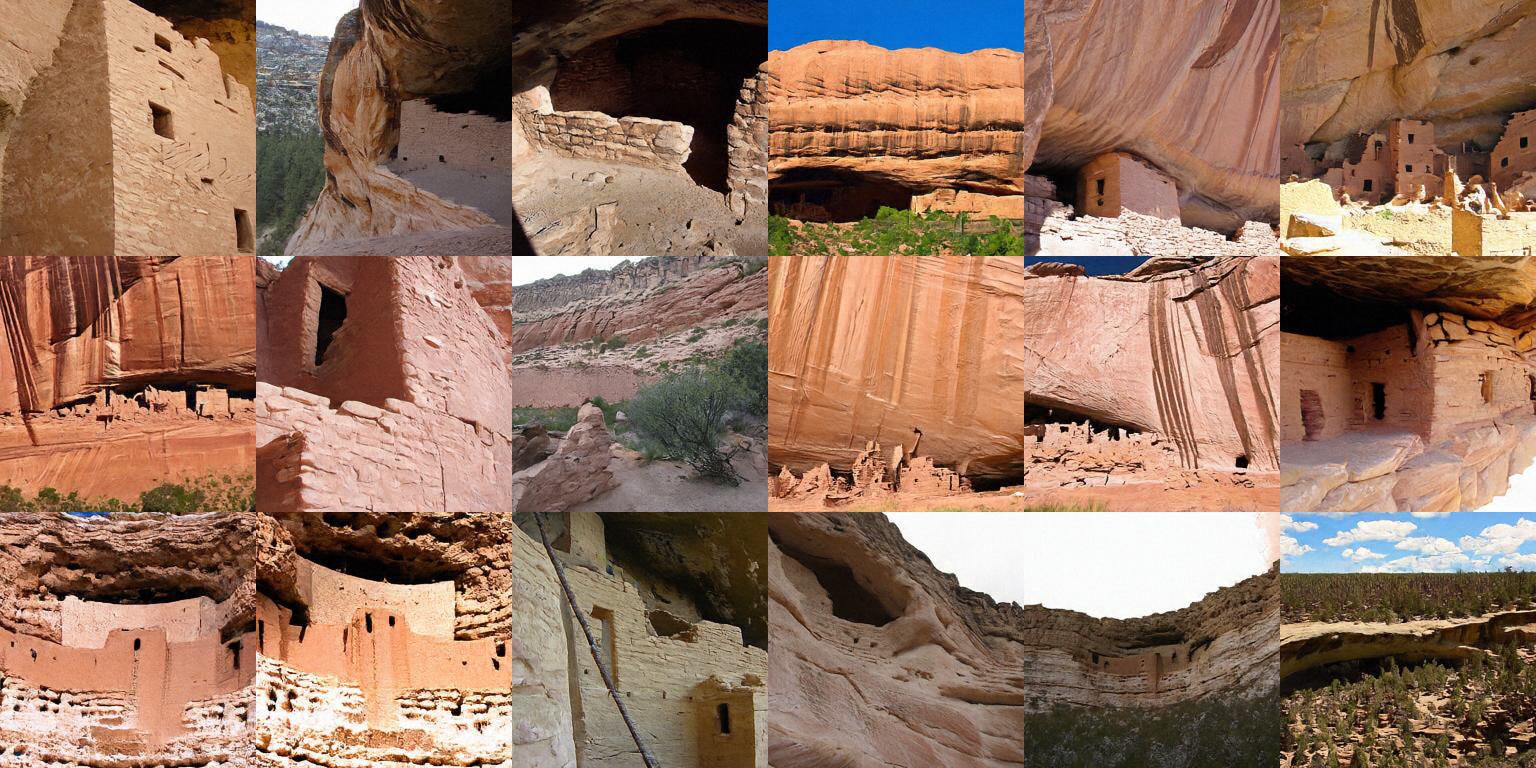}{n03042490}
  \qualrow{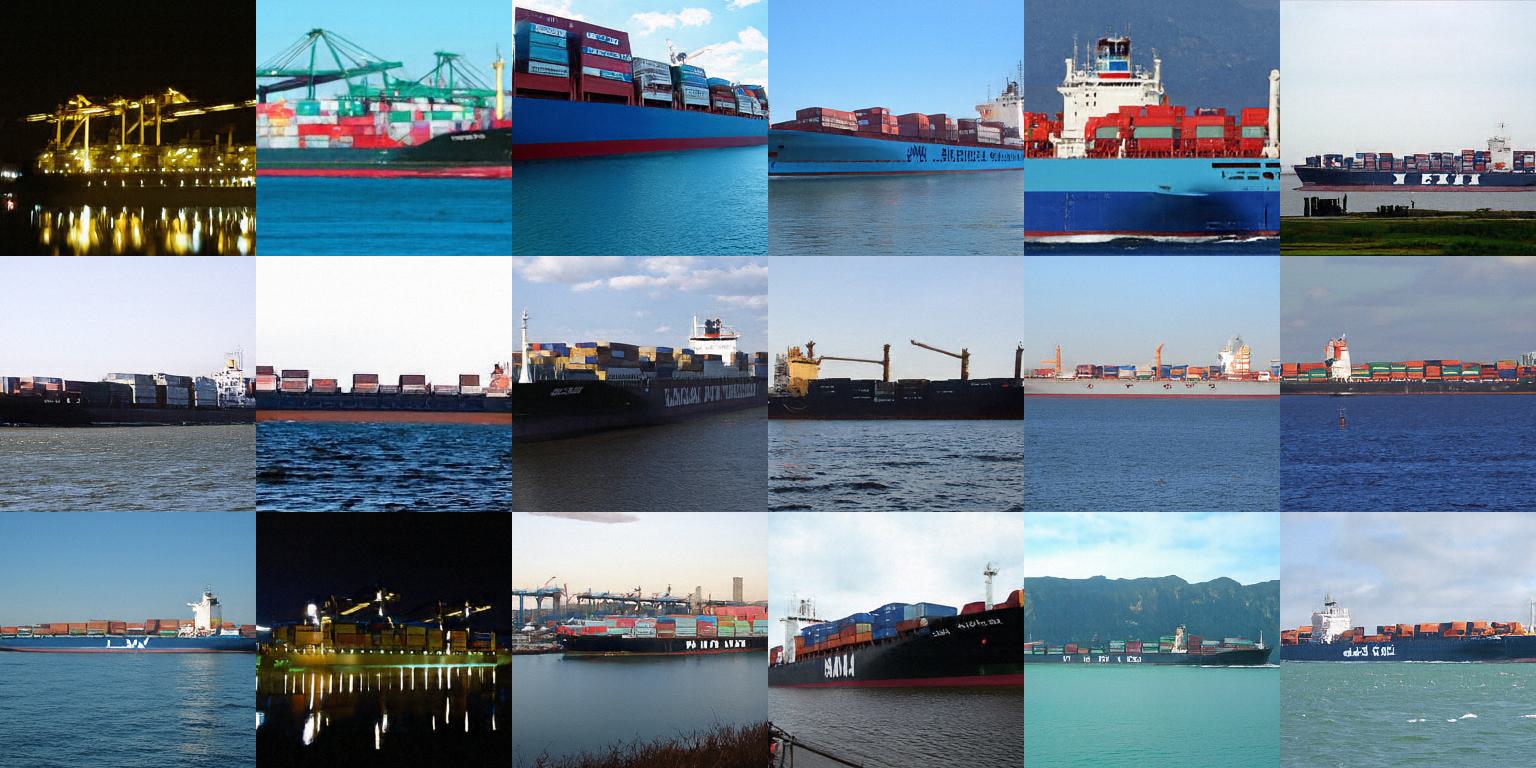}{n03095699}{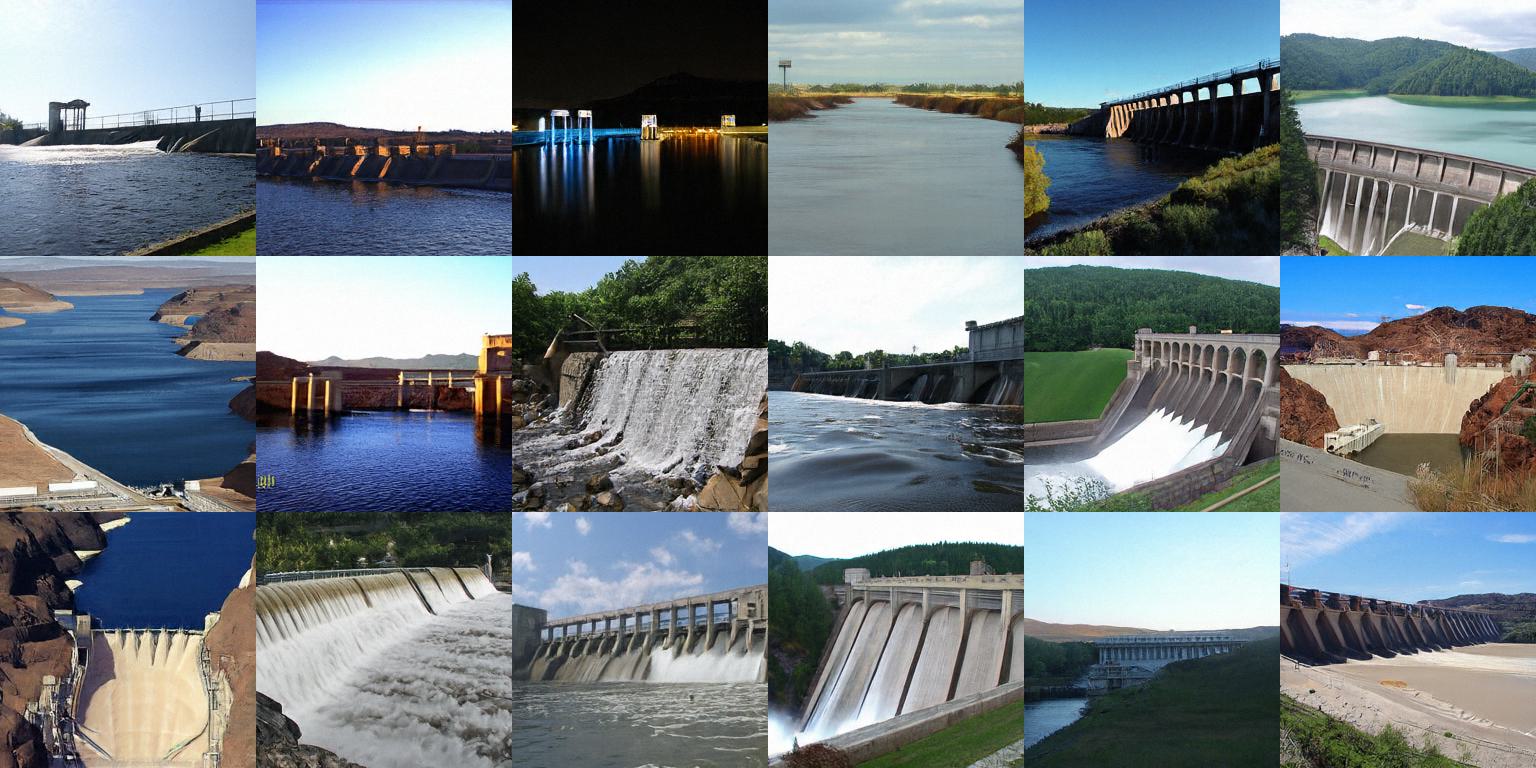}{n03160309}
  \qualrow{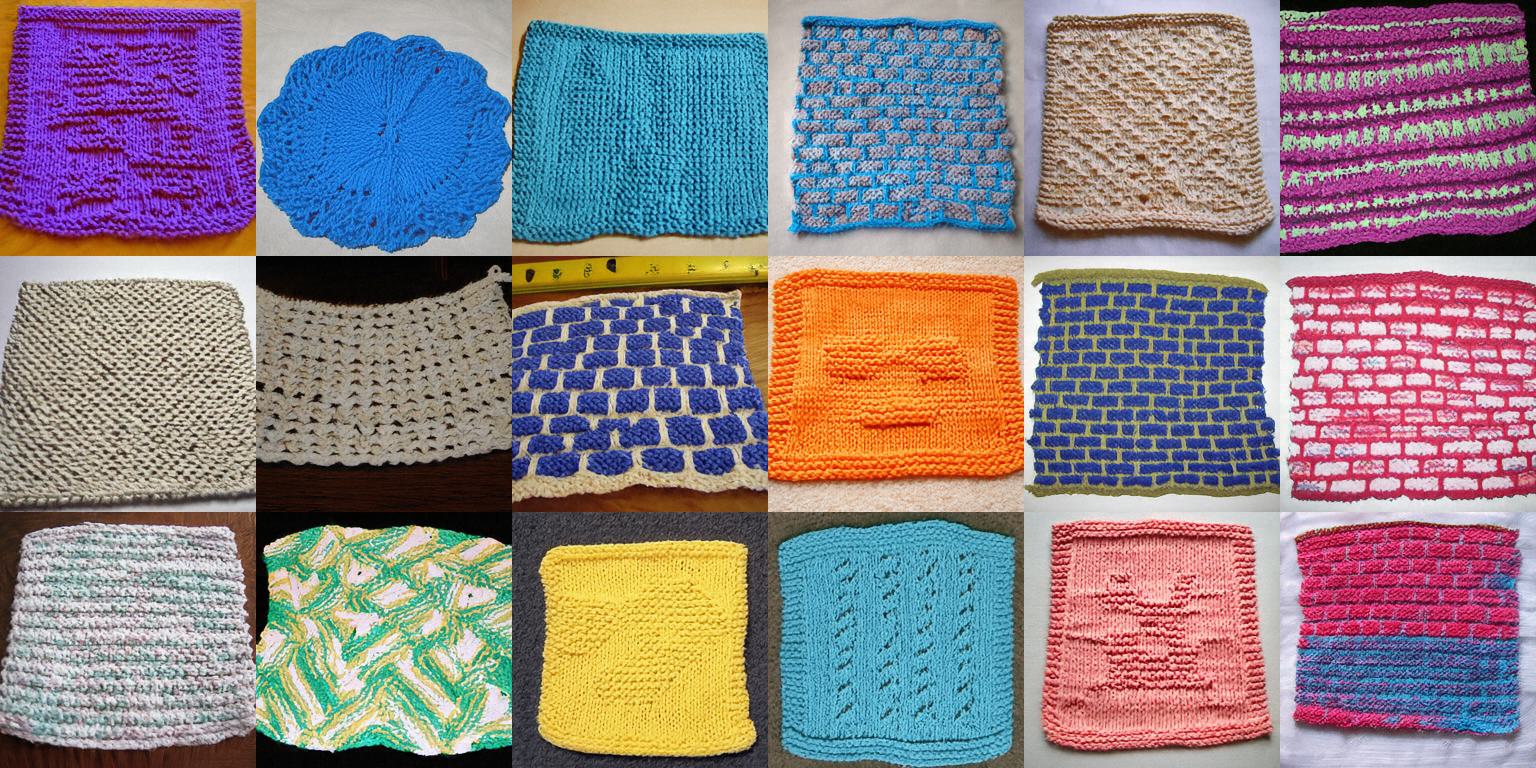}{n03207743}{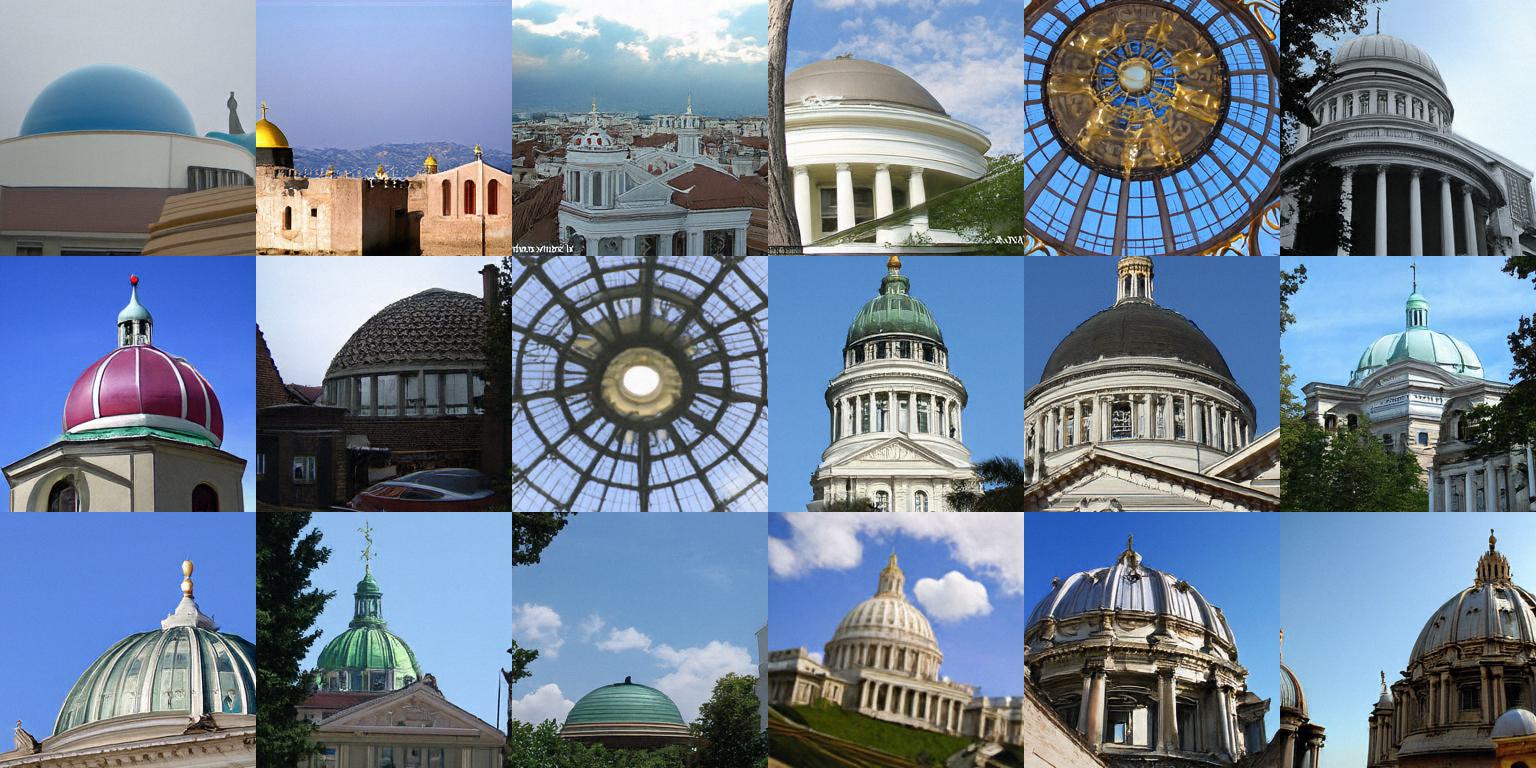}{n03220513}
  \qualrow{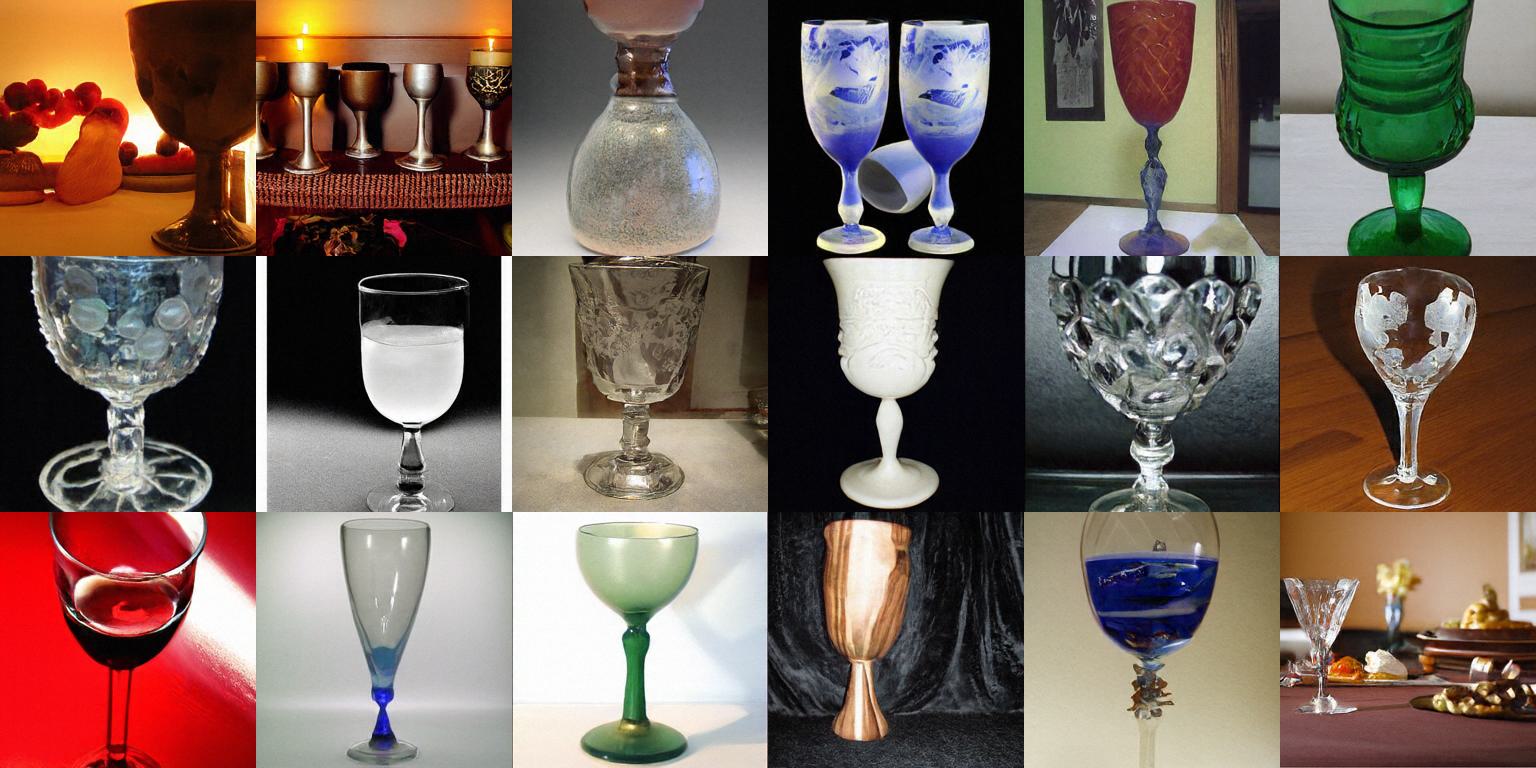}{n03443371}{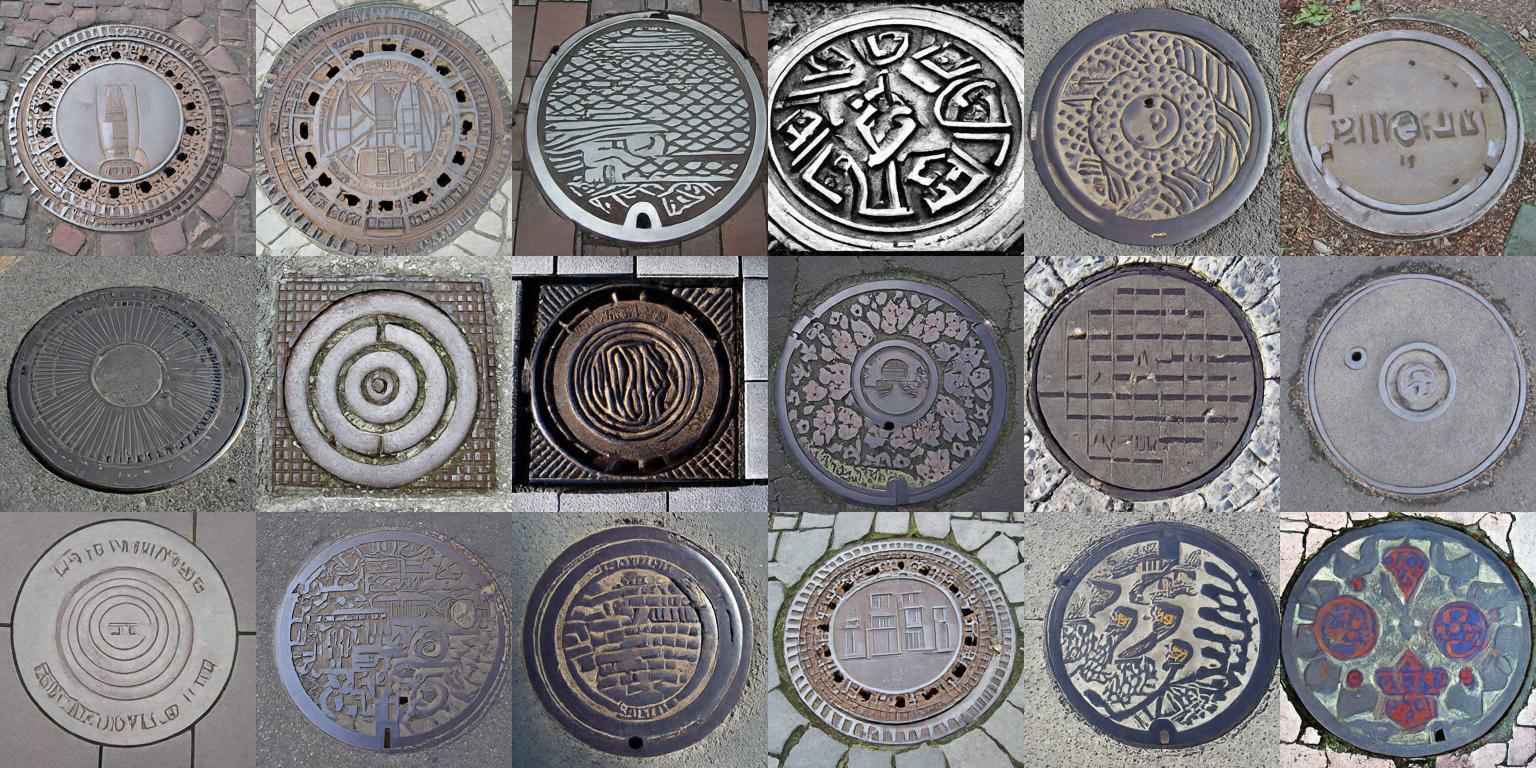}{n03717622}
  \qualcenter{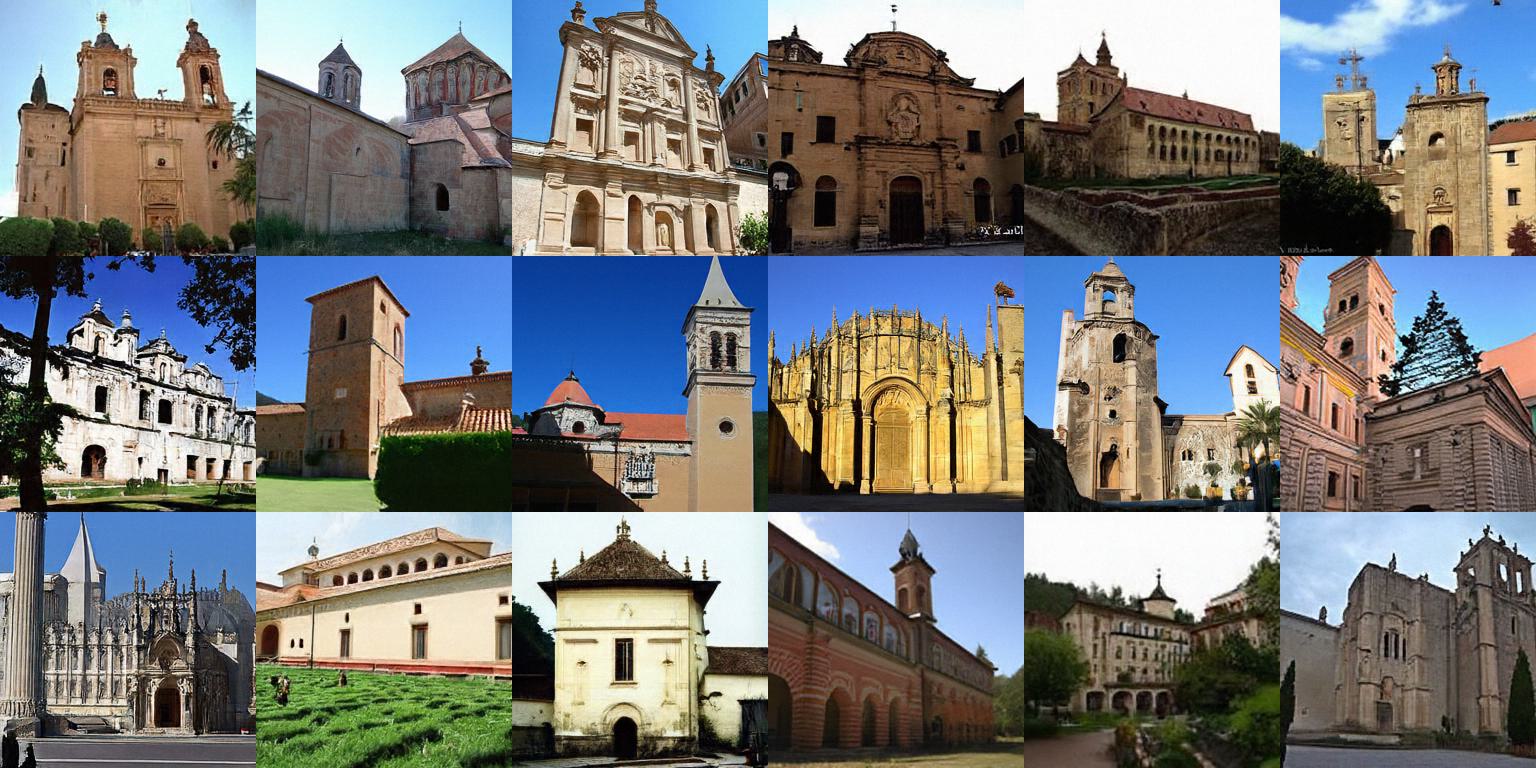}{n03781244}
  \caption{Uncurated samples from JiT-H/16 + \ours{} on ImageNet $256\times256$.}
  \label{fig:app-qualitative-6}
\end{figure}

\begin{figure}[H]
  \centering
  \qualrow{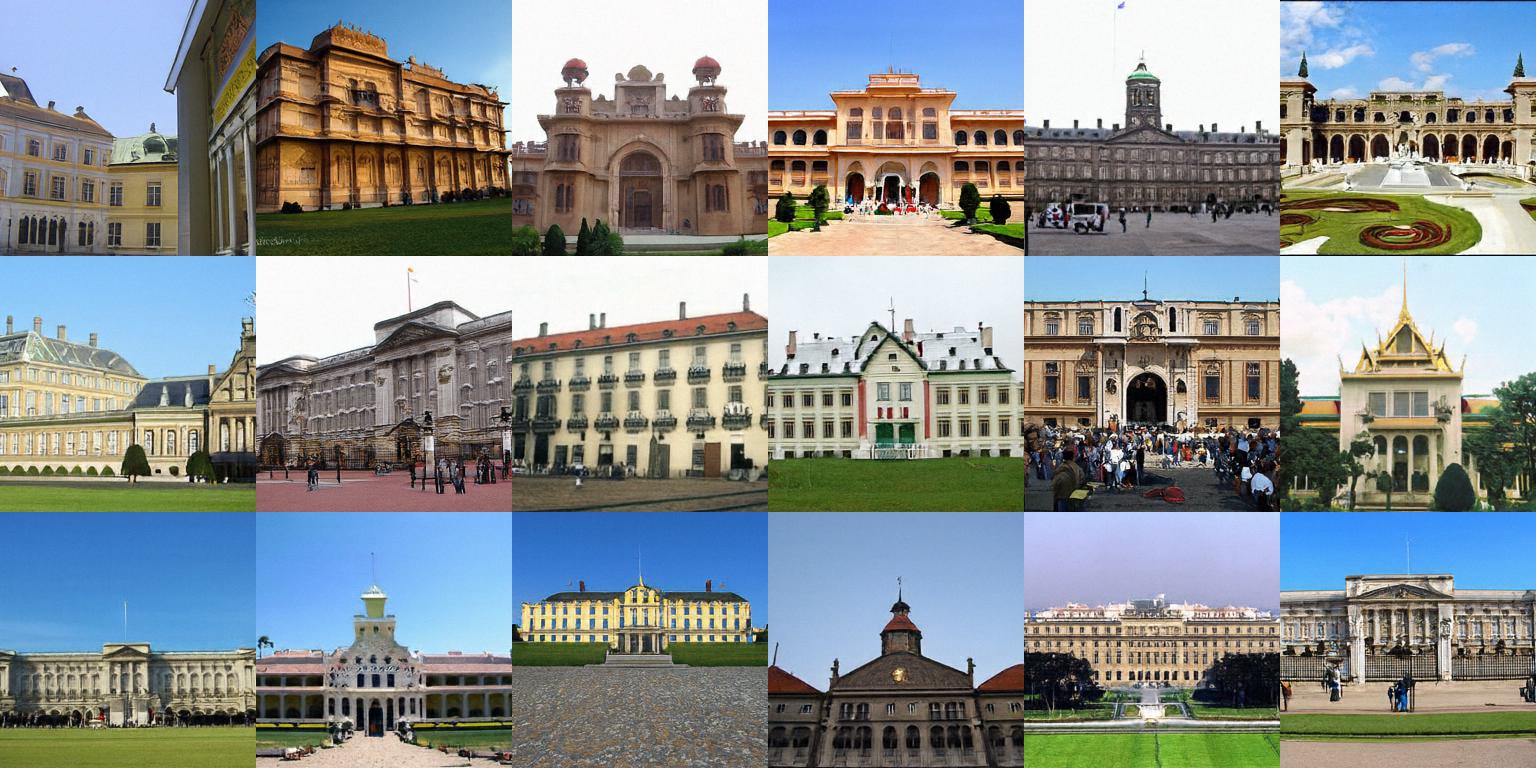}{n03877845}{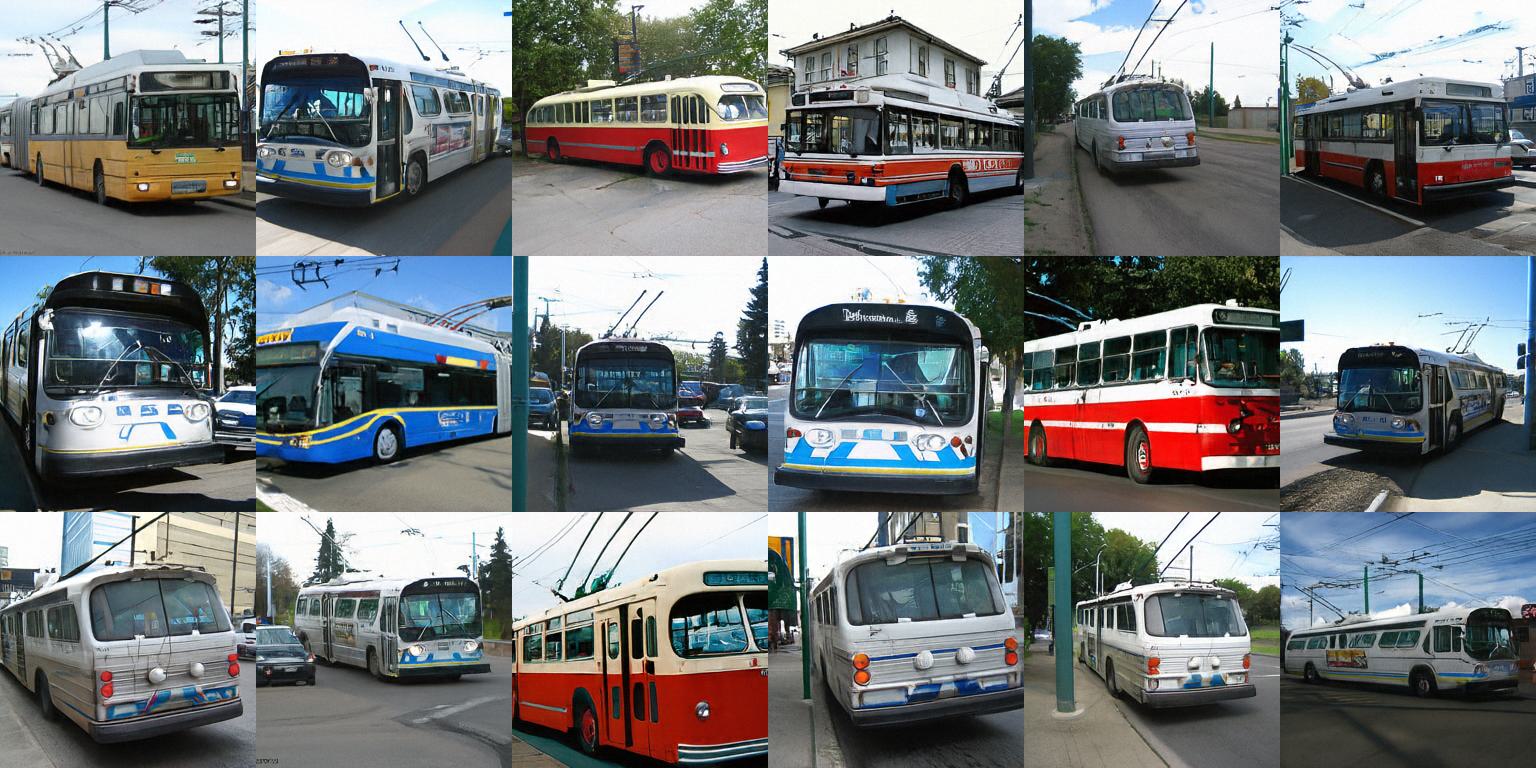}{n04487081}
  \qualrow{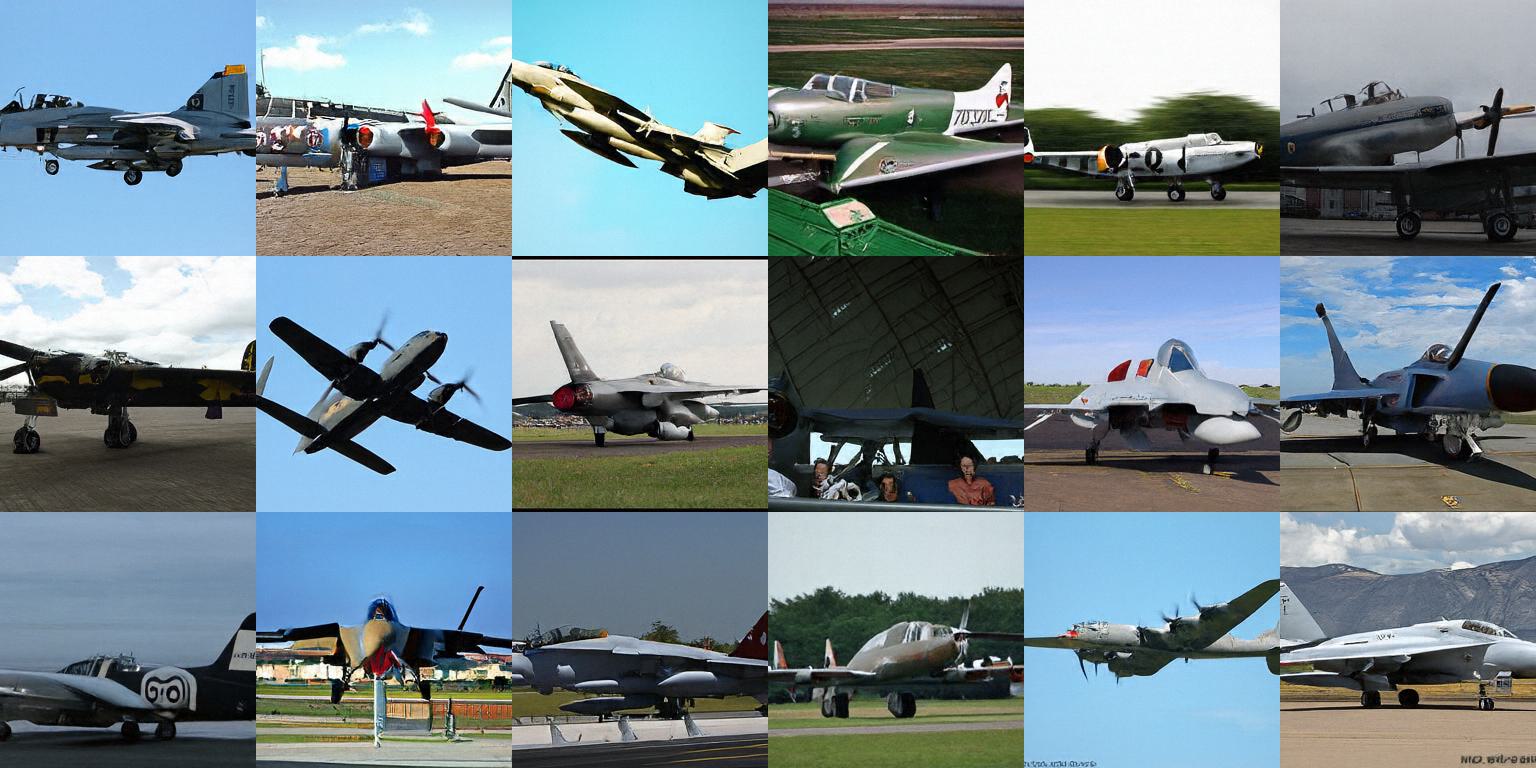}{n04552348}{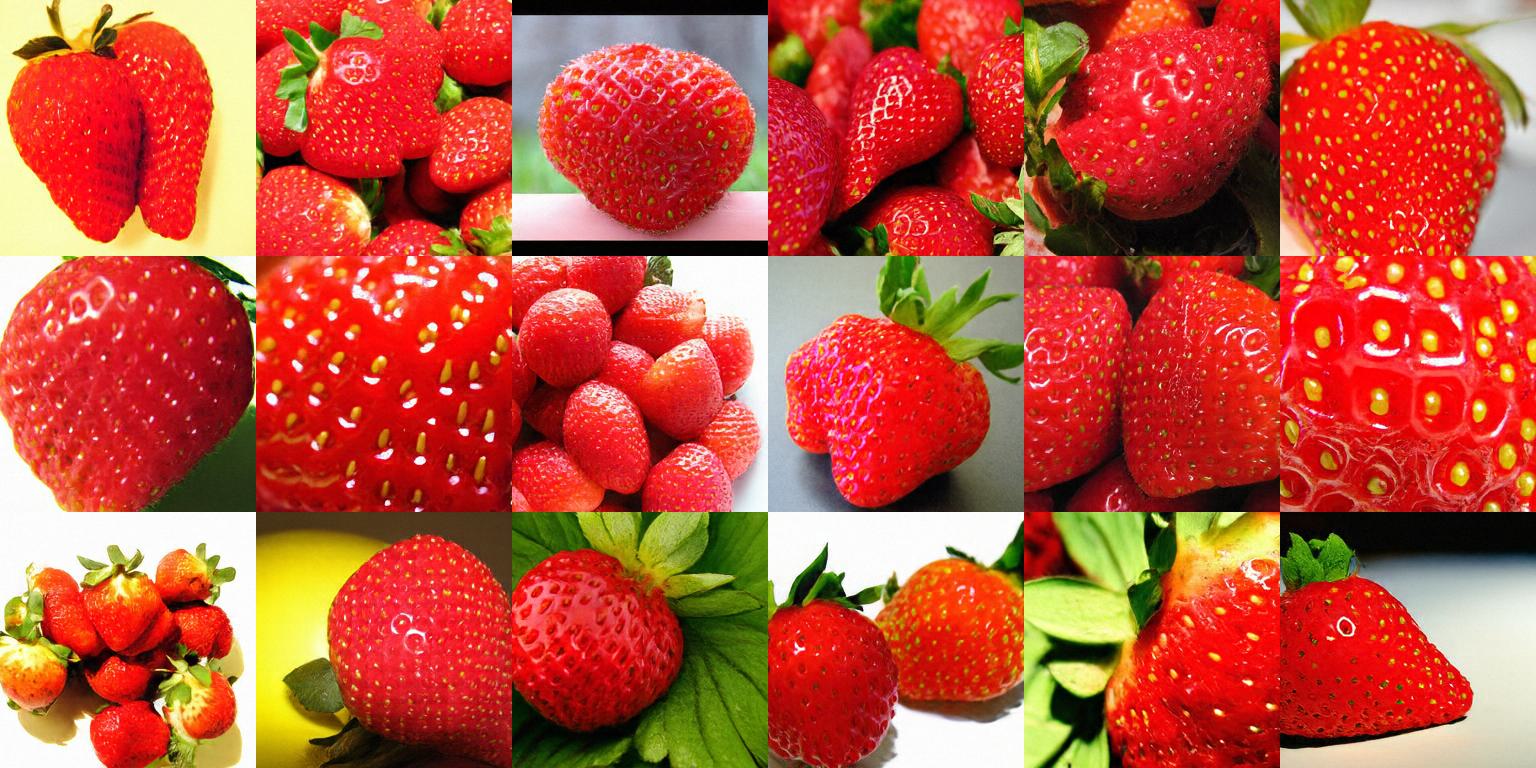}{n07745940}
  \qualrow{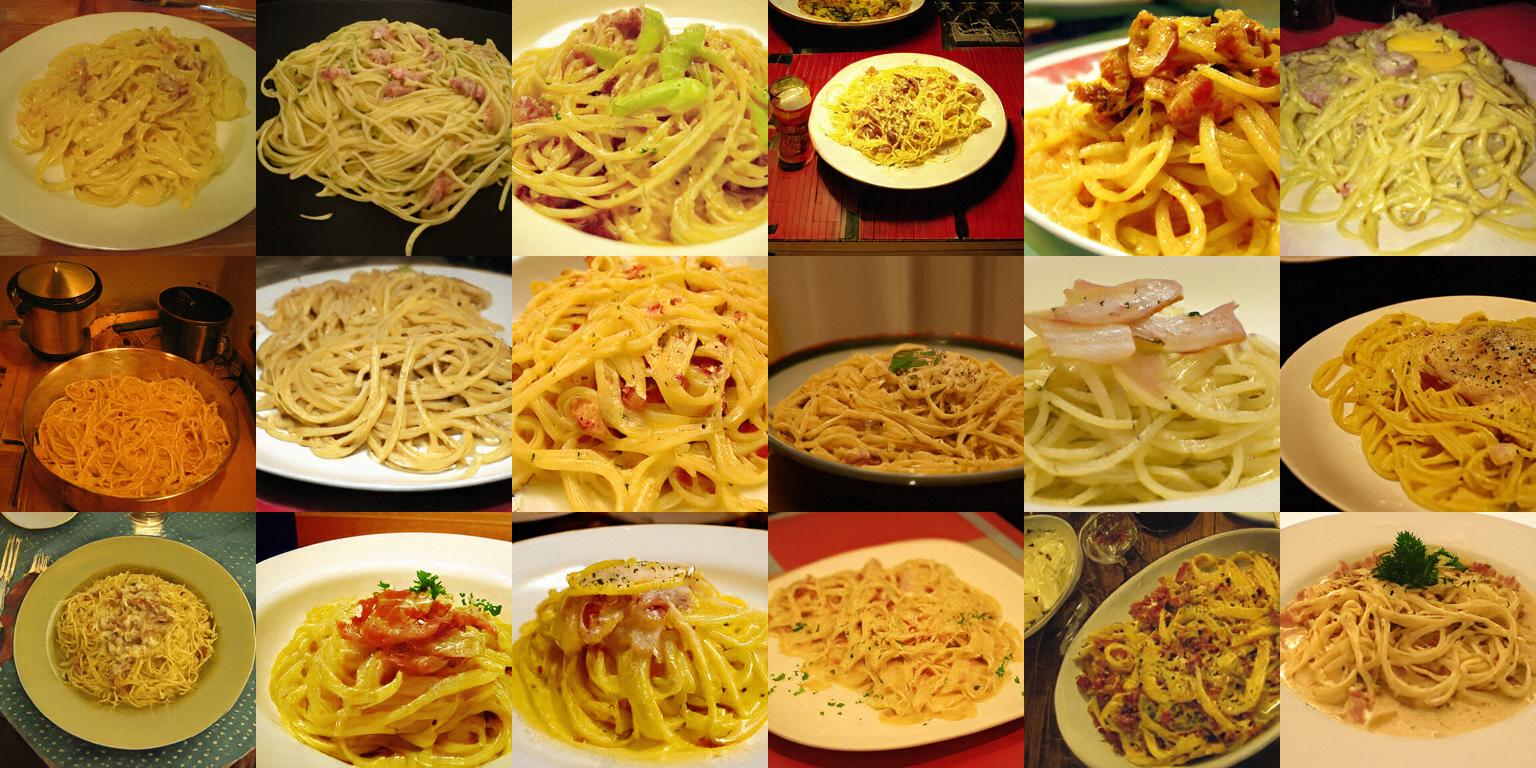}{n07831146}{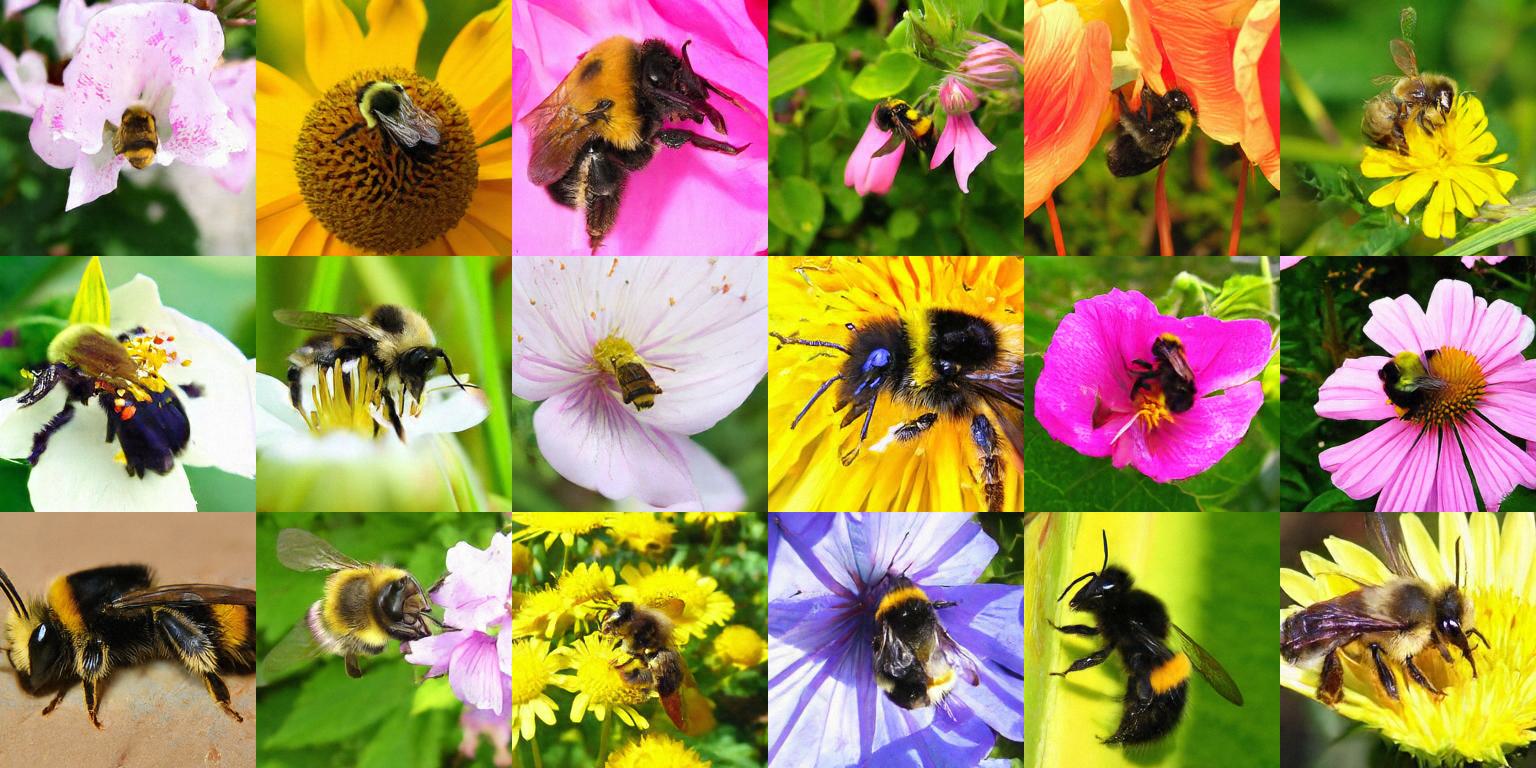}{n02206856}
  \qualrow{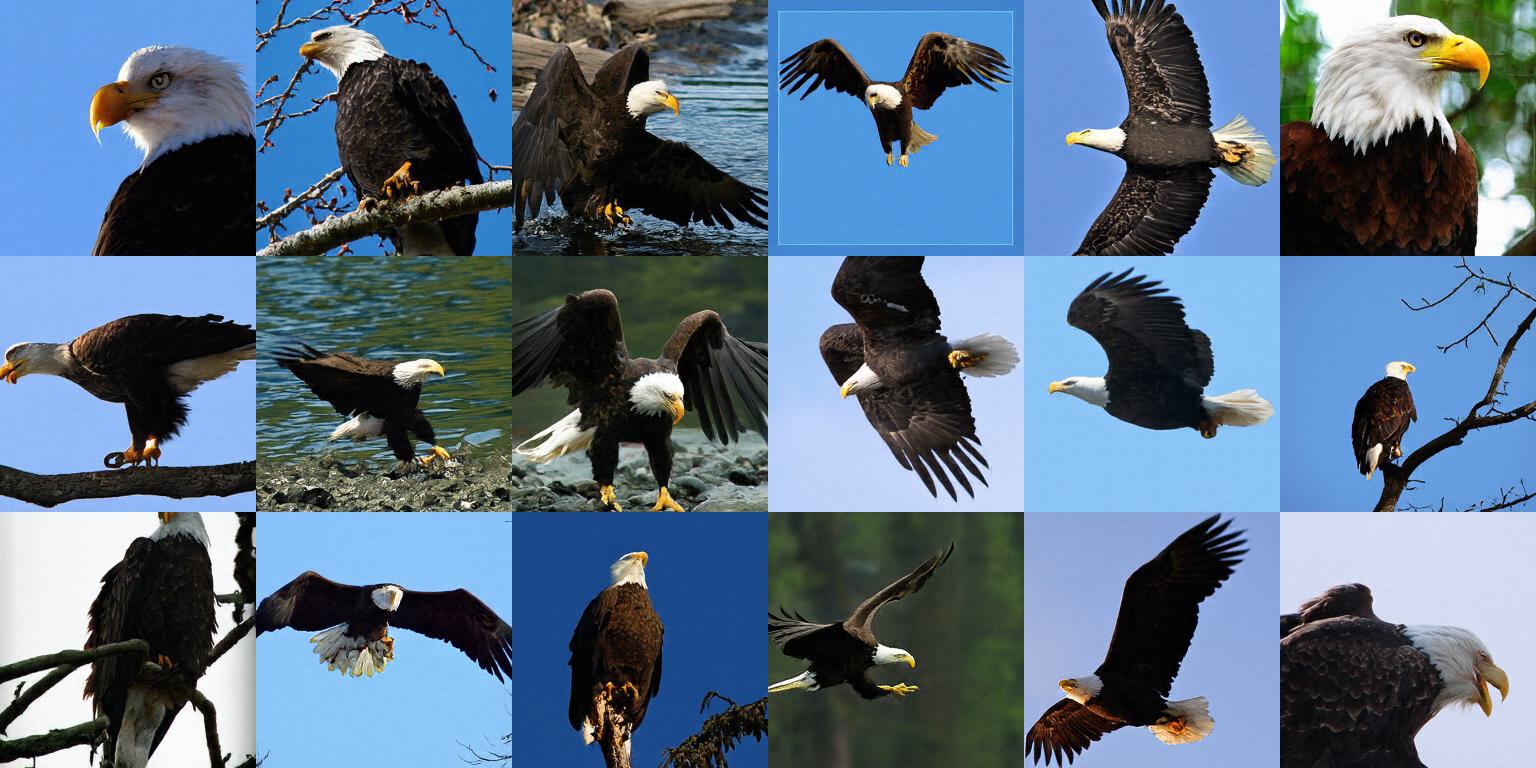}{n01614925}{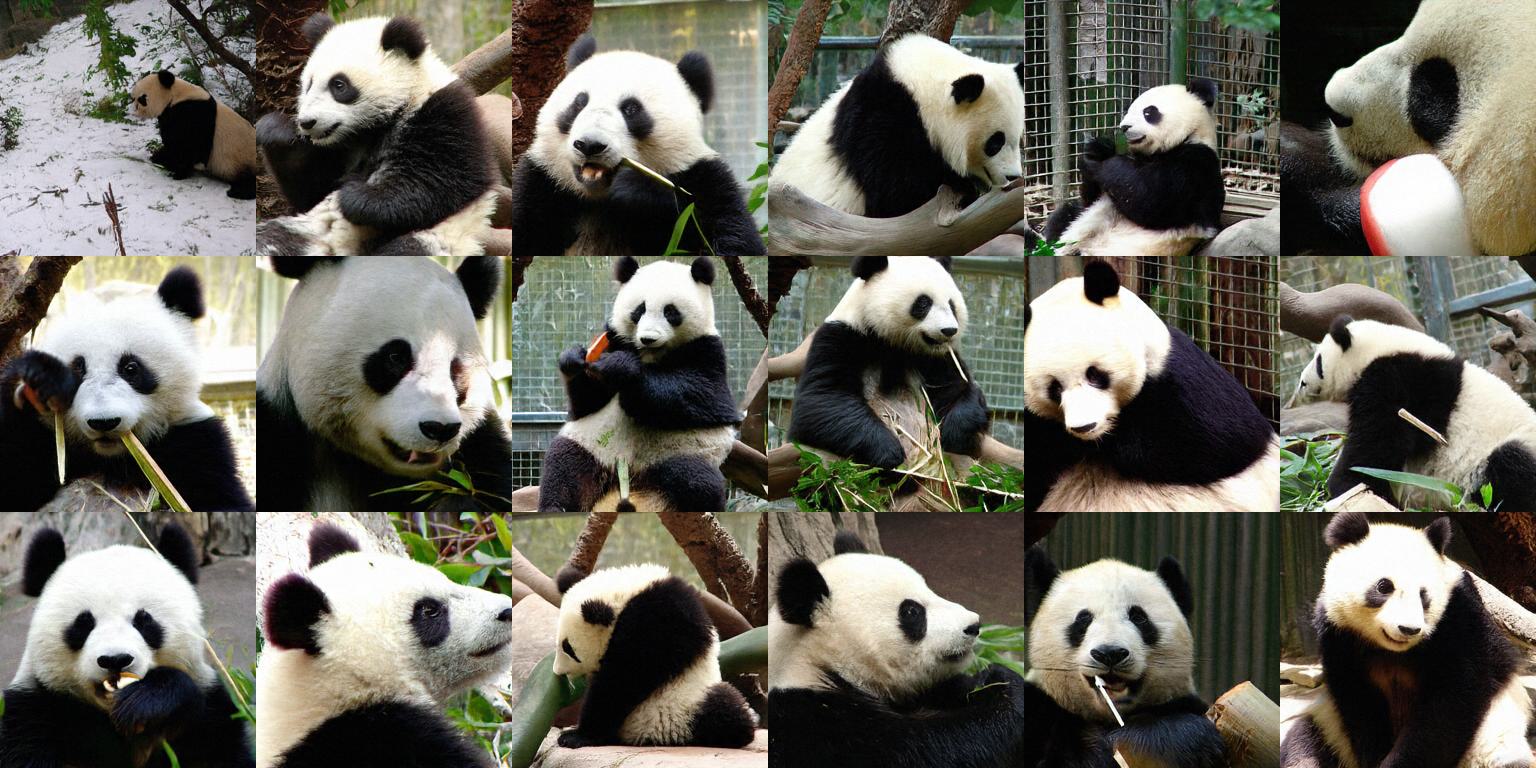}{n02510455}
  \qualcenter{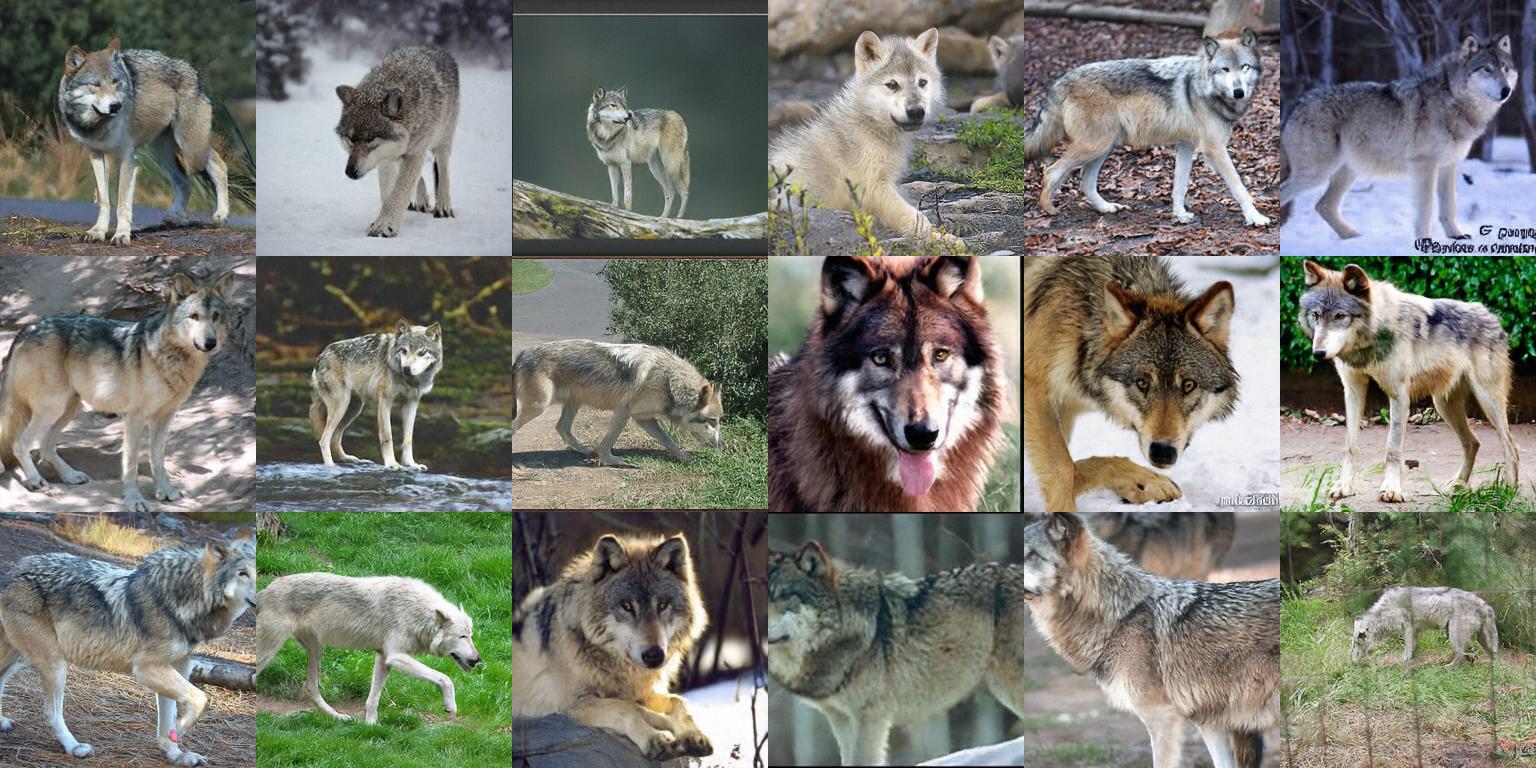}{n02114367}
  \caption{Uncurated samples from JiT-H/16 + \ours{} on ImageNet $256\times256$.}
  \label{fig:app-qualitative-7}
\end{figure}

\endgroup

\fi
\end{document}